\documentclass[10pt,twocolumn,letterpaper]{article}

\usepackage{cvpr}
\usepackage{times}
\usepackage{epsfig}
\usepackage{graphicx}
\usepackage{amsmath}
\usepackage{amssymb}
\usepackage{multirow}
\usepackage{array}
\usepackage{xcolor}
\usepackage{booktabs}
\usepackage{enumitem}

\newcommand*{\rowstyle}[1]{
	\gdef\@rowstyle{#1}
	\@rowstyle\ignorespaces%
}
\newcolumntype{=}{
	>{\gdef\@rowstyle{}}
}
\newcolumntype{+}{
	>{\@rowstyle}
}

\usepackage{caption}
\usepackage{subcaption}
\usepackage[vlined,ruled,commentsnumbered,linesnumbered]{algorithm2e}

\usepackage{listings}

\newcommand{\centered}[1]{\multicolumn{1}{c|}{\centering #1}}

\newcommand{\lyft}{Lyft\xspace}
\newcommand{\argo}{Argoverse\xspace}
\newcommand{\nusc}{nuScenes\xspace}
\newcommand{\waymo}{Waymo\xspace}
\newcommand{\kitti}{KITTI\xspace}

\newif\ifnewthreshold
\newif\ifdifficultybydistance
\difficultybydistancetrue

\usepackage[pagebackref=true,breaklinks=true,letterpaper=true,colorlinks,bookmarks=false]{hyperref}

\usepackage{amssymb}
\usepackage{amsmath,amsfonts}
\usepackage{amsopn}
\usepackage{bm} 
\usepackage{multirow}

\newcommand{\ProbOpr}[1]{\mathbb{#1}}

\newcommand{\expect}[2]{%
\ifthenelse{\equal{#2}{}}{\ProbOpr{E}_{#1}}
{\ifthenelse{\equal{#1}{}}{\ProbOpr{E}\left[#2\right]}{\ProbOpr{E}_{#1}\left[#2\right]}}} 
\newcommand{\var}[2]{%
\ifthenelse{\equal{#2}{}}{\ProbOpr{VAR}_{#1}}
{\ifthenelse{\equal{#1}{}}{\ProbOpr{VAR}\left[#2\right]}{\ProbOpr{VAR}_{#1}\left[#2\right]}}} 

\newcommand{\eat}[1]{}
\newcommand{\method}[1]{\textsc{#1}}

\newcommand{\PIXOR}{\method{PIXOR}\xspace}
\newcommand{\PRCNN}{\method{PointRCNN}\xspace}

\newcommand{\APBEV}{AP$_\text{BEV}$\xspace}
\newcommand{\AP}{AP$_\text{3D}$\xspace}

\cvprfinalcopy 
\def\cvprPaperID{5327} 

\ifcvprfinal\fi
\begin{document}
	
	\title{Train in Germany, Test in The USA: Making 3D Object Detectors Generalize}
	\author{Yan Wang$^{*1}$\hspace{20pt}
	Xiangyu Chen\thanks{\hspace{1pt}Equal contributions}\hspace{4pt}$^{1}$\hspace{20pt}
	Yurong You$^1$\hspace{20pt}
	Li Erran Li$^{2,3}$ \\
	Bharath Hariharan$^1$\hspace{20pt}
	Mark Campbell$^1$\hspace{20pt}
	Kilian Q. Weinberger$^1$\hspace{20pt}
	Wei-Lun Chao$^4$\\
		$^1$Cornell University\hspace{20pt} $^2$Scale AI\hspace{20pt}$^3$Columbia University\hspace{20pt}$^4$The Ohio State University\\
		{\tt\small \{yw763, xc429, yy785, bh497, mc288, kqw4\}@cornell.edu\hspace{5pt}  erranlli@gmail.com\hspace{5pt}
		chao.209@osu.edu}
	}

	\maketitle
\begin{abstract}
	In the domain of autonomous driving, deep learning has substantially improved the 3D object detection accuracy for LiDAR and stereo camera data alike. While deep networks are great at generalization, they are also notorious to over-fit to all kinds of spurious artifacts, such as brightness, car sizes and models, that may appear consistently throughout the data. In fact, most datasets for autonomous driving are collected within a narrow subset of cities within one country, typically  under similar weather conditions. In this paper we consider the task of adapting 3D object detectors from one dataset to another. We observe that na\"ively, this appears to be a very challenging task, resulting in drastic drops in accuracy levels. We provide extensive experiments to investigate the true adaptation challenges and arrive at a surprising conclusion: the primary adaptation hurdle to overcome are differences in car sizes across geographic areas. A simple correction based on the average car size
	yields a strong correction of the adaptation gap. Our proposed method is simple and easily incorporated into most 3D object detection frameworks. It provides a first baseline for 3D object detection adaptation across countries, and gives hope that the underlying problem may be more within grasp than one may have hoped to believe. Our code is available at \url{https://github.com/cxy1997/3D_adapt_auto_driving}. 
\end{abstract}

\begin{figure}[t]
	\centering
	\small
	\begin{subfigure}[b]{\linewidth}
		\includegraphics[width=\linewidth]{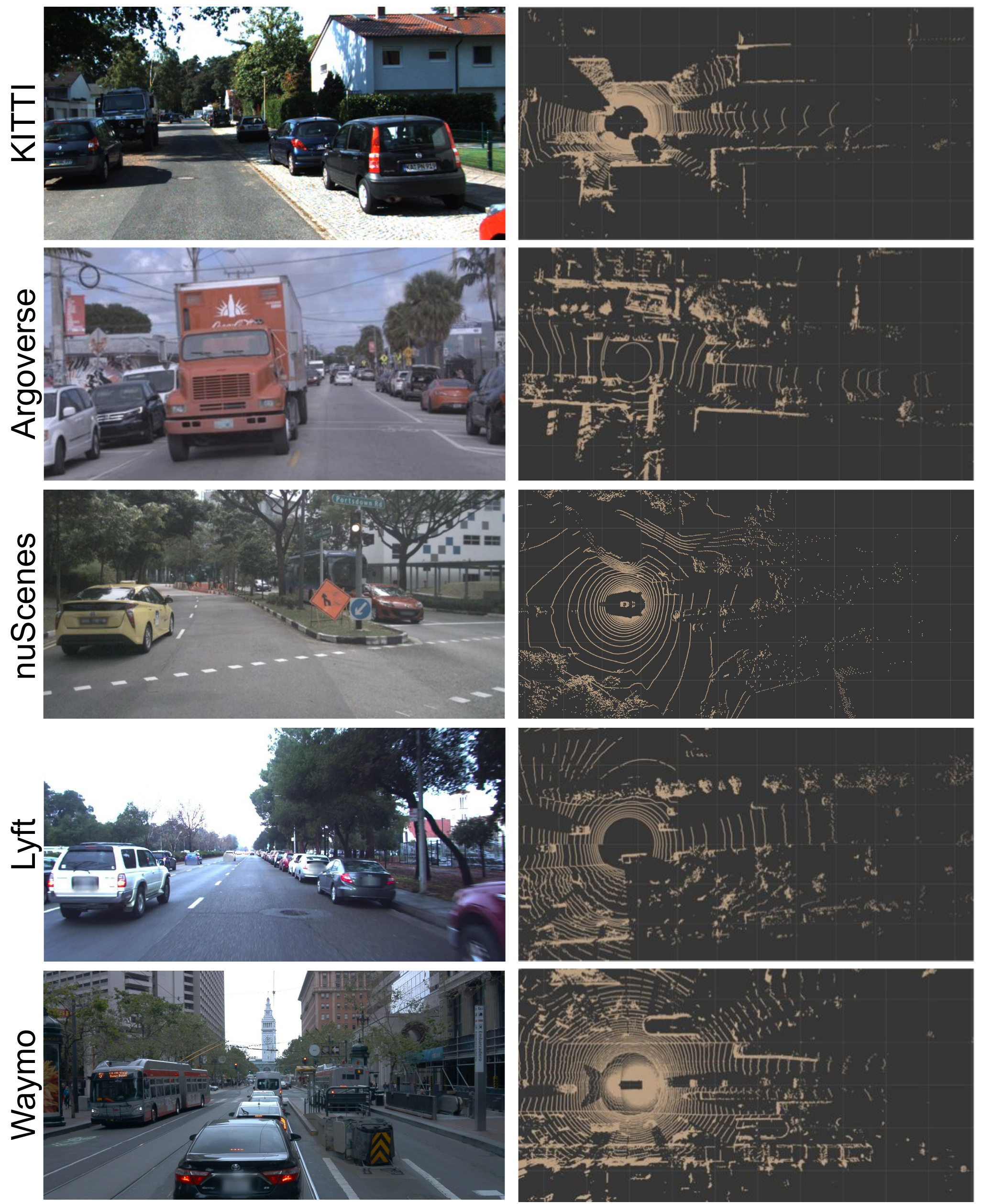}
	\end{subfigure}
	\vskip -5pt
	\caption{\small\textbf{Datasets.} We show frontal view images (left) and the corresponding LiDAR signals (right) from the bird's-eye view for five datasets: \kitti~\cite{geiger2013vision,geiger2012we}, \argo~\cite{argoverse}, \nusc~\cite{nuscenes2019}, \lyft~\cite{lyft2019}, and \waymo~\cite{waymo_open_dataset}. These datasets not only capture scenes at different geo-locations, but also use different LiDAR models, making generalizing 3D object detectors a challenging problem.}
	\label{fig:1}
	\vskip -10pt
\end{figure}

\section{Introduction}
\label{sec:intro}
Autonomous cars need to accurately detect and localize vehicles and pedestrians in 3D to drive safely.
As such, the past few years have seen a flurry of interest on the problem of 3D object detection, resulting in large gains in accuracy on the \kitti benchmark~\cite{chen2017multi,chen2019fast,chen2020dsgn,du2018general,geiger2013vision,geiger2012we,konigshofrealtime,ku2018joint,lang2019pointpillars,li2019gs3d,li2019stereo,liang2019multi,liang2018deep,meyer2019lasernet,pon2019object,qi2018frustum,shi2019pointrcnn,shi2020points,shi2020point,shi2019pv,xu2020zoomnet,yan2018second,yang2018pixor,yang20203dssd,yang2019std,zhou2019end,zhou2018voxelnet}.
However, in the excitement this has garnered, it has often been forgotten that \kitti is a fairly small ($\sim$15K scenes) object detection dataset obtained from a narrow domain: it was collected using a fixed sensing apparatus by driving through a mid-sized German city and the German countryside, in clear weather, during the day.
Thus, the 3D object detection algorithms trained on \kitti may  have picked up all sorts of biases: they may expect the road to be visible or the sky to be blue. They may identify only certain brands of cars, and might have even over-fit to the idiosyncrasies of German drivers and pedestrians.
Carrying these biases over to a new environment in a different part of the world might cause the object detector to miss cars or pedestrians, with devastating consequences~\cite{uber}.

It is, therefore, crucial that we (a) understand the biases that our 3D object detectors are picking up before we deploy them in safety-critical applications, and (b) identify techniques to mitigate these biases. 
Our goal in this paper is to address both of these challenges.

Our first goal is to understand if any biases have crept into current 3D object detectors.
For this, we leverage multiple recently released datasets with similar types of sensors to \kitti~\cite{geiger2013vision,geiger2012we} (cameras and LiDAR) and with 3D annotations, each of them collected in different cities~\cite{waymo_open_dataset, nuscenes2019,argoverse,lyft2019} (see~\autoref{fig:1} for an illustration).
Interestingly, they are also recorded with different sensor configurations (\ie, the LiDAR and camera models as well as their mounting arrangements can be different). 
We first train two representative LiDAR-based 3D object detectors (\PIXOR~\cite{yang2018pixor} and \PRCNN~\cite{shi2019pointrcnn}) on each dataset and test on the others.
We find that when tested on a different dataset, 3D object detectors fail dramatically: a detector trained on \kitti performs \textbf{36 percent worse} on \waymo~\cite{waymo_open_dataset} compared to the one trained on \waymo.
This indicates that the detector has indeed \emph{over-fitted} to its training domain.

What domain differences are causing such catastrophic failure?
One can think of many possibilities.
There may be differences in low-level statistics of the images.
The LiDAR sensors might have more or fewer beams, and may be oriented differently.
But the differences can also be in the \emph{physical world} being sensed.
There may be differences in the number of vehicles, their orientation, and also their sizes and shapes.
We present an extensive analysis of these potential biases that points to one major issue --- statistical differences in the sizes and shapes of cars.

In hindsight, this difference makes sense. 
The best selling car in the USA is a 5-meter long truck (Ford F-series)~\cite{usatop}, while the best selling car in Germany is a 4-meter long compact car (Volkswagen Golf\footnote{\url{https://www.best-selling-cars.com/germany/2019-q1-germany-best-selling-car-brands-and-models/}}).
Because of such differences, cars in KITTI tend to be smaller than cars in other datasets, a bias that 3D object detectors happily learn.
As a counter to this bias, we propose an extremely simple approach that leverages aggregate statistics of car sizes (\ie, mean)
to correct for this bias, \emph{in both the output annotations and the input signals}.
Such statistics might be acquired from the department of motor vehicles, or car sales data.
This single correction results in a massive improvement in cross-dataset performance, raising the 3D easy part average precision by $41.4$ points and results in a much more robust 3D object detector.

Taken together, our contributions are two-fold:
\begin{itemize}[noitemsep]
\item We present an extensive evaluation of the domain differences between self-driving car environments and how they impact 3D detector performance. Our results suggest a single core issue: size statistics of cars in different locations.
\item We present a simple and effective approach to mitigate this issue by using easily obtainable aggregate statistics of car sizes, and show dramatic improvements in cross-dataset performance as a result.
\end{itemize} 
Based on our results, we recommend that vision researchers and self-driving car companies alike be cognizant of such domain differences for large-scale deployment of 3D detection systems.

\section{Related Work}
\label{sec:related}

We review 3D object detection for autonomous driving, and domain adaptation for 2D segmentation and detection in street scenes.

\noindent\textbf{LiDAR-based detection.} 
Most existing techniques of 3D object detection use LiDAR (sometimes with images) as the input signal, which provides accurate 3D points of the surrounding environment. The main challenge is thus on properly encoding the points so as to predict point labels or draw bounding boxes in 3D to locate objects. Frustum PointNet~\cite{qi2018frustum} applies PointNet~\cite{qi2017pointnet,qi2017pointnet++} to each frustum proposal from a 2D object detector; 
\PRCNN \cite{shi2019pointrcnn} learns 3D proposals from PointNet++ features~\cite{qi2017pointnet++}.
MV3D~\cite{chen2017multi} projects LiDAR points into frontal and bird's-eye views (BEV) to obtain multi-view features; PIXOR~\cite{yang2018pixor} and LaserNet~\cite{meyer2019lasernet} show that properly encoding features in one view is sufficient to localize objects. VoxelNet~\cite{zhou2018voxelnet} and PointPillar~\cite{lang2019pointpillars} encode 3D points into voxels and extracts features by 3D convolutions and PointNet. UberATG-ContFuse~\cite{liang2018deep} and UberATG-MMF~\cite{liang2019multi} perform continuous convolutions~\cite{wang2018deep} to fuse visual and LiDAR features.

\noindent\textbf{Image-based detection.}
While providing accurate 3D points, LiDAR sensors are notoriously expensive. A 64-line LiDAR (\eg, the one used in KITTI~\cite{geiger2012we,geiger2013vision}) costs around \$$75,000$ (US dollars). As an alternative, researchers have also been investigating purely image-based 3D detection.
Existing algorithms are largely built upon 2D object detection~\cite{ren2015faster,he2017mask,lin2017feature}, imposing extra geometric constraints~\cite{chabot2017deep,chen2016monocular,mousavian20173d,xiang2017subcategory} to create 3D proposals. \cite{chen20153d,chen20183d,pham2017robust,xu2018multi} apply stereo-based depth estimation to obtain 3D coordinates of each pixel. These 3D coordinates are either entered as additional input channels into a 2D detection pipeline, or used to extract hand-crafted features. 
The recently proposed pseudo-LiDAR~\cite{pseudoLiDAR,E2EPL,you2019pseudo} combined stereo-based depth estimation with LiDAR-based detection, converting the depth map into a 3D point cloud and processing it exactly as LiDAR signal. The pseudo-LiDAR framework has largely improved image-based detection, yet a notable gap is still remained compared to LiDAR. In this work, we therefore focus on LiDAR-based object detectors.

\begin{table*}[]
\small
\caption{\small  Dataset overview. We focus on their properties related to frontal-view images, LiDAR, and 3D object detection. The dataset size refers to the number of synchronized $($image, LiDAR$)$ pairs. For \waymo and \nusc, we subsample the data. See text for details. 
} 
	\label{tab:datastat_s}
	\vskip -5pt
	\centering
	\tabcolsep 4pt
	\begin{tabular}{lccccccc}
		\hline
		Dataset & Size & LiDAR Type & Beam Angles & Object Types & Rainy Weather & Night Time \\ 
		\hline
		\kitti~\cite{geiger2013vision,geiger2012we} & $14,999$ & $1 \times 64\text{-beam}$ & $ [-24^\circ, 4^\circ]$ & 8 & No & No \\ 
		\argo~\cite{argoverse} & $22,305$ & $2 \times 32\text{-beam}$ & $[-26^\circ, 25^\circ]$ & 17 & No & Yes \\ 
		\nusc~\cite{nuscenes2019} & $34,149$ & $1 \times 32\text{-beam}$ & $[-16^\circ, 11^\circ]$ & 23 & Yes & Yes \\ 
		\lyft~\cite{lyft2019} & $18,634$ & $1\times40 \text{ or } 64 + 2 \times 40\text{-beam}$ & $[-29^\circ , 5^\circ]$ & 9 & No & No \\ 
		\waymo~\cite{waymo_open_dataset} & $192,484$ & $1 \times 64 + 4 \times 200\text{-beam}$ & $[-18^\circ, 2^\circ]$ & 4 & Yes & Yes \\
		\hline
	\end{tabular}
	\vskip -10pt
\end{table*}

\noindent\textbf{Domain adaptation.} (Unsupervised) domain adaptation has also been studied in autonomous driving scenes, but mainly for the tasks of 2D semantic segmentation~\cite{chen2018road,hoffman2017cycada,huang2018domain,luo2019taking,saito2018maximum,saleh2018effective,sankaranarayanan2018learning,tsai2018learning,zhang2017curriculum,zou2018unsupervised} and 2D object detection~\cite{cai2019exploring,chen2018domain,he2019multi,hsu2020progressive,khodabandeh2019robust,kim2019diversify,rodriguez2019domain,saito2019strong,wang2019few,zhuang2020ifan,zhu2019adapting}. The common setting is to adapt a model trained from one labeled source domain (\eg, synthetic images) to an unlabeled target domain (e.g., real images). The domain difference is mostly from the input signal (\eg, image styles), and many algorithms have built upon adversarial feature matching and style transfer~\cite{ganin2016domain,hoffman2017cycada,zhu2017unpaired} to minimize the domain gap in the input or feature space. Our work contrasts these methods by studying 3D object detection. We found that, the output space (\eg, car sizes) can also contribute to the domain gap; properly leveraging the statistics of the target domain can largely improve the model's generalization ability.


\section{Datasets}
\label{sec:data}

We review \kitti~\cite{geiger2013vision,geiger2012we} and introduce the other four datasets used in our experiments: \argo~\cite{argoverse}, \lyft~\cite{lyft2019}, \nusc~\cite{nuscenes2019}, and \waymo~\cite{waymo_open_dataset}. 
We focus on data related to 3D object detection.
All the datasets provide ground-truth 3D bounding box labels for several kinds of objects. We summarize the five datasets in detail in \autoref{tab:datastat_s}. 

\noindent\textbf{\kitti.}
The \kitti object detection benchmark~\cite{geiger2013vision, geiger2012we} contains $7,481$ (left) images for training and $7,518$ images for testing.
The training set is further separated into $3,712$ training and $3,769$ validation images as suggested by \cite{chen20153d}. All the scenes are pictured around Karlsruhe, Germany in clear weather and day time.
For each (left) image, KITTI provides its corresponding 64-beam Velodyne LiDAR point cloud and the right stereo image.

\noindent\textbf{\argo.}
The \argo dataset~\cite{argoverse} is collected around Miami and Pittsburgh, USA in multiple weathers and during different times of a day. 
It provides images from stereo cameras and another seven cameras that cover $360^\circ$ information.
It also provides $64$-beam LiDAR point clouds captured by two $32$-beam Velodyne LiDAR sensors stacked vertically. 
We extracted synchronized frontal-view images and corresponding point clouds from the original \argo dataset, with a timestamp tolerance of 51 ms between LiDAR sweeps and images. 
The resulting dataset we use contains $13,122$ images for training, $5,015$ images for validation, $4,168$ images for testing.

\noindent\textbf{\nusc.}
The \nusc dataset~\cite{nuscenes2019}
contains $28,130$ training and $6,019$ validation images. We treat the validation images as test images, and re-split and subsample the $28,130$ training images into $11,040$ training and $3,026$ validation images.
The scenes are pictured around Boston, USA and Singapore in multiple weathers and during different times of a day. For each image, \nusc provides the point cloud captured by a $32$-beam roof LiDAR. It also provides images from another five cameras that cover $360^\circ$ information. 

\noindent\textbf{\lyft.}
The \lyft Level 5 dataset~\cite{lyft2019} contains $18,634$ frontal-view images and we separate them into $12,599$ images for training, $3,024$ images for validation, $3,011$ images for testing. 
The scenes are pictured around Palo Auto, USA in clear weathers and during day time. 
For each image, \lyft provides the point cloud captured by a $40$ (or $64$)-beam roof LiDAR and two $40$-beam bumper LiDAR sensors. 
It also provides images from another five cameras that cover $360^\circ$ information and one long-focal-length camera.

\noindent\textbf{\waymo.}
The \waymo dataset~\cite{waymo_open_dataset} contains $122,000$ training, $30,407$ validation, and $40,077$ test images and we sub-sample them into $12,000$ , $3,000$, and $3,000$, respectively. 
The scenes are pictured at Phoenix, Mountain View, and San Francisco in multiple weathers and at multiple times of a day. 
For each image, \waymo provides the combined point cloud captured by five LiDAR sensors (one on the roof). 
It also provides images from another four cameras.

\noindent\textbf{Data format.}
A non-negligible difficulty in conducting cross-dataset analysis lies in the differences of data formats. 
\emph{Considering that most existing algorithms are developed using the \kitti format, we transfer all the other four datasets into its format.} 
See the Supplementary Material for details.

\section{Experiments and Analysis}
\label{sec:exp}

\subsection{Setup}
\label{ssec:exp_setup}
\noindent\textbf{3D object detection algorithms.}
We apply two LiDAR-based models \PRCNN~\cite{shi2019pointrcnn} and \PIXOR~\cite{yang2018pixor} to detect objects in 3D by outputting the surrounding 3D bounding boxes.
\PIXOR represents LiDAR point clouds by 3D tensors after voxelization, while \PRCNN applies PointNet++~\cite{qi2017pointnet++} to extract point-wise features. Both methods do not rely on images.
We train both models on the five 3D object detection datasets. \PRCNN has two sub-networks, the region proposal network (RPN) and region-CNN (RCNN), that are trained separately. The RPN is trained first, for $200$ epochs with batch size $16$ and learning rate $0.02$. The RCNN is trained for $70$ epochs with batch size $4$ and learning rate $0.02$. We use online ground truth boxes augmentation, which copies object boxes and inside points from one scene to the same locations in another scene. For \PIXOR, we train it with batch size $4$ and initial learning rate $5\times10^{-5}$, which will be decreased 10 times on the 50th and 80th epoch. We do randomly horizontal flip and rotate during training. 

\noindent\textbf{Metric.}
We follow \kitti to evaluate object detection in 3D and the bird's-eye view (BEV). 
We focus on the \emph{Car} category, which has been the main focus in existing works. 
We report average precision (AP) with the IoU thresholds at 0.7: a car is correctly detected if the intersection over union (IoU) with the predicted 3D box is larger than $0.7$.
We denote AP for the 3D and BEV tasks by \AP and \APBEV.

\kitti evaluates three cases: \emph{Easy}, \emph{Moderate}, and \emph{Hard}. 
Specifically, it labels each ground truth box with four levels (0 to 3) of occlusion / truncation. 
The \emph{Easy} case contains level-0 cars whose bounding box heights in 2D are larger than $40$ pixels; 
the \emph{Moderate} case contains level-\{0, 1\} cars whose bounding box heights in 2D are larger than $25$ pixels; 
the \emph{Hard} case contains level-\{0, 1, 2\} cars whose bounding box heights in 2D are larger than $25$ pixels.
The heights are meant to separate cars by their depths with respect to the observing car. 
Nevertheless, since different datasets have different image resolutions, such criteria might not be aligned across datasets. 
We thus replace the constraints of ``larger than $40, 25$ pixels'' by ``within $30, 70$ meters''. 
We further evaluate cars of level-\{0, 1, 2\} within three depth ranges: $0-30$, $30-50$, and $50-70$ meters, following~\cite{yang2018pixor}.

We mainly report and discuss results of \PRCNN on the \emph{validation} set  in the main paper. We report results of \PIXOR in the Supplementary Material.

\begin{table*}[h]
\small	
\caption{\small\textbf{3D object detection across multiple datasets} (evaluated on the validation sets). We report average precision (AP) of the \emph{Car} category in bird's-eye view (AP$_{\text{BEV}}$) and 3D (AP$_{\text{3D}}$) at $\text{IoU}=0.7$, using the \PRCNN detector~\cite{shi2019pointrcnn}. We report results at different difficulties (following the \kitti benchmark, but we replace the $40$, $25$, $25$ pixel thresholds on 2D bounding boxes with $30$, $70$, $70$ meters on object depths, for \emph{Easy}, \emph{Moderate}, and \emph{Hard} cases, respectively) and different depth ranges (using the same truncation and occlusion thresholds as \kitti \emph{Hard} case). The results show a significant performance drop in cross-dataset inference. We indicate the best generalization results per column and per setting
by red fonts and the worst by blue fonts. We indicate in-domain results by bold fonts.}
\vskip-5pt
\label{tab:cross_inference}
\centering
\begin{tabular}{c|l|ccccc}
\hline
 Setting & Source$\backslash$Target & \kitti & \argo & \nusc & \lyft & \waymo \\
\hline
\multirow{5}{*}{Easy} & \kitti & \color{black}{\textbf{88.0}} \color{black}{/} \color{black}{\textbf{82.5}} & \color{blue}{55.8} \color{black}{/} \color{black}{27.7} & \color{blue}{47.4} \color{black}{/} \color{blue}{13.3} & \color{blue}{81.7} \color{black}{/} \color{black}{51.8} & \color{blue}{45.2} \color{black}{/} \color{blue}{11.9} \\
 & \argo & \color{black}{69.5} \color{black}{/} \color{black}{33.9} & \color{black}{\textbf{79.2}} \color{black}{/} \color{black}{\textbf{57.8}} & \color{black}{52.5} \color{black}{/} \color{black}{21.8} & \color{black}{86.9} \color{black}{/} \color{black}{67.4} & \color{black}{83.8} \color{black}{/} \color{black}{40.2} \\
 & \nusc & \color{blue}{49.7} \color{black}{/} \color{black}{13.4} & \color{black}{73.2} \color{black}{/} \color{blue}{21.8} & \color{black}{\textbf{73.4}} \color{black}{/} \color{black}{\textbf{38.1}} & \color{red}{89.0} \color{black}{/} \color{blue}{38.2} & \color{black}{78.8} \color{black}{/} \color{black}{36.7} \\
 & \lyft & \color{red}{74.3} \color{black}{/} \color{red}{39.4} & \color{red}{77.1} \color{black}{/} \color{red}{45.8} & \color{red}{63.5} \color{black}{/} \color{red}{23.9} & \color{black}{\textbf{90.2}} \color{black}{/} \color{black}{\textbf{87.3}} & \color{red}{87.0} \color{black}{/} \color{red}{64.7} \\
 & \waymo & \color{black}{51.9} \color{black}{/} \color{blue}{13.1} & \color{black}{76.4} \color{black}{/} \color{black}{42.6} & \color{black}{55.5} \color{black}{/} \color{black}{21.6} & \color{black}{87.9} \color{black}{/} \color{red}{74.5} & \color{black}{\textbf{90.1}} \color{black}{/} \color{black}{\textbf{85.3}} \\
\hline\multirow{5}{*}{Moderate} & \kitti & \color{black}{\textbf{80.6}} \color{black}{/} \color{black}{\textbf{68.9}} & \color{blue}{44.9} \color{black}{/} \color{black}{22.3} & \color{blue}{26.2} \color{black}{/} \color{blue}{8.3} & \color{blue}{61.8} \color{black}{/} \color{black}{33.7} & \color{blue}{43.9} \color{black}{/} \color{blue}{12.3} \\
 & \argo & \color{black}{56.6} \color{black}{/} \color{black}{31.4} & \color{black}{\textbf{69.9}} \color{black}{/} \color{black}{\textbf{44.2}} & \color{black}{27.6} \color{black}{/} \color{black}{11.8} & \color{black}{66.6} \color{black}{/} \color{black}{42.1} & \color{black}{72.3} \color{black}{/} \color{black}{35.1} \\
 & \nusc & \color{blue}{39.8} \color{black}{/} \color{blue}{10.7} & \color{black}{56.6} \color{black}{/} \color{blue}{17.1} & \color{black}{\textbf{40.7}} \color{black}{/} \color{black}{\textbf{21.2}} & \color{black}{71.4} \color{black}{/} \color{blue}{25.0} & \color{black}{68.2} \color{black}{/} \color{black}{30.8} \\
 & \lyft & \color{red}{61.1} \color{black}{/} \color{red}{34.3} & \color{black}{62.5} \color{black}{/} \color{red}{35.3} & \color{red}{33.6} \color{black}{/} \color{black}{12.3} & \color{black}{\textbf{83.7}} \color{black}{/} \color{black}{\textbf{65.5}} & \color{red}{77.6} \color{black}{/} \color{red}{53.2} \\
 & \waymo & \color{black}{45.8} \color{black}{/} \color{black}{13.2} & \color{red}{64.4} \color{black}{/} \color{black}{29.8} & \color{black}{28.9} \color{black}{/} \color{red}{13.7} & \color{red}{74.2} \color{black}{/} \color{red}{53.8} & \color{black}{\textbf{85.9}} \color{black}{/} \color{black}{\textbf{67.9}} \\
\hline\multirow{5}{*}{Hard} & \kitti & \color{black}{\textbf{81.9}} \color{black}{/} \color{black}{\textbf{66.7}} & \color{blue}{42.5} \color{black}{/} \color{black}{22.2} & \color{blue}{24.9} \color{black}{/} \color{blue}{8.8} & \color{blue}{57.4} \color{black}{/} \color{black}{34.2} & \color{blue}{41.5} \color{black}{/} \color{blue}{12.6} \\
 & \argo & \color{black}{58.5} \color{black}{/} \color{black}{33.3} & \color{black}{\textbf{69.9}} \color{black}{/} \color{black}{\textbf{42.8}} & \color{black}{26.8} \color{black}{/} \color{red}{14.5} & \color{black}{64.4} \color{black}{/} \color{black}{42.7} & \color{black}{68.5} \color{black}{/} \color{black}{36.8} \\
 & \nusc & \color{blue}{39.6} \color{black}{/} \color{blue}{10.1} & \color{black}{53.3} \color{black}{/} \color{blue}{16.7} & \color{black}{\textbf{40.2}} \color{black}{/} \color{black}{\textbf{20.5}} & \color{black}{67.7} \color{black}{/} \color{blue}{25.7} & \color{black}{66.9} \color{black}{/} \color{black}{29.0} \\
 & \lyft & \color{red}{60.7} \color{black}{/} \color{red}{33.9} & \color{red}{62.9} \color{black}{/} \color{red}{35.9} & \color{red}{30.6} \color{black}{/} \color{black}{11.7} & \color{black}{\textbf{79.3}} \color{black}{/} \color{black}{\textbf{65.5}} & \color{red}{77.0} \color{black}{/} \color{red}{53.9} \\
 & \waymo & \color{black}{46.3} \color{black}{/} \color{black}{12.6} & \color{black}{61.6} \color{black}{/} \color{black}{29.0} & \color{black}{28.4} \color{black}{/} \color{black}{14.1} & \color{red}{74.1} \color{black}{/} \color{red}{54.5} & \color{black}{\textbf{80.4}} \color{black}{/} \color{black}{\textbf{67.7}} \\
\hline\hline\multirow{5}{*}{0-30m} & \kitti & \color{black}{\textbf{88.8}} \color{black}{/} \color{black}{\textbf{84.9}} & \color{blue}{58.4} \color{black}{/} \color{black}{34.7} & \color{blue}{47.9} \color{black}{/} \color{blue}{14.9} & \color{blue}{77.8} \color{black}{/} \color{black}{54.2} & \color{blue}{48.0} \color{black}{/} \color{blue}{14.0} \\
 & \argo & \color{black}{74.2} \color{black}{/} \color{red}{46.8} & \color{black}{\textbf{83.3}} \color{black}{/} \color{black}{\textbf{63.3}} & \color{black}{55.3} \color{black}{/} \color{red}{26.9} & \color{black}{87.7} \color{black}{/} \color{black}{69.5} & \color{black}{85.7} \color{black}{/} \color{black}{44.4} \\
 & \nusc & \color{blue}{50.7} \color{black}{/} \color{blue}{13.9} & \color{black}{73.7} \color{black}{/} \color{blue}{26.0} & \color{black}{\textbf{73.2}} \color{black}{/} \color{black}{\textbf{42.8}} & \color{red}{89.1} \color{black}{/} \color{blue}{43.8} & \color{black}{79.8} \color{black}{/} \color{black}{43.4} \\
 & \lyft & \color{red}{75.1} \color{black}{/} \color{black}{45.2} & \color{red}{81.0} \color{black}{/} \color{red}{54.0} & \color{red}{61.6} \color{black}{/} \color{black}{25.4} & \color{black}{\textbf{90.4}} \color{black}{/} \color{black}{\textbf{88.5}} & \color{red}{88.6} \color{black}{/} \color{red}{70.9} \\
 & \waymo & \color{black}{56.8} \color{black}{/} \color{black}{15.0} & \color{black}{80.6} \color{black}{/} \color{black}{48.1} & \color{black}{57.8} \color{black}{/} \color{black}{24.0} & \color{black}{88.4} \color{black}{/} \color{red}{76.2} & \color{black}{\textbf{90.4}} \color{black}{/} \color{black}{\textbf{87.2}} \\
\hline\multirow{5}{*}{30m-50m} & \kitti & \color{black}{\textbf{70.2}} \color{black}{/} \color{black}{\textbf{51.4}} & \color{black}{46.5} \color{black}{/} \color{black}{19.0} & \color{black}{9.8} \color{black}{/} \color{blue}{4.5} & \color{blue}{60.1} \color{black}{/} \color{black}{34.5} & \color{blue}{50.5} \color{black}{/} \color{blue}{21.4} \\
 & \argo & \color{black}{33.9} \color{black}{/} \color{black}{11.8} & \color{black}{\textbf{72.2}} \color{black}{/} \color{black}{\textbf{39.5}} & \color{blue}{9.5} \color{black}{/} \color{black}{9.1} & \color{black}{65.9} \color{black}{/} \color{black}{39.1} & \color{black}{75.9} \color{black}{/} \color{black}{42.1} \\
 & \nusc & \color{blue}{24.1} \color{black}{/} \color{blue}{3.8} & \color{blue}{46.3} \color{black}{/} \color{blue}{6.4} & \color{black}{\textbf{17.1}} \color{black}{/} \color{black}{\textbf{4.1}} & \color{black}{70.1} \color{black}{/} \color{blue}{18.9} & \color{black}{69.4} \color{black}{/} \color{black}{29.2} \\
 & \lyft & \color{red}{39.3} \color{black}{/} \color{red}{16.6} & \color{red}{59.2} \color{black}{/} \color{red}{21.8} & \color{red}{11.2} \color{black}{/} \color{black}{9.1} & \color{black}{\textbf{83.8}} \color{black}{/} \color{black}{\textbf{62.7}} & \color{red}{79.4} \color{black}{/} \color{red}{55.5} \\
 & \waymo & \color{black}{31.7} \color{black}{/} \color{black}{9.3} & \color{black}{58.0} \color{black}{/} \color{black}{18.8} & \color{black}{9.9} \color{black}{/} \color{red}{9.1} & \color{red}{74.5} \color{black}{/} \color{red}{51.4} & \color{black}{\textbf{87.5}} \color{black}{/} \color{black}{\textbf{68.8}} \\
\hline\multirow{5}{*}{50m-70m} & \kitti & \color{black}{\textbf{28.8}} \color{black}{/} \color{black}{\textbf{12.0}} & \color{blue}{9.2} \color{black}{/} \color{black}{3.0} & \color{black}{1.1} \color{black}{/} \color{black}{0.0} & \color{blue}{33.2} \color{black}{/} \color{black}{9.6} & \color{blue}{27.1} \color{black}{/} \color{blue}{12.0} \\
 & \argo & \color{black}{10.9} \color{black}{/} \color{blue}{1.3} & \color{black}{\textbf{29.9}} \color{black}{/} \color{black}{\textbf{6.9}} & \color{blue}{0.5} \color{black}{/} \color{blue}{0.0} & \color{black}{35.1} \color{black}{/} \color{black}{14.5} & \color{black}{46.2} \color{black}{/} \color{black}{23.0} \\
 & \nusc & \color{black}{6.5} \color{black}{/} \color{black}{1.5} & \color{black}{15.2} \color{black}{/} \color{blue}{2.3} & \color{black}{\textbf{9.1}} \color{black}{/} \color{black}{\textbf{9.1}} & \color{black}{41.8} \color{black}{/} \color{blue}{5.3} & \color{black}{37.9} \color{black}{/} \color{black}{15.2} \\
 & \lyft & \color{red}{13.6} \color{black}{/} \color{red}{4.6} & \color{black}{23.1} \color{black}{/} \color{black}{3.9} & \color{red}{1.1} \color{black}{/} \color{red}{0.0} & \color{black}{\textbf{62.7}} \color{black}{/} \color{black}{\textbf{33.1}} & \color{red}{54.6} \color{black}{/} \color{red}{27.5} \\
 & \waymo & \color{blue}{5.6} \color{black}{/} \color{black}{1.8} & \color{red}{26.9} \color{black}{/} \color{red}{5.6} & \color{black}{0.9} \color{black}{/} \color{black}{0.0} & \color{red}{50.8} \color{black}{/} \color{red}{21.3} & \color{black}{\textbf{63.5}} \color{black}{/} \color{black}{\textbf{41.1}} \\
\hline
\end{tabular}
\vskip-10pt
\end{table*}

\subsection{Results within each dataset}
\label{ssec:within}
We first evaluate if existing 3D object detection models that have shown promising results on the \kitti benchmark can also be learned and perform well on newly released datasets.
We summarize the results in \autoref{tab:cross_inference}: the rows are the source domains that a detector is trained on, and the columns are the target domains the detector is being tested on. The \textbf{bold} font indicates the within domain performance (\ie, training and testing using the same dataset).

We see that \PRCNN works fairly well on the \kitti, \lyft, and \waymo datasets, for all the easy, moderate, and hard cases. The results get slightly worse on \argo, and then \nusc. We hypothesize that this may result from the relatively poor LiDAR input: \nusc has only $32$ beams; while \argo has $64$ beams, every two of them are very close due to the configurations that the signal is captured by two stacked LiDAR sensors.

We further analyze at different ranges in \autoref{tab:cross_inference} (bottom). We see a drastic drop on \argo and \nusc for the far-away ranges, which supports our hypothesis: with fewer beams, the far-away objects can only be rendered by very sparse LiDAR points and thus are hard to detect. We also see poor accuracies at $50-70$ meters on \kitti, which may result from very few labeled training instances there. 

Overall, both 3D object detection algorithms work fairly well when  being trained and tested using the same dataset, as long as the input sensor signal is of high quality and the labeled instances are sufficient. 

\subsection{Results across datasets}
\label{exp: across}
We further experiment with generalizing a trained detector across datasets. 
We indicate the best result per column and per setting by red fonts and the worst by blue fonts.

We see a clear trend of performance drop. For instance, the \PRCNN model trained on \kitti achieves only $45.2\%$ \APBEV (Moderate) on \waymo, lower than the model trained on \waymo by over $40\%$. The gap becomes even larger in \AP: the same \kitti model attains only $11.9\%$ \AP, while the \waymo model attains $85.3\%$.
We hypothesize that the car height is hard to get right. 
In terms of the target (test) domain, \lyft and \waymo suffer the least drop if the detector is trained from the other datasets, followed by \argo. \kitti and \nusc suffer the most drop, which might result from their different geo-locations (one is from Germany and the other contains data from Singapore). The \nusc dataset might also suffer from its relatively fewer beams in the input and other models may therefore not be able to 
apply.
By considering different ranges, we also find that the deeper the range is, the bigger the drop is.

In terms of the source (training) domain, we see that the detector trained on \kitti seems to be the worst to transfer to others. In every $5\times1$ block that is evaluated on a single dataset in a single setting, the \kitti model is mostly outperformed by others. Surprisingly, \nusc model can perform fairly well when being tested on the other datasets: the results are even higher than on its own. We thus have two arguments: The quality of sensors is more important in testing than in training; \kitti data (\eg, car styles, time, and weather) might be too limited or different from others and therefore cannot transfer well to others. In the following subsections, we provide detailed analysis.

\begin{figure}[htbp!]
	\centering
	\small
	\centerline{\includegraphics[width=0.9\linewidth]{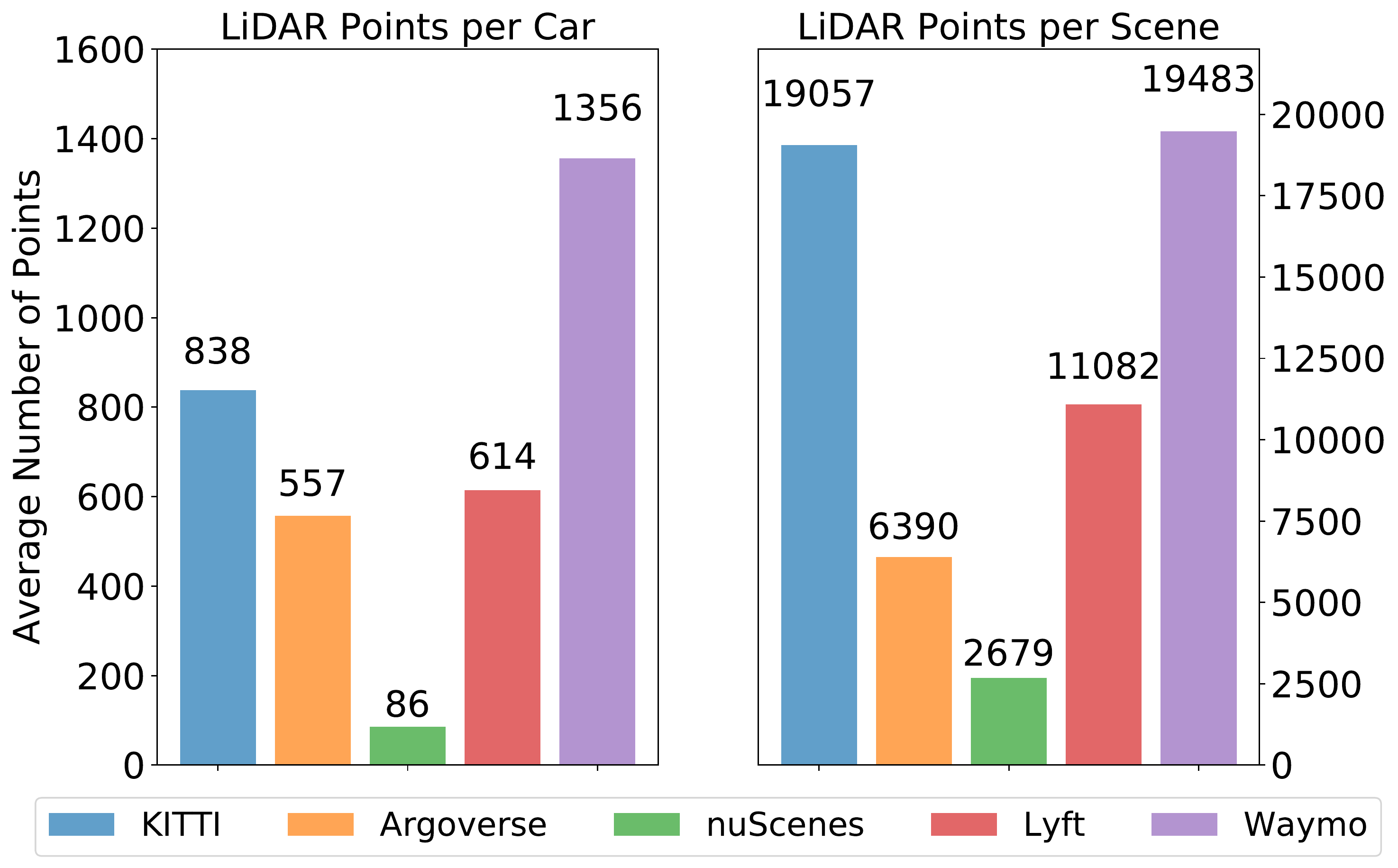}}
	\vskip -5pt
	\caption{\small  The average numbers of 3D points per car (left) and per scene (right). We only include points within the frontal-view camera view and cars whose depths are within $70$ meters.}
	\label{fig:2}
	\vskip -15pt
\end{figure}

\begin{figure*}[t!]
	\centering
	\includegraphics[width=0.31\linewidth]{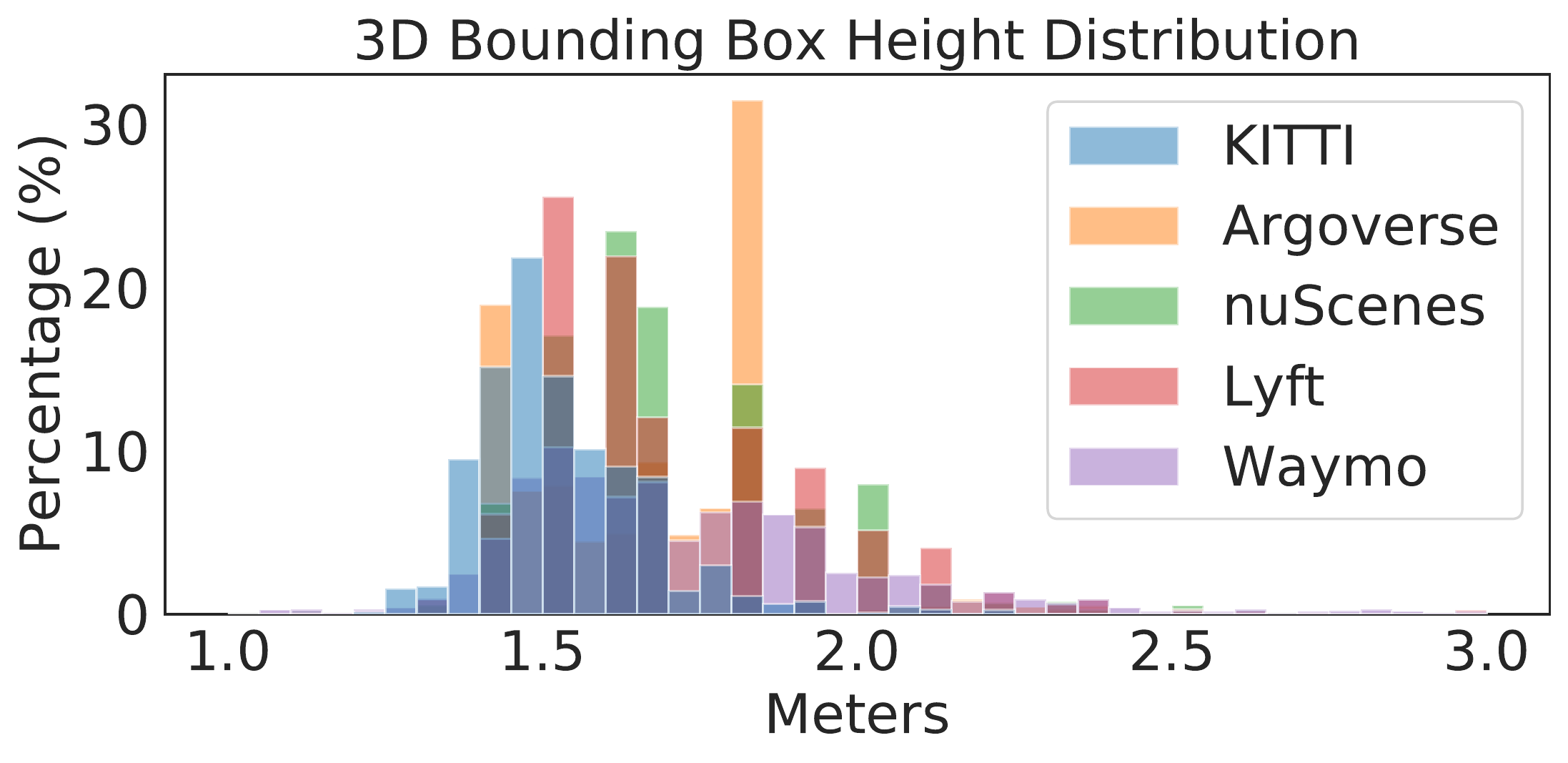}\hspace{5pt}
	\includegraphics[width=0.31\linewidth]{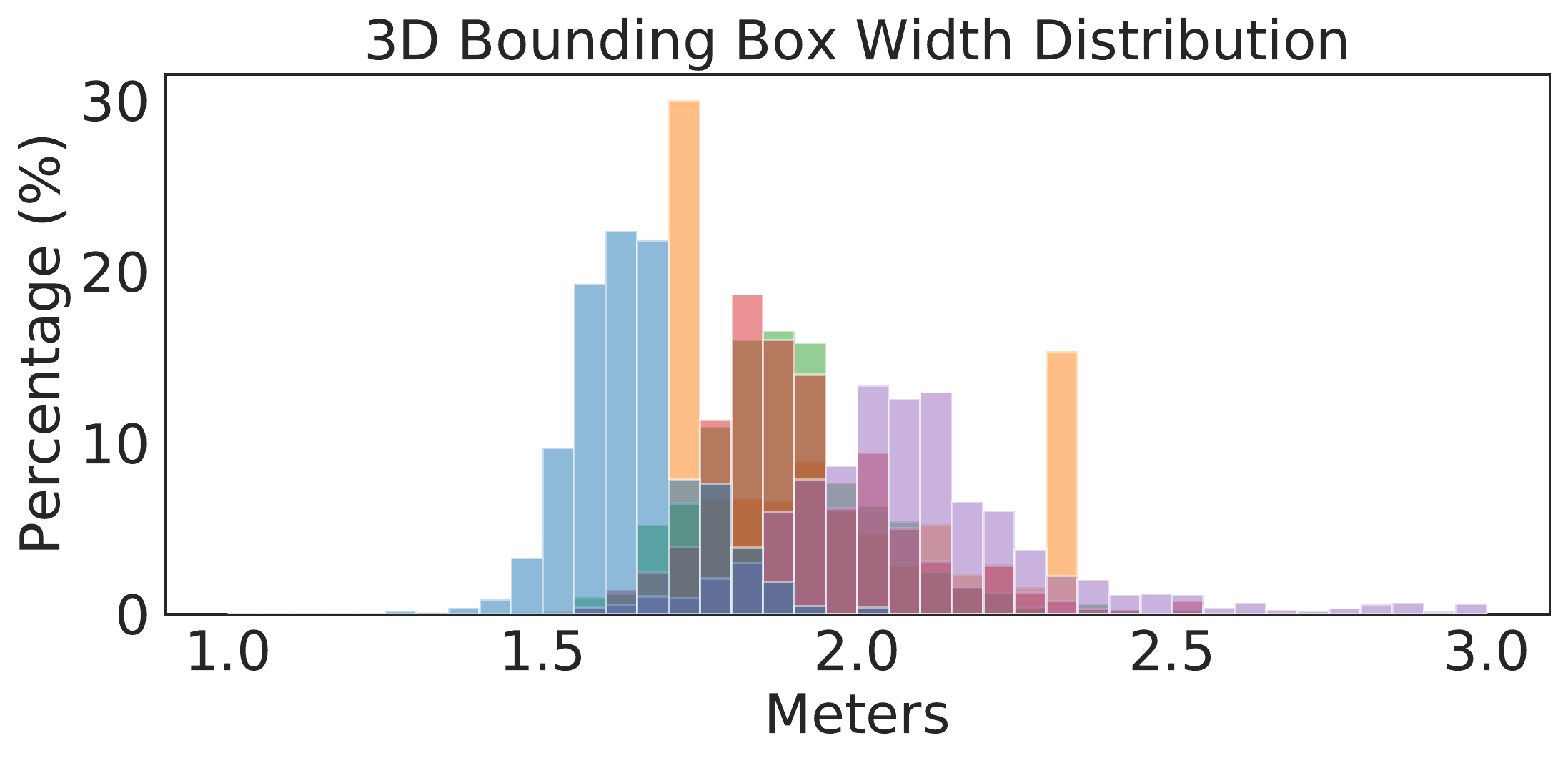}\hspace{5pt}
	\includegraphics[width=0.31\linewidth]{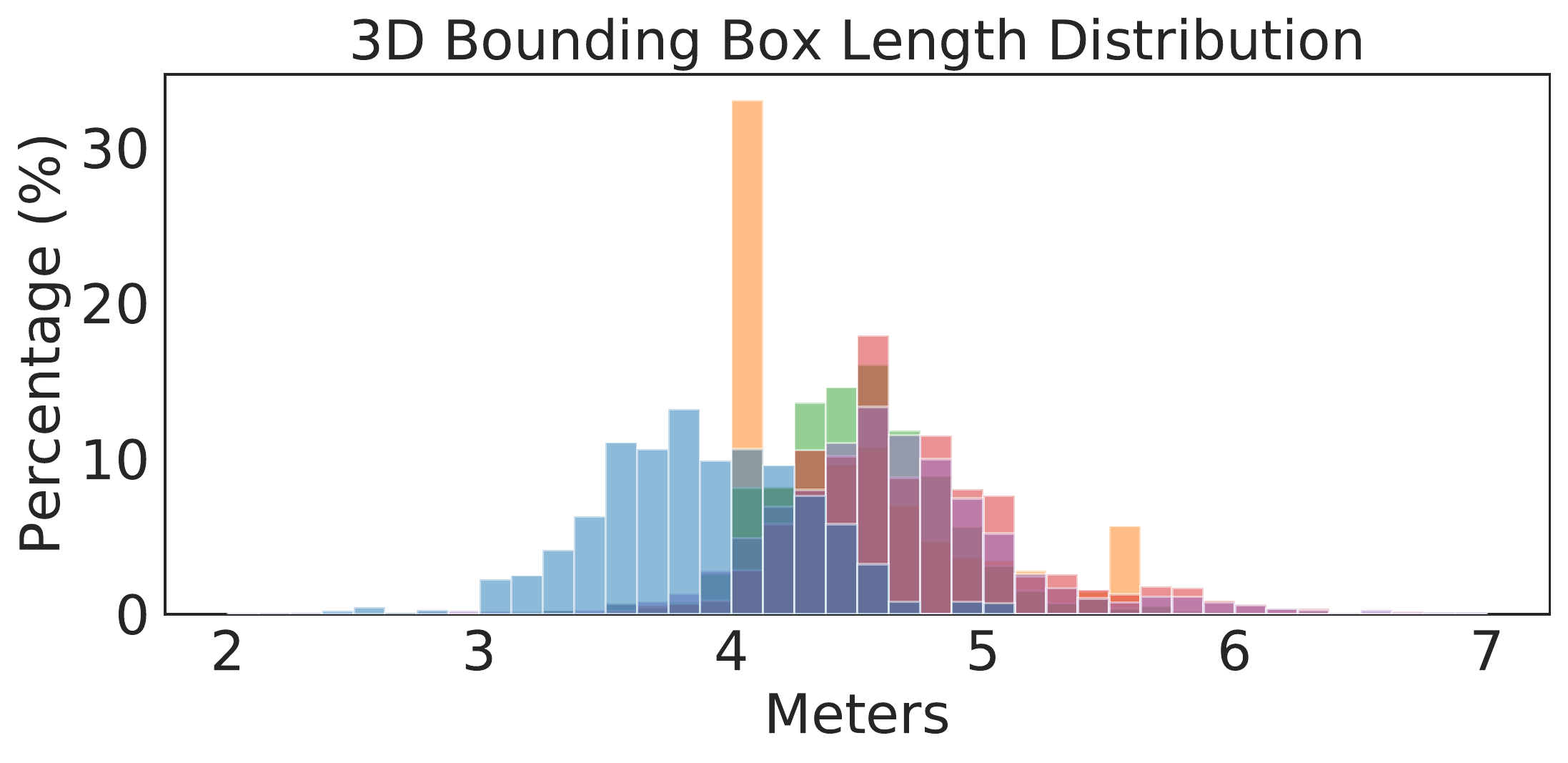}
	\vskip -10pt
	\caption{\small Car size statistics of different datasets.}
	\label{fig:stat}
\end{figure*}

\begin{figure*}[ht!]
	\centering
	\includegraphics[width=0.31\linewidth]{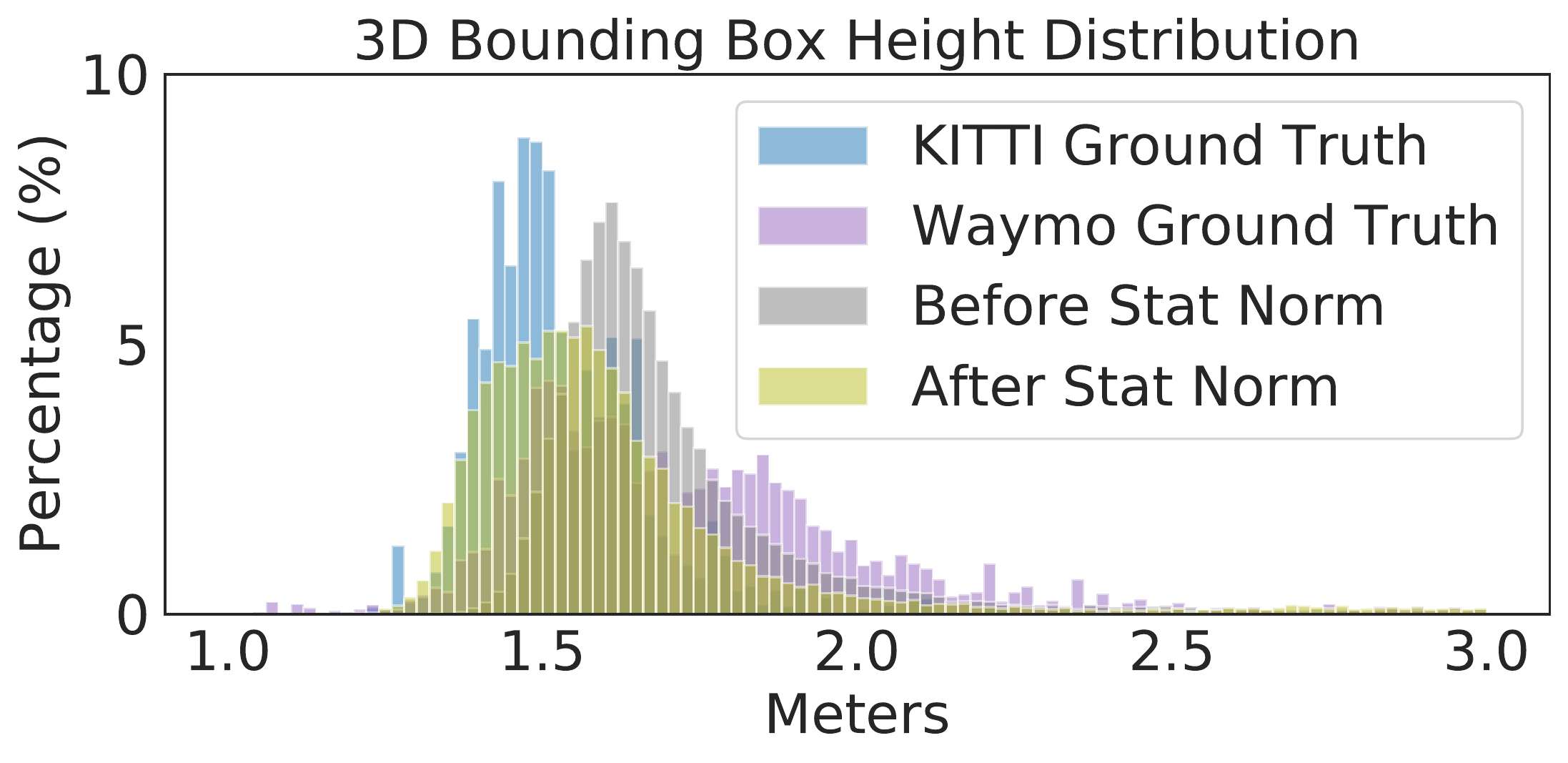}\hspace{5pt}
	\includegraphics[width=0.31\linewidth]{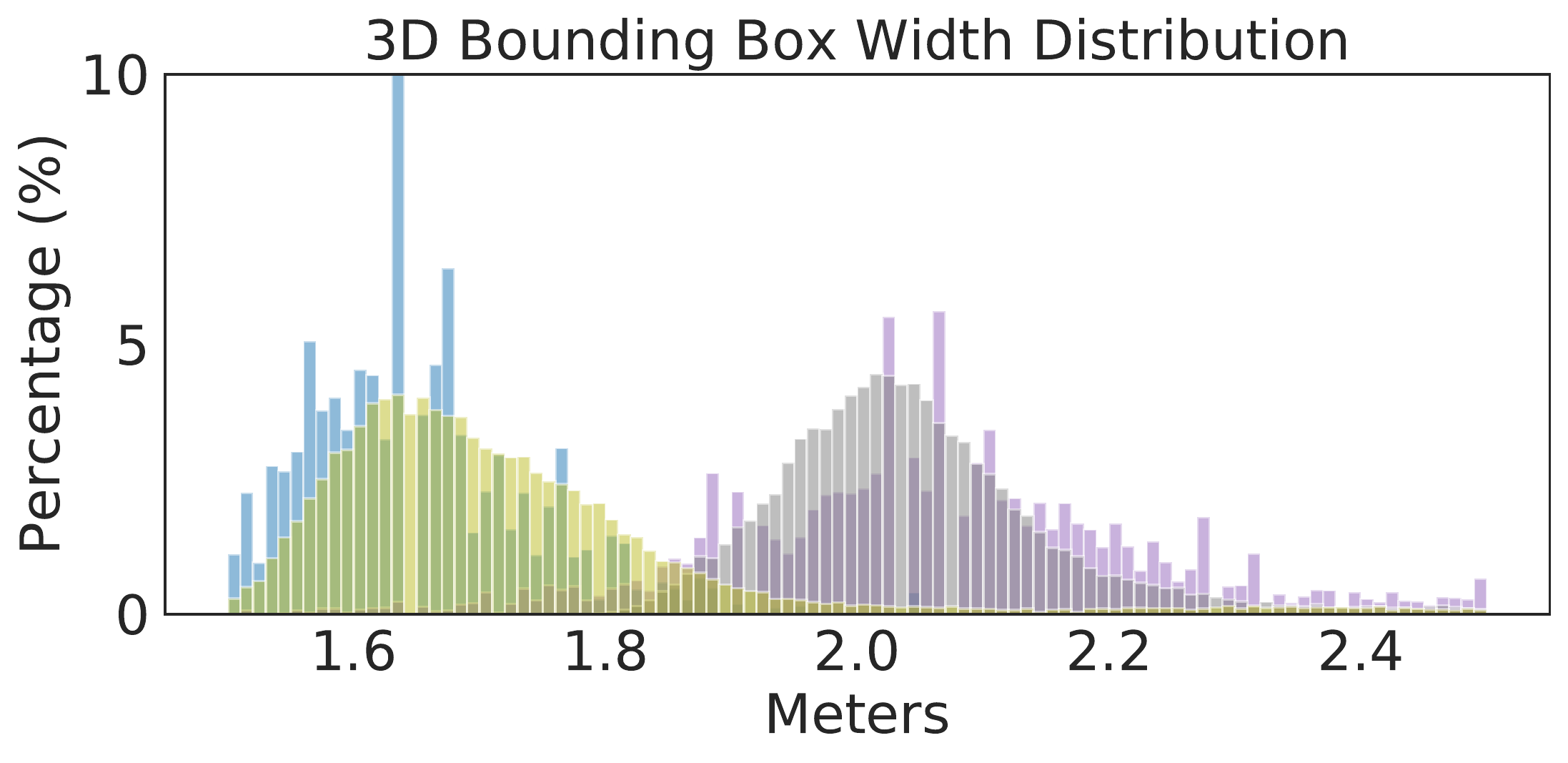}\hspace{5pt}
	\includegraphics[width=0.31\linewidth]{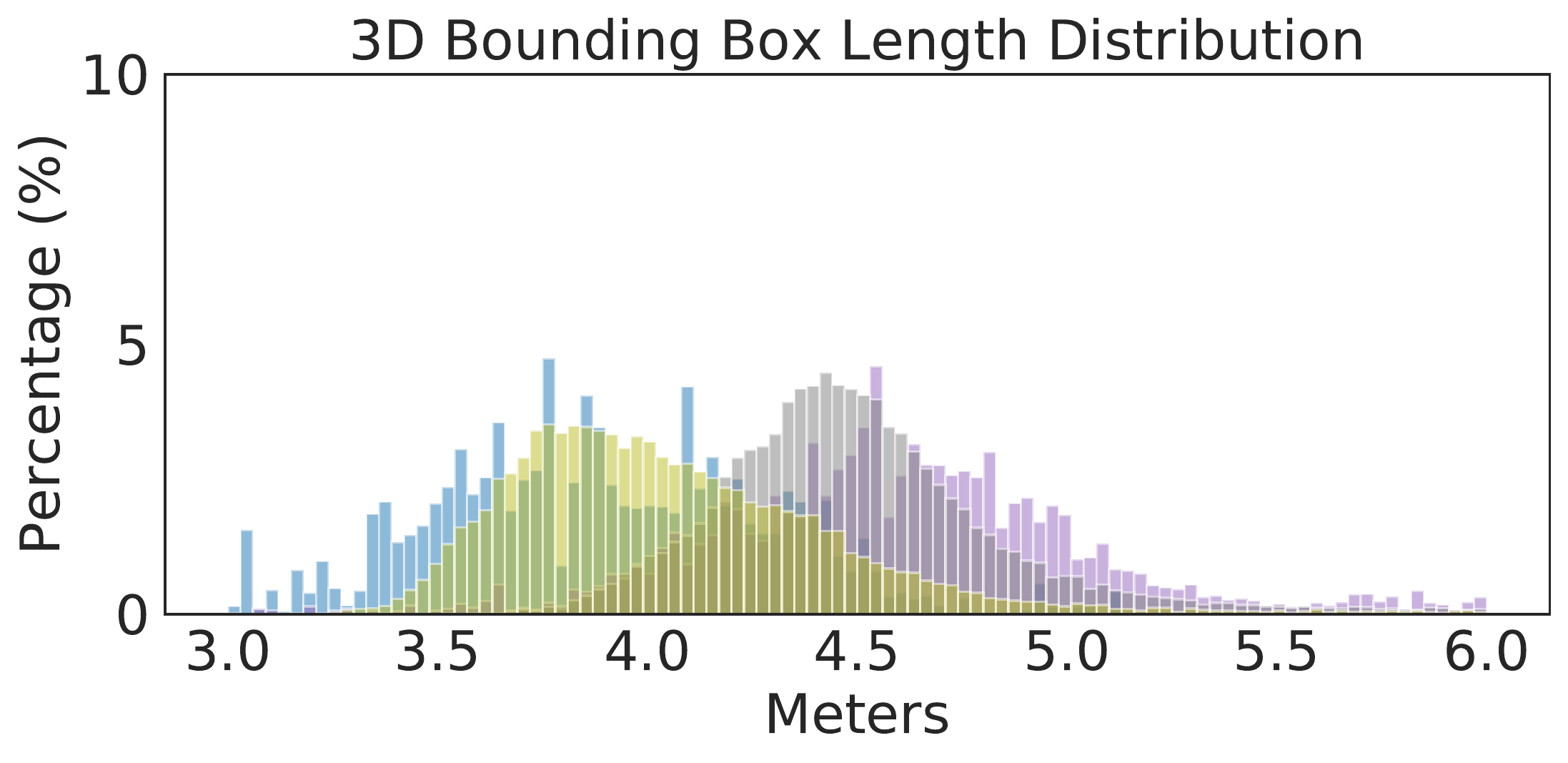}
	\vskip -10pt
	\caption{\small Sizes of detected bounding boxes before and after our Statistical Normalization (Stat Norm). The detector is trained on \waymo (w/o or w/ Stat Norm) and tested on \kitti. We also show the distribution of ground-truth box sizes in both datasets.}
	\vskip -10pt
	\label{fig:adapt}
\end{figure*}

\begin{figure}[]
    \centering
    \includegraphics[width=0.78\linewidth]{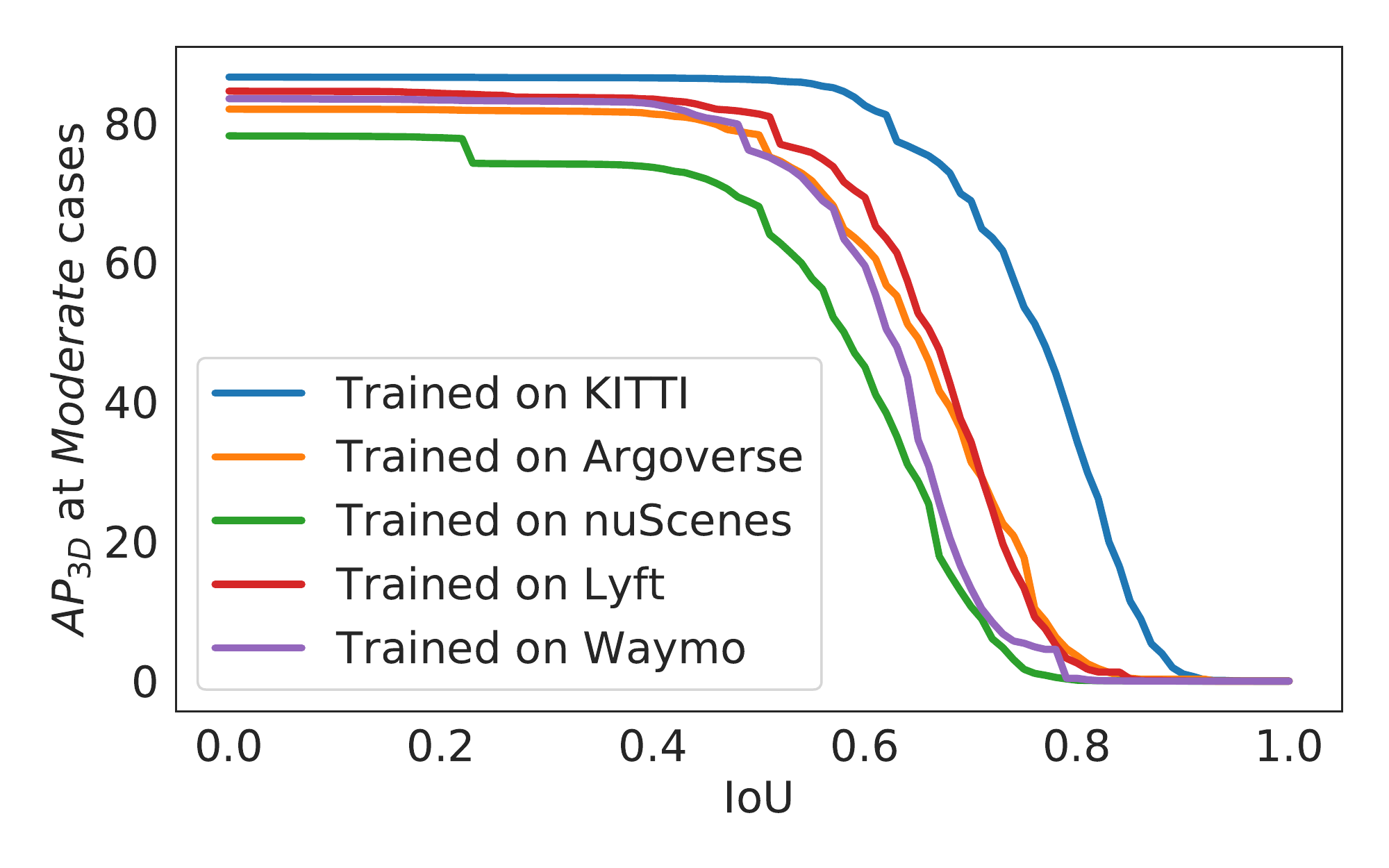}
    \vskip -15pt
    \caption{\small Car detection accuracy (\AP at \emph{Moderate} cases) on \kitti, using \PRCNN models trained on different datasets. We vary the IoU threshold from $0.0$ to $1.0$ (x-axis).
    The curves indicate that models trained on different datasets have similar detection abilities (converge at low IoU) but they differ in localization (diverge at high IoU).}
    \vskip -10pt
    \label{fig:overview}
\end{figure}

\subsection{Analysis of domain idiosyncrasies}
\label{ssec:analysis}

\autoref{tab:cross_inference} and \autoref{exp: across} reveal drastic accuracy drops in generalizing 3D object detectors across datasets (domains). We hypothesize that there exist significant idiosyncrasies in each dataset.
In particular, \autoref{fig:1} shows that the images and point clouds are quite different across datasets. One one hand, different datasets are collected by cars of different sensor configurations. For example, \nusc uses a single $32$-beam LiDAR; the point clouds are thus sparser than the other datasets. On the other hand, these datasets are collected at different locations; the environments and the foreground object styles may also be different. 

To provide a better understanding, we compute the average number of LiDAR points per scene and per car (using the ground-truth 3D bounding box) in~\autoref{fig:2}. We see a large difference: \waymo has ten times of points per car than \nusc\footnote{We note that \PRCNN applies point re-sampling so that every scene (in RPN) and object proposal (in RCNN) will have the same numbers of input points while \PIXOR applies voxelization. Both operations can reduce but cannot fully resolve point cloud differences across domains.}. We further analyze the size of bounding boxes per car.
\autoref{fig:stat} shows the histograms of each dataset. We again see mismatches between different datasets: \kitti seems to have the smallest box sizes while \waymo has the largest. We conduct an analysis and find that most of the bounding boxes tightly contain the points of cars inside. We, therefore, argue that this difference of box sizes is related to the car styles captured in different datasets.

\subsection{Analysis of  detector performance}
\label{ssec:detector_perf}
So what are the idiosyncrasies that account for the majority of performance gap?
There are two factors that can lead to an miss-detected car (\ie, IoU $< 0.7$): the car might be entirely missed by the detector, or it is detected but poorly localized. 
To identify the main factor, we lower down the IoU threshold using \kitti as the target domain (see~\autoref{fig:overview}). We observe an immediate increase in \AP, and the results become saturated when IoU is lower than $0.4$.  
Surprisingly, \PRCNN models trained from other datasets perform on a par with the model trained on \kitti. 
In other words, poor generalization resides primarily in localization.

We investigate one cause of mislocalization\footnote{Mislocalization can result from wrong box centers, rotations, or sizes.}: inaccurate box size. 
To this end, we replace the size of every detected car that has IoU $> 0.2$ to a ground-truth car with the corresponding ground-truth box size, while keeping its bottom center and rotation unchanged. We see an immediate performance boost in~\autoref{tab:align_gain} (see the Supplementary Material for complete results across all pairs of datasets). In other words, the detector trained from one domain just cannot predict the car size right in the other domains. This observation correlates with our findings in~\autoref{fig:stat} that these datasets have different car sizes. By further analyzing the detected boxes (in~\autoref{fig:adapt}, we apply the detector trained from~\waymo to \kitti), we find that the detector tends to predict box sizes that are similar to the ground-truth sizes in source domain, even though cars in the target domain are indeed physically smaller. We think this is because the detectors trained from the source data carry the learned bias to the target data.

\begin{table}
\small
\caption{\small Cross-dataset performance and gain (in parentheses) by assigning ground-truth box sizes to detected cars while keeping their centers and rotations unchanged. We report \AP of the \emph{Car} category at $\text{IoU}=0.7$, using \PRCNN~\cite{shi2019pointrcnn}. We show adaptation from \kitti to other datasets, and vice versa.}
\vskip-5pt
\label{tab:align_gain}
\centering
\begin{tabular}{c|l|c|c}
Setting & Dataset & \multicolumn{1}{c|}{From \kitti} & \multicolumn{1}{c}{To \kitti} \\ 
\hline
\multirow{4}{*}{Easy} & \argo & \small{\color{black}{65.7} (\color{black}{+38.0})} & \small{\color{black}{59.2} (\color{black}{+25.3})} \\
 & \nusc & \small{\color{black}{33.5} (\color{black}{+20.2})} & \small{\color{black}{63.9} (\color{black}{+50.5})} \\
 & \lyft & \small{\color{black}{74.8} (\color{black}{+23.1})} & \small{\color{black}{58.4} (\color{black}{+19.0})} \\
 & \waymo & \small{\color{black}{77.1} (\color{black}{+65.2})} & \small{\color{black}{78.2} (\color{black}{+65.1})} \\
\hline\multirow{4}{*}{Moderate} & \argo & \small{\color{black}{50.9} (\color{black}{+28.6})} & \small{\color{black}{51.0} (\color{black}{+19.6})} \\
 & \nusc & \small{\color{black}{18.2} (\color{black}{+9.9})} & \small{\color{black}{47.3} (\color{black}{+36.6})} \\
 & \lyft & \small{\color{black}{54.3} (\color{black}{+20.6})} & \small{\color{black}{49.4} (\color{black}{+15.1})} \\
 & \waymo & \small{\color{black}{63.0} (\color{black}{+50.7})} & \small{\color{black}{60.6} (\color{black}{+47.4})} \\
\hline\multirow{4}{*}{Hard} & \argo & \small{\color{black}{49.3} (\color{black}{+27.1})} & \small{\color{black}{52.5} (\color{black}{+19.2})} \\
 & \nusc & \small{\color{black}{17.7} (\color{black}{+8.9})} & \small{\color{black}{45.7} (\color{black}{+35.6})} \\
 & \lyft & \small{\color{black}{53.0} (\color{black}{+18.8})} & \small{\color{black}{52.0} (\color{black}{+18.1})} \\
 & \waymo & \small{\color{black}{59.1} (\color{black}{+46.5})} & \small{\color{black}{60.7} (\color{black}{+48.1})} \\
\hline
\end{tabular}
\vskip-10pt
\end{table}

\begin{table*}
\tabcolsep 1.5pt
	\small
\caption{\small \textbf{Improved 3D object detection across datasets} (evaluated on the validation sets). 
We report \APBEV / \AP of the \emph{Car} category at $\text{IoU}=0.7$, using \PRCNN~\cite{shi2019pointrcnn}.
We investigate \textbf{(OT)} \emph{output transformation} by directly adjusting the predicted box sizes,
\textbf{(SN)} \emph{statistical normalization}, and \textbf{(FS)} \emph{few-shot fine-tuning} (with $10$ labeled instances). We also include \textbf{(Direct)} directly applying the detectors trained on the source domain
and \textbf{(Within)} applying the detectors trained on the target domain for comparison.
We show adaption results from \kitti to other datasets, and vice versa. We mark the best result among Direct, OT, SN, and FS in red fonts, and worst in blue fonts. 
}
\vskip-5pt
\label{tab:big_all}
\centering
\begin{tabular}{c|l|c|c|c|c|c||c|c|c|c|c}
& & \multicolumn{5}{c||}{\textbf{From \kitti} (\kitti as the source; others as the target)} & \multicolumn{5}{c}{\textbf{To \kitti} (\kitti as the target; others as the source)} \\ \hline
Setting & Dataset & Direct & OT & SN & FS & Within & Direct & OT & SN & FS & Within\\
\hline
\multirow{4}{*}{Easy} & \argo & \small{\color{blue}{55.8} \color{black}{/} \color{black}{27.7}} & \small{\color{black}{72.7} \color{black}{/} \color{blue}{9.0}} & \small{\color{black}{74.7} \color{black}{/} \color{black}{48.2}} & \small{\color{red}{75.8} \color{black}{/} \color{red}{49.2}} & \small{\color{black}{79.2} \color{black}{/} \color{black}{57.8}} & \small{\color{black}{69.5} \color{black}{/} \color{black}{33.9}} & \small{\color{blue}{53.3} \color{black}{/} \color{blue}{5.7}} & \small{\color{black}{76.2} \color{black}{/} \color{black}{46.1}} & \small{\color{red}{80.0} \color{black}{/} \color{red}{49.7}} & \small{\color{black}{88.0} \color{black}{/} \color{black}{82.5}} \\
 & \nusc & \small{\color{blue}{47.4} \color{black}{/} \color{black}{13.3}} & \small{\color{black}{55.0} \color{black}{/} \color{blue}{10.4}} & \small{\color{red}{60.8} \color{black}{/} \color{red}{23.9}} & \small{\color{black}{54.7} \color{black}{/} \color{black}{21.7}} & \small{\color{black}{73.4} \color{black}{/} \color{black}{38.1}} & \small{\color{blue}{49.7} \color{black}{/} \color{blue}{13.4}} & \small{\color{black}{75.4} \color{black}{/} \color{black}{31.5}} & \small{\color{black}{83.2} \color{black}{/} \color{black}{35.6}} & \small{\color{red}{83.8} \color{black}{/} \color{red}{58.7}} & \small{\color{black}{88.0} \color{black}{/} \color{black}{82.5}} \\
 & \lyft & \small{\color{blue}{81.7} \color{black}{/} \color{black}{51.8}} & \small{\color{black}{88.2} \color{black}{/} \color{blue}{23.5}} & \small{\color{black}{88.3} \color{black}{/} \color{black}{73.3}} & \small{\color{red}{89.0} \color{black}{/} \color{red}{78.1}} & \small{\color{black}{90.2} \color{black}{/} \color{black}{87.3}} & \small{\color{black}{74.3} \color{black}{/} \color{black}{39.4}} & \small{\color{blue}{71.9} \color{black}{/} \color{blue}{4.7}} & \small{\color{black}{83.5} \color{black}{/} \color{black}{72.1}} & \small{\color{red}{85.3} \color{black}{/} \color{red}{72.5}} & \small{\color{black}{88.0} \color{black}{/} \color{black}{82.5}} \\
 & \waymo & \small{\color{blue}{45.2} \color{black}{/} \color{blue}{11.9}} & \small{\color{black}{86.1} \color{black}{/} \color{black}{16.2}} & \small{\color{black}{84.6} \color{black}{/} \color{black}{53.3}} & \small{\color{red}{87.4} \color{black}{/} \color{red}{70.9}} & \small{\color{black}{90.1} \color{black}{/} \color{black}{85.3}} & \small{\color{blue}{51.9} \color{black}{/} \color{black}{13.1}} & \small{\color{black}{64.0} \color{black}{/} \color{blue}{3.9}} & \small{\color{red}{82.1} \color{black}{/} \color{black}{48.7}} & \small{\color{black}{81.0} \color{black}{/} \color{red}{67.0}} & \small{\color{black}{88.0} \color{black}{/} \color{black}{82.5}} \\
\hline\multirow{4}{*}{Mod.} & \argo & \small{\color{blue}{44.9} \color{black}{/} \color{black}{22.3}} & \small{\color{black}{59.9} \color{black}{/} \color{blue}{7.9}} & \small{\color{red}{61.5} \color{black}{/} \color{red}{38.2}} & \small{\color{black}{60.7} \color{black}{/} \color{black}{37.3}} & \small{\color{black}{69.9} \color{black}{/} \color{black}{44.2}} & \small{\color{black}{56.6} \color{black}{/} \color{black}{31.4}} & \small{\color{blue}{52.2} \color{black}{/} \color{blue}{7.3}} & \small{\color{black}{67.2} \color{black}{/} \color{black}{40.5}} & \small{\color{red}{68.8} \color{black}{/} \color{red}{42.8}} & \small{\color{black}{80.6} \color{black}{/} \color{black}{68.9}} \\
 & \nusc & \small{\color{blue}{26.2} \color{black}{/} \color{black}{8.3}} & \small{\color{black}{30.8} \color{black}{/} \color{blue}{6.8}} & \small{\color{red}{32.9} \color{black}{/} \color{red}{16.4}} & \small{\color{black}{28.7} \color{black}{/} \color{black}{12.5}} & \small{\color{black}{40.7} \color{black}{/} \color{black}{21.2}} & \small{\color{blue}{39.8} \color{black}{/} \color{blue}{10.7}} & \small{\color{black}{58.5} \color{black}{/} \color{black}{27.3}} & \small{\color{red}{67.4} \color{black}{/} \color{black}{31.0}} & \small{\color{black}{67.2} \color{black}{/} \color{red}{45.5}} & \small{\color{black}{80.6} \color{black}{/} \color{black}{68.9}} \\
 & \lyft & \small{\color{blue}{61.8} \color{black}{/} \color{black}{33.7}} & \small{\color{black}{70.1} \color{black}{/} \color{blue}{17.8}} & \small{\color{black}{73.7} \color{black}{/} \color{black}{53.1}} & \small{\color{red}{74.2} \color{black}{/} \color{red}{53.4}} & \small{\color{black}{83.7} \color{black}{/} \color{black}{65.5}} & \small{\color{black}{61.1} \color{black}{/} \color{black}{34.3}} & \small{\color{blue}{60.8} \color{black}{/} \color{blue}{5.6}} & \small{\color{black}{73.6} \color{black}{/} \color{red}{57.9}} & \small{\color{red}{73.9} \color{black}{/} \color{black}{56.2}} & \small{\color{black}{80.6} \color{black}{/} \color{black}{68.9}} \\
 & \waymo & \small{\color{blue}{43.9} \color{black}{/} \color{blue}{12.3}} & \small{\color{black}{69.1} \color{black}{/} \color{black}{13.1}} & \small{\color{black}{74.9} \color{black}{/} \color{black}{49.4}} & \small{\color{red}{75.9} \color{black}{/} \color{red}{55.3}} & \small{\color{black}{85.9} \color{black}{/} \color{black}{67.9}} & \small{\color{blue}{45.8} \color{black}{/} \color{black}{13.2}} & \small{\color{black}{54.9} \color{black}{/} \color{blue}{3.7}} & \small{\color{red}{71.3} \color{black}{/} \color{black}{47.1}} & \small{\color{black}{66.8} \color{black}{/} \color{red}{51.8}} & \small{\color{black}{80.6} \color{black}{/} \color{black}{68.9}} \\
\hline\multirow{4}{*}{Hard} & \argo & \small{\color{blue}{42.5} \color{black}{/} \color{black}{22.2}} & \small{\color{black}{59.3} \color{black}{/} \color{blue}{9.3}} & \small{\color{red}{60.6} \color{black}{/} \color{red}{37.1}} & \small{\color{black}{59.8} \color{black}{/} \color{black}{36.5}} & \small{\color{black}{69.9} \color{black}{/} \color{black}{42.8}} & \small{\color{black}{58.5} \color{black}{/} \color{black}{33.3}} & \small{\color{blue}{53.5} \color{black}{/} \color{blue}{8.6}} & \small{\color{red}{68.5} \color{black}{/} \color{black}{41.9}} & \small{\color{black}{66.3} \color{black}{/} \color{red}{43.0}} & \small{\color{black}{81.9} \color{black}{/} \color{black}{66.7}} \\
 & \nusc & \small{\color{blue}{24.9} \color{black}{/} \color{black}{8.8}} & \small{\color{black}{27.8} \color{black}{/} \color{blue}{7.6}} & \small{\color{red}{31.9} \color{black}{/} \color{red}{15.8}} & \small{\color{black}{27.5} \color{black}{/} \color{black}{12.4}} & \small{\color{black}{40.2} \color{black}{/} \color{black}{20.5}} & \small{\color{blue}{39.6} \color{black}{/} \color{blue}{10.1}} & \small{\color{black}{59.5} \color{black}{/} \color{black}{27.8}} & \small{\color{red}{65.2} \color{black}{/} \color{black}{30.8}} & \small{\color{black}{64.7} \color{black}{/} \color{red}{44.5}} & \small{\color{black}{81.9} \color{black}{/} \color{black}{66.7}} \\
 & \lyft & \small{\color{blue}{57.4} \color{black}{/} \color{black}{34.2}} & \small{\color{black}{66.5} \color{black}{/} \color{blue}{19.1}} & \small{\color{red}{73.1} \color{black}{/} \color{red}{53.5}} & \small{\color{black}{71.8} \color{black}{/} \color{black}{52.9}} & \small{\color{black}{79.3} \color{black}{/} \color{black}{65.5}} & \small{\color{blue}{60.7} \color{black}{/} \color{black}{33.9}} & \small{\color{black}{63.1} \color{black}{/} \color{blue}{6.9}} & \small{\color{red}{75.2} \color{black}{/} \color{red}{58.9}} & \small{\color{black}{74.1} \color{black}{/} \color{black}{56.2}} & \small{\color{black}{81.9} \color{black}{/} \color{black}{66.7}} \\
 & \waymo & \small{\color{blue}{41.5} \color{black}{/} \color{blue}{12.6}} & \small{\color{black}{68.7} \color{black}{/} \color{black}{13.9}} & \small{\color{black}{69.4} \color{black}{/} \color{black}{49.4}} & \small{\color{red}{70.1} \color{black}{/} \color{red}{54.4}} & \small{\color{black}{80.4} \color{black}{/} \color{black}{67.7}} & \small{\color{blue}{46.3} \color{black}{/} \color{black}{12.6}} & \small{\color{black}{58.0} \color{black}{/} \color{blue}{4.1}} & \small{\color{red}{73.0} \color{black}{/} \color{black}{49.7}} & \small{\color{black}{68.1} \color{black}{/} \color{red}{52.9}} & \small{\color{black}{81.9} \color{black}{/} \color{black}{66.7}} \\
\hline\multirow{4}{*}{0-30} & \argo & \small{\color{blue}{58.4} \color{black}{/} \color{black}{34.7}} & \small{\color{black}{73.0} \color{black}{/} \color{blue}{13.7}} & \small{\color{black}{73.1} \color{black}{/} \color{black}{54.2}} & \small{\color{red}{73.6} \color{black}{/} \color{red}{55.2}} & \small{\color{black}{83.3} \color{black}{/} \color{black}{63.3}} & \small{\color{black}{74.2} \color{black}{/} \color{black}{46.8}} & \small{\color{blue}{64.9} \color{black}{/} \color{blue}{10.1}} & \small{\color{black}{83.3} \color{black}{/} \color{black}{53.9}} & \small{\color{red}{84.0} \color{black}{/} \color{red}{56.9}} & \small{\color{black}{88.8} \color{black}{/} \color{black}{84.9}} \\
 & \nusc & \small{\color{blue}{47.9} \color{black}{/} \color{black}{14.9}} & \small{\color{black}{56.2} \color{black}{/} \color{blue}{13.9}} & \small{\color{red}{60.0} \color{black}{/} \color{red}{29.2}} & \small{\color{black}{54.0} \color{black}{/} \color{black}{23.6}} & \small{\color{black}{73.2} \color{black}{/} \color{black}{42.8}} & \small{\color{blue}{50.7} \color{black}{/} \color{blue}{13.9}} & \small{\color{black}{74.6} \color{black}{/} \color{black}{36.6}} & \small{\color{red}{83.6} \color{black}{/} \color{black}{42.8}} & \small{\color{black}{81.2} \color{black}{/} \color{red}{59.8}} & \small{\color{black}{88.8} \color{black}{/} \color{black}{84.9}} \\
 & \lyft & \small{\color{blue}{77.8} \color{black}{/} \color{black}{54.2}} & \small{\color{black}{88.4} \color{black}{/} \color{blue}{27.5}} & \small{\color{black}{88.8} \color{black}{/} \color{black}{75.4}} & \small{\color{red}{89.3} \color{black}{/} \color{red}{77.6}} & \small{\color{black}{90.4} \color{black}{/} \color{black}{88.5}} & \small{\color{black}{75.1} \color{black}{/} \color{black}{45.2}} & \small{\color{blue}{74.8} \color{black}{/} \color{blue}{9.1}} & \small{\color{black}{87.4} \color{black}{/} \color{black}{73.6}} & \small{\color{red}{87.5} \color{black}{/} \color{red}{73.9}} & \small{\color{black}{88.8} \color{black}{/} \color{black}{84.9}} \\
 & \waymo & \small{\color{blue}{48.0} \color{black}{/} \color{blue}{14.0}} & \small{\color{black}{87.7} \color{black}{/} \color{black}{22.2}} & \small{\color{black}{87.1} \color{black}{/} \color{black}{60.1}} & \small{\color{red}{88.7} \color{black}{/} \color{red}{74.1}} & \small{\color{black}{90.4} \color{black}{/} \color{black}{87.2}} & \small{\color{blue}{56.8} \color{black}{/} \color{black}{15.0}} & \small{\color{black}{71.3} \color{black}{/} \color{blue}{4.4}} & \small{\color{red}{85.7} \color{black}{/} \color{black}{59.0}} & \small{\color{black}{84.8} \color{black}{/} \color{red}{71.0}} & \small{\color{black}{88.8} \color{black}{/} \color{black}{84.9}} \\
\hline\multirow{4}{*}{30-50} & \argo & \small{\color{blue}{46.5} \color{black}{/} \color{black}{19.0}} & \small{\color{black}{56.1} \color{black}{/} \color{blue}{5.4}} & \small{\color{red}{61.5} \color{black}{/} \color{red}{31.5}} & \small{\color{black}{59.0} \color{black}{/} \color{black}{29.9}} & \small{\color{black}{72.2} \color{black}{/} \color{black}{39.5}} & \small{\color{blue}{33.9} \color{black}{/} \color{black}{11.8}} & \small{\color{black}{35.1} \color{black}{/} \color{blue}{9.1}} & \small{\color{red}{48.9} \color{black}{/} \color{red}{25.7}} & \small{\color{black}{47.9} \color{black}{/} \color{black}{23.8}} & \small{\color{black}{70.2} \color{black}{/} \color{black}{51.4}} \\
 & \nusc & \small{\color{black}{9.8} \color{black}{/} \color{black}{4.5}} & \small{\color{black}{10.8} \color{black}{/} \color{red}{9.1}} & \small{\color{red}{11.0} \color{black}{/} \color{blue}{2.3}} & \small{\color{blue}{9.5} \color{black}{/} \color{black}{6.1}} & \small{\color{black}{17.1} \color{black}{/} \color{black}{4.1}} & \small{\color{blue}{24.1} \color{black}{/} \color{blue}{3.8}} & \small{\color{black}{35.5} \color{black}{/} \color{black}{15.5}} & \small{\color{black}{44.9} \color{black}{/} \color{black}{18.6}} & \small{\color{red}{45.0} \color{black}{/} \color{red}{25.1}} & \small{\color{black}{70.2} \color{black}{/} \color{black}{51.4}} \\
 & \lyft & \small{\color{blue}{60.1} \color{black}{/} \color{black}{34.5}} & \small{\color{black}{67.4} \color{black}{/} \color{blue}{10.7}} & \small{\color{red}{73.8} \color{black}{/} \color{red}{52.2}} & \small{\color{black}{73.7} \color{black}{/} \color{black}{50.4}} & \small{\color{black}{83.8} \color{black}{/} \color{black}{62.7}} & \small{\color{blue}{39.3} \color{black}{/} \color{black}{16.6}} & \small{\color{black}{43.3} \color{black}{/} \color{blue}{3.9}} & \small{\color{red}{58.3} \color{black}{/} \color{red}{38.0}} & \small{\color{black}{57.7} \color{black}{/} \color{black}{33.3}} & \small{\color{black}{70.2} \color{black}{/} \color{black}{51.4}} \\
 & \waymo & \small{\color{blue}{50.5} \color{black}{/} \color{black}{21.4}} & \small{\color{black}{73.6} \color{black}{/} \color{blue}{10.4}} & \small{\color{red}{78.1} \color{black}{/} \color{black}{54.9}} & \small{\color{black}{78.1} \color{black}{/} \color{red}{57.2}} & \small{\color{black}{87.5} \color{black}{/} \color{black}{68.8}} & \small{\color{blue}{31.7} \color{black}{/} \color{black}{9.3}} & \small{\color{black}{39.8} \color{black}{/} \color{blue}{4.5}} & \small{\color{red}{57.3} \color{black}{/} \color{red}{36.3}} & \small{\color{black}{49.2} \color{black}{/} \color{black}{29.2}} & \small{\color{black}{70.2} \color{black}{/} \color{black}{51.4}} \\
\hline\multirow{4}{*}{50-70} & \argo & \small{\color{blue}{9.2} \color{black}{/} \color{black}{3.0}} & \small{\color{black}{20.5} \color{black}{/} \color{blue}{1.0}} & \small{\color{red}{23.8} \color{black}{/} \color{black}{5.6}} & \small{\color{black}{20.1} \color{black}{/} \color{red}{6.3}} & \small{\color{black}{29.9} \color{black}{/} \color{black}{6.9}} & \small{\color{red}{10.9} \color{black}{/} \color{black}{1.3}} & \small{\color{blue}{8.0} \color{black}{/} \color{blue}{0.8}} & \small{\color{black}{9.1} \color{black}{/} \color{black}{2.6}} & \small{\color{black}{8.1} \color{black}{/} \color{red}{3.8}} & \small{\color{black}{28.8} \color{black}{/} \color{black}{12.0}} \\
 & \nusc & \small{\color{blue}{1.1} \color{black}{/} \color{blue}{0.0}} & \small{\color{black}{1.5} \color{black}{/} \color{black}{1.0}} & \small{\color{black}{3.0} \color{black}{/} \color{red}{2.3}} & \small{\color{red}{3.3} \color{black}{/} \color{black}{1.2}} & \small{\color{black}{9.1} \color{black}{/} \color{black}{9.1}} & \small{\color{blue}{6.5} \color{black}{/} \color{blue}{1.5}} & \small{\color{black}{7.8} \color{black}{/} \color{black}{5.1}} & \small{\color{black}{9.4} \color{black}{/} \color{black}{5.1}} & \small{\color{red}{12.9} \color{black}{/} \color{red}{5.7}} & \small{\color{black}{28.8} \color{black}{/} \color{black}{12.0}} \\
 & \lyft & \small{\color{blue}{33.2} \color{black}{/} \color{black}{9.6}} & \small{\color{black}{41.3} \color{black}{/} \color{blue}{6.8}} & \small{\color{red}{49.9} \color{black}{/} \color{red}{22.2}} & \small{\color{black}{46.8} \color{black}{/} \color{black}{19.4}} & \small{\color{black}{62.7} \color{black}{/} \color{black}{33.1}} & \small{\color{black}{13.6} \color{black}{/} \color{black}{4.6}} & \small{\color{blue}{12.7} \color{black}{/} \color{blue}{0.9}} & \small{\color{red}{21.1} \color{black}{/} \color{black}{6.7}} & \small{\color{black}{17.5} \color{black}{/} \color{red}{8.0}} & \small{\color{black}{28.8} \color{black}{/} \color{black}{12.0}} \\
 & \waymo & \small{\color{blue}{27.1} \color{black}{/} \color{black}{12.0}} & \small{\color{black}{42.6} \color{black}{/} \color{blue}{4.2}} & \small{\color{red}{46.8} \color{black}{/} \color{red}{25.1}} & \small{\color{black}{45.2} \color{black}{/} \color{black}{24.3}} & \small{\color{black}{63.5} \color{black}{/} \color{black}{41.1}} & \small{\color{blue}{5.6} \color{black}{/} \color{black}{1.8}} & \small{\color{black}{7.7} \color{black}{/} \color{blue}{1.1}} & \small{\color{red}{14.4} \color{black}{/} \color{red}{5.7}} & \small{\color{black}{10.5} \color{black}{/} \color{black}{4.8}} & \small{\color{black}{28.8} \color{black}{/} \color{black}{12.0}} \\
\hline
\end{tabular}
\vskip-5pt
\end{table*}

\section{Domain Adaptation Approaches}
\label{sec:approach}

The poor performance due to mislocalization rather than misdetection opens the possibility of adapting a learned detector to a new domain with relatively smaller efforts. 
We investigate two scenarios: (1) a few labeled scenes (\ie, point clouds with 3D box annotations) or (2) the car size statistics of the target domain are available. 
We argue that both scenarios are practical: we can simply annotate for every place a few labeled instances, or get the statistics from the local vehicle offices or car-selling websites. \emph{In the main paper, we will mainly focus on training from \kitti and testing on the others, and vice versa.} We leave other results in the Supplementary Material.

\begin{figure}
\vskip -5pt
	\centering
		\includegraphics[width=0.8\linewidth]{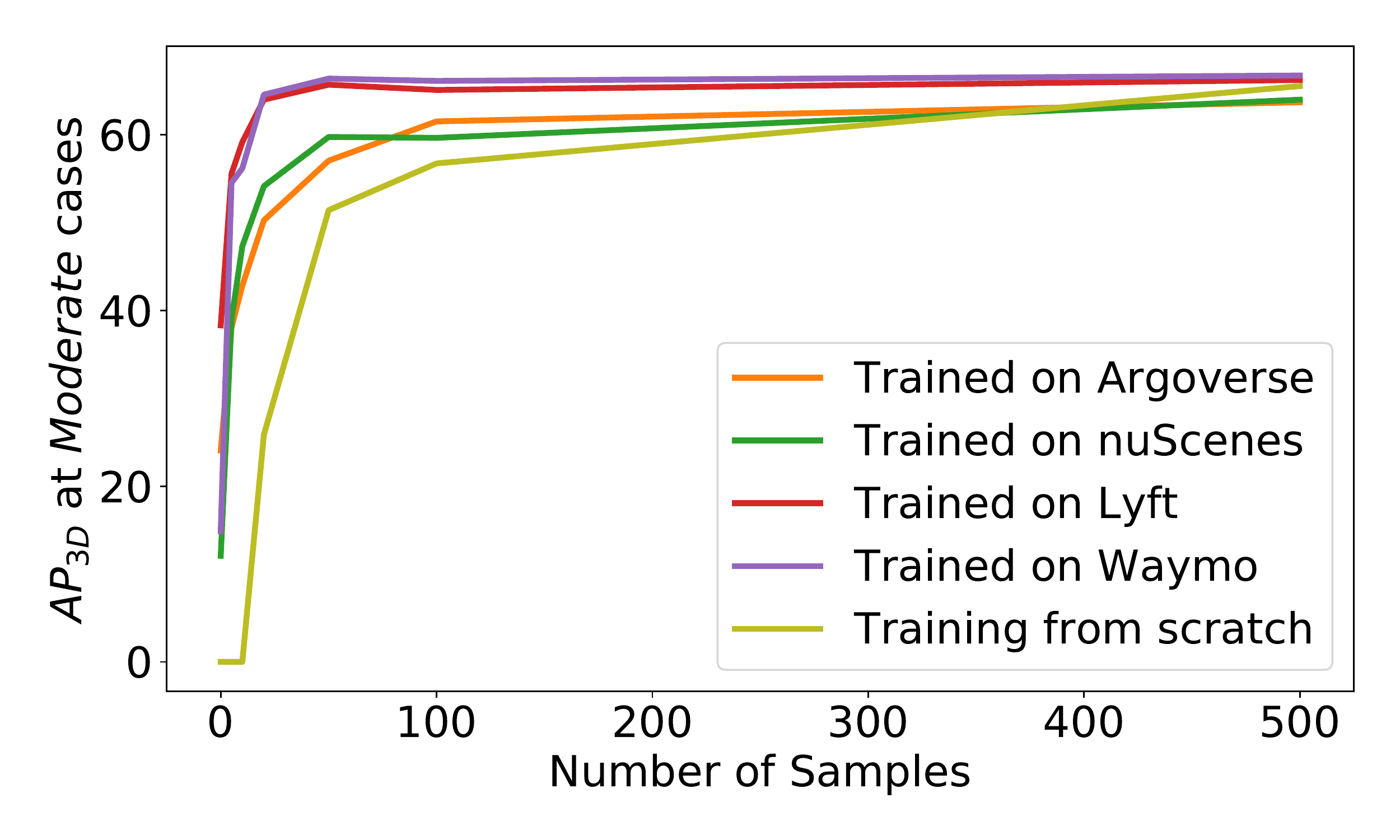}
	\vskip -15pt
	\caption{\small The few-shot fine-tuning performance on \kitti validation set with the model pre-trained on \argo, \nusc, \lyft, and \waymo datasets. The x-axis indicates how many \kitti training images are used for fine-tuning. The y-axis marks \AP (moderate cases). \emph{Scratch} denotes the model trained on the sampled \kitti training images with randomly initialized weights.}
	\label{fig:inat}
	\vskip-10pt
\end{figure}

\noindent\textbf{Few-shot (FS) fine-tuning.} In the first scenario where a few labeled scenes from the target domain are accessible, we investigate fine-tuning the already trained object detector with these few-shot examples. 
As shown in~\autoref{tab:big_all}, using only $10$ labeled scenes (average over five rounds of experiments) of the target domain, we can already improve the \AP by over $20.4\%$ on average when adapting \kitti to other datasets and $24.4\%$ on average when adapting other datasets to \kitti. \autoref{fig:inat} further shows the performance by fine-tuning with different number of scenes. With merely $20$ labeled target scenes, the adapted detector from \lyft and \waymo can already be on a par with that trained from scratch in the target domain with $500$ scenes.

\begin{figure}[htbp]
	\centering
		\includegraphics[width=.7\linewidth]{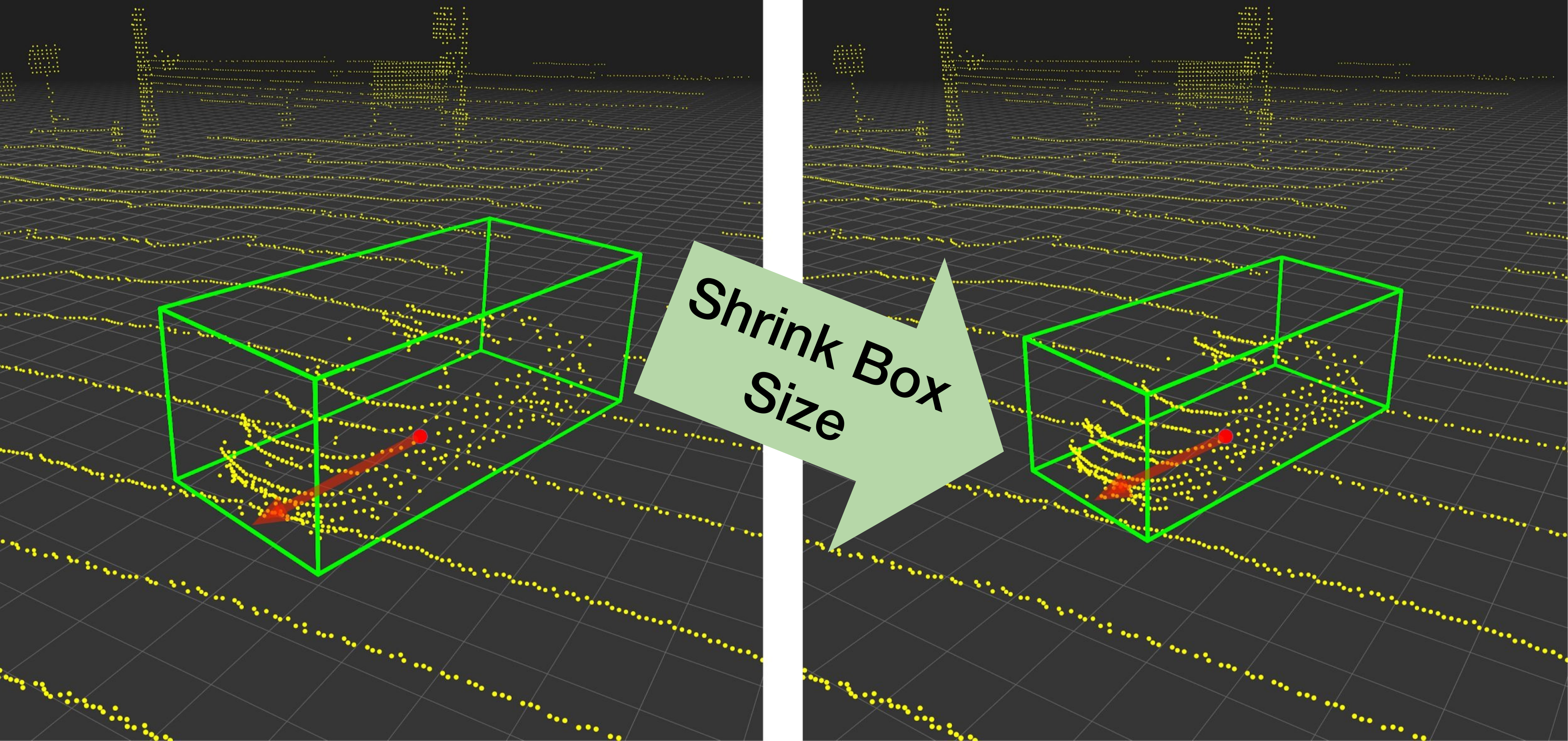}
		\vskip -7pt
	\caption{\small \textbf{ Statistical Normalization (SN).} We shrink (or enlarge) the bounding box sizes (in the output space) and the corresponding point clouds (in the input space) in the training scenes of the source domain to match the \emph{mean} statistics of the target domain. We fine-tune the detector with these modified source scenes.}
	\label{fig:stat_1}
	\vskip-10pt
\end{figure}

\noindent\textbf{Statistical normalization (SN).} For the second scenario where the target statistics (\ie, average height, width, and length of cars) are accessible, we investigate modifying the already trained object detector so that its predicted box sizes can better match the target statistics. We propose a data modification scheme named \emph{statistical normalization} by adjusting the source domain data, as illustrated in~\autoref{fig:stat_1}.
Specifically, we compute the
difference of mean car sizes  between the target domain (TD) and source domain (SD), 
$\Delta = (\Delta h, \Delta w, \Delta l) = (h_\text{TD}, w_\text{TD}, L_\text{TD}) - (h_\text{SD}, w_\text{SD}, L_\text{SD})$,
where $h, w, l$ stand for the height, width, and length, respectively\footnote{Here we obtain the target statistics directly from the dataset. We investigate using the car sales data online in the Supplementary Material.}.
We then modify both the point clouds and the labels in the source domain with respect to $\Delta$. 
For each annotated bounding box of cars, we adjust its size by adding $(\Delta h, \Delta w, \Delta l)$. We also crop the points inside the original box, scale up or shrink their coordinates to fit the adjusted bounding box size accordingly, and paste them back to the point cloud of the scene.
By doing so, we generate new point clouds and labels whose car sizes are much similar to the target domain data.
We then fine-tune the already trained model on the source domain with these data.

\begin{figure}[t]
	\centering
	\includegraphics[width=\linewidth]{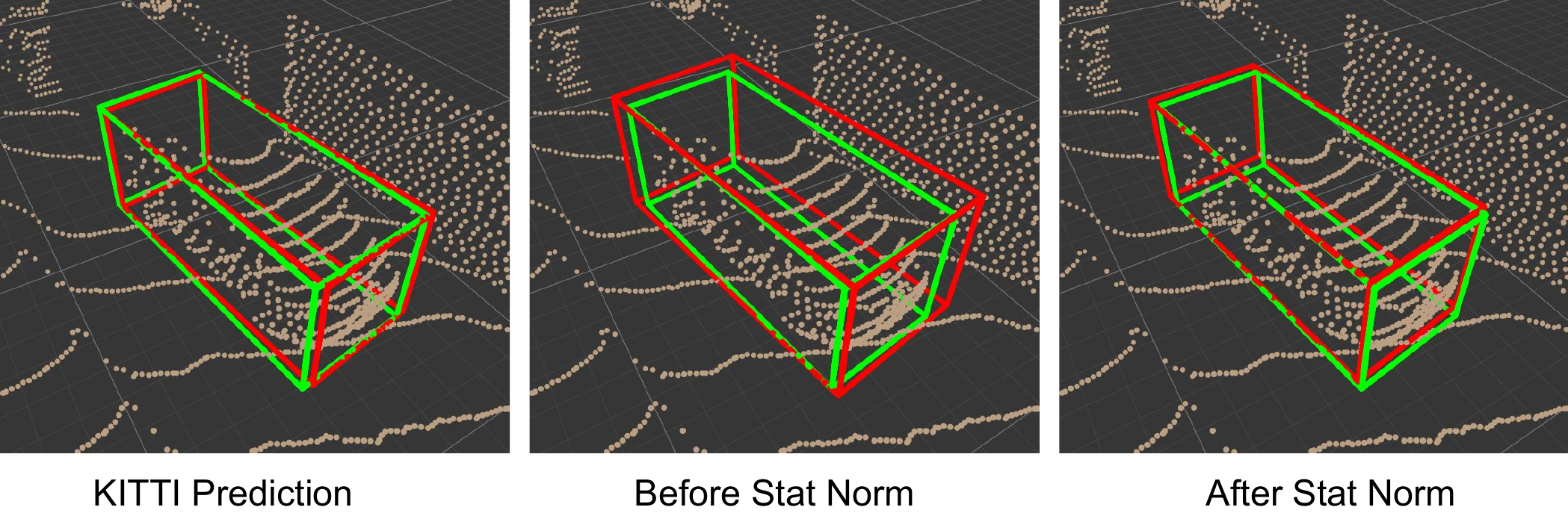}
	\vskip -7pt
	\caption{\small \textbf{Illustration of car prediction on \kitti w/o and w/ statistical normalization (Stat Norm).} The green boxes and red boxes indicate the ground truth and prediction, respectively. The box in the left image is predicted by \PRCNN trained on \kitti. The middle image shows \PRCNN that is pre-trained on \waymo and directly tested on \kitti. With statistical normalization, the model trained on \waymo only (with modified data) can accurately predict the bounding box shown in the right image.}
	\label{fig:stat_visual}
	\vskip -10pt
\end{figure}

Surprisingly, with such a simple method that does not requires labeled target domain data, the performance is significantly improved (see~\autoref{tab:big_all}) between \kitti and other datasets that obviously contain cars of different styles (\ie, one in Germany, and others in the USA).
\autoref{fig:adapt} and \autoref{fig:stat_visual} further analyze the prediction before and after statistical normalization. We see a clear shift of the histogram (predicted box) from the source to the target domain.

\noindent\textbf{Output transformation (OT).} We investigate an even simpler approach by directly adjusting the detector's prediction without fine-tuning --- by adding $(\Delta h, \Delta w, \Delta l)$ to the predicted size. As shown in~\autoref{tab:big_all}, this approach does not always improve but sometimes degrade the accuracy. This is because when we apply the source detector to the target domain, the predicted box sizes do slightly deviate from the source statistics to the target ones due to the difference of object sizes in the input signals (see~\autoref{fig:adapt}). Thus, simply adding $(\Delta h, \Delta w, \Delta l)$ may \emph{over-correct} the bias. We hypothesize that by searching a suitable scale for addition or designing more intelligent output transformations can alleviate this problem and we leave them for future work.

\noindent\textbf{Discussion.} 
As shown in \autoref{tab:big_all}, statistical normalization largely improves over direct applying the source-domain detector. For some pairs of data sets (\eg, from \kitti to \lyft, the \APBEV after statistical normalization is encouraging, largely closing the gap to the \textbf{Within} performance.

Compared to domain adaptation on 2D images, there are more possible factors of domain gaps in 3D. While the box size difference is just one factor, we find addressing it to be highly effective in closing the gaps. 
This factor is rarely discussed in other domain adaptation tasks. 
We thus expect it and our solution to be valuable additions to the community.

\section{Conclusion}
\label{sec:disc}
In conclusion, in this paper we are the first (to our knowledge) to provide and investigate a standardized form of most widely-used 3D object detection datasets for autonomous driving. Although na\"ive adaptation across datasets is unsurprisingly difficult, we observe that, surprisingly, there appears to be a single dominant factor that explains a majority share of the adaptation gap: varying car sizes across different geographic regions. That car sizes play such an important role in adaptation ultimately makes sense. No matter if the detection is based on LiDAR or stereo cameras, cars are only observed from one side --- and the depth of the bounding boxes must be estimated based on experience. If a deep network trained in Germany encounters an American Ford F-Series truck (with $5.3$m length), it has little chance to correctly estimate the corresponding bounding box. It is surprising, however, that just matching the mean size of cars in the areas during fine-tuning already reduces this uncertainty so much. We hope that this publication will kindle interests in the exciting problem of cross-dataset domain adaptation for 3D object detection and localization, and that researchers will be careful to first apply simple global corrections before developing new computer vision algorithms to tackle the remaining adaptation gap. 
	
	\section*{Acknowledgments}
	{\small
    This research is supported by grants from the National Science Foundation NSF (III-1618134, III-1526012, IIS-1149882, IIS-1724282, OAC-1934714, and TRIPODS-1740822), the Office of Naval Research DOD (N00014-17-1-2175), the Bill and Melinda Gates Foundation, and the Cornell Center for Materials Research with funding from the NSF MRSEC program (DMR-1719875). We are thankful for generous support by Zillow and SAP America Inc.}
	{\small
		\bibliographystyle{ieee_fullname}
		\bibliography{main}
	}
		
	\clearpage
\appendix
\begin{center}
	\textbf{\Large Supplementary Material}
\end{center}
\definecolor{deepblue}{rgb}{0,0,0.5}
\definecolor{deepred}{rgb}{0.6,0,0}
\definecolor{deepgreen}{rgb}{0,0.5,0}
\renewcommand{\thesection}{S\arabic{section}}
\renewcommand{\thetable}{S\arabic{table}}
\renewcommand{\thefigure}{S\arabic{figure}}

In this Supplementary Material, we provide details omitted in the main paper.
\begin{itemize}
    \item \autoref{suppl-sec:format}: data format conversion (\autoref{sec:data} of the main paper).
    \item \autoref{suppl-sec:metric}: evaluation metric (\autoref{ssec:exp_setup} of the main paper).
	\item \autoref{suppl-sec:dataset}: additional results on dataset discrepancy (\autoref{ssec:analysis} and \autoref{ssec:detector_perf} of the main paper).
	\item \autoref{suppl-sec:pixor}: object detection using PIXOR~\cite{yang2018pixor} (\autoref{ssec:within} and \autoref{exp: across} of the main paper).
	\item \autoref{suppl-sec:tables}: object detection using \PRCNN with different adaptation methods (\autoref{ssec:detector_perf} and \autoref{sec:approach} of the main paper).
	\item \autoref{suppl-sec:Quali}: additional qualitative results (\autoref{sec:approach} of the main paper).
\end{itemize}

\section{Converting Datasets into \kitti Format}
\label{suppl-sec:format}
In this section we describe in detail how we convert \argo~\cite{argoverse}, \nusc~\cite{nuscenes2019}, \lyft~\cite{lyft2019}, and \waymo~\cite{waymo_open_dataset} into \kitti~\cite{geiger2013vision,geiger2012we} format. As the formatting of images, point clouds and camera calibration information is trivial, and label fields such as $alpha$ and $rotation_y$ have been well-defined, we only discuss the labeling process with non-deterministic definitions.

\subsection{Object filtering}
Due to the fact that \kitti focuses on objects that appear in the camera view, we follow its setting and discard all object annotations outside the frontal camera view.
To allow truncated objects, we project the 8 corners of each object's 3D bounding box onto the image plane. An object will be discarded if all its 8 corners fall out of the image boundary. 
To make other datasets consistent with \kitti, we do not consider labeled objects farther than $70$ meters.

\subsection{Matching categories with \kitti}
Since the taxonomy of object categories among datasets are misaligned, it is necessary to re-label each dataset in the same way as \kitti does. As we focus on car detection, here we describe how we construct the new \emph{car} and \emph{truck} categories for each dataset except \kitti in \autoref{tab:newcar}. The \emph{truck} category is also important since detected trucks are treated as false positives when we look at the \emph{car} category. We would like to point out that \waymo labels all kinds of vehicles as \emph{car}s. A model trained on \waymo thus will tend to predict trucks or other vehicles as cars. Therefore, directly applying a model trained on \waymo to other datasets will lead to higher false positive rates. For other datasets, the definition between categories can vary (\eg, \argo label Ford F-Series as cars; \nusc labels some as trucks) and result in cross-domain accuracy drop even if the data are collected at similar locations with similar sensors.

\begin{table}[t!]
\caption{The original categories in each dataset that we include into the \emph{car} and \emph{truck} categories following the \kitti label formulation.}
\label{tab:newcar}
\centering
\center
\begin{tabular}{l|c|c}
\hline
Dataset & Car & Truck \\
\hline
\centered{\argo} & \centered{\{VEHICLE\}} & \shortstack{\{LARGE\_VEHICLE,\\ BUS, TRAILER, \\ SCHOOL\_BUS\}} \\
\hline
\centered{\nusc} & \centered{\{car\}} & \shortstack{\{bus, trailer,\\ construction\_vehicle,\\ truck\}} \\
\hline
\centered{\lyft} & \centered{\{Car\}} & \shortstack{\{other\_vehicle,\\ truck, bus,\\ emergency\_vehicle\}} \\
\hline
\centered{\waymo} & \centered{\{Car\}} & $\varnothing$ \\
\hline
\end{tabular}
\end{table}

\subsection{Handling missing 2D bounding boxes}
To annotate each object in the image with a 2D bounding box (the information is used by the original \kitti metric), we first compute 8 corners of its 3D bounding box, and then calculate their pixel coordinates $\{(cx_n, cy_n), 1\leq n \leq 8\}$. We then draw the smallest bounding box $(x_1, y_1, x_2, y_2)$ that contains all corners whose projections fall in the image plane:
\begin{align}
x_1&=\max(\mathop{\min}_{0 \leq n < 8} cx_n, 0),\nonumber\\
y_1&=\max(\mathop{\min}_{0 \leq n < 8} cy_n, 0),\nonumber\\
x_2&=\min(\mathop{\max}_{0 \leq n < 8} cx_n, width),\label{eqn:2dbbox}\\
y_2&=\min(\mathop{\max}_{0 \leq n < 8} cy_n, height),\nonumber
\end{align}
where $width$ and $height$ denote the width and height of the 2D image, respectively.

\subsection{Calculating truncation values}
Following the \kitti formulation, the \emph{truncation} value refers to how much of an object locates beyond image boundary. With \autoref{eqn:2dbbox} we estimate it by calculating how much of the object's 2D uncropped bounding box is outside the image boundary:
\begin{align}
x_1'&=\mathop{\min}_{0 \leq n < 8} cx_n, \nonumber\\
y_1'&=\mathop{\min}_{0 \leq n < 8} cy_n,\nonumber\\
x_2'&=\mathop{\max}_{0 \leq n < 8} cx_n,\\
y_2'&=\mathop{\max}_{0 \leq n < 8} cy_n,\nonumber\\
\text{truncation} &=1 - \frac{(x_2 - x_1) \times (y_2 - y_1)}{(x_2' - x_1') \times (y_2' - y_1')}.\nonumber
\end{align}

\subsection{Calculating occlusion values}
We estimate the occlusion value of objects by approximating car shapes with corresponding 2D bounding boxes. The \emph{occlusion} value is thus derived by computing the percentage of pixels occluded by bounding boxes from closer objects. We discretize the $0\text{--}1$ $\text{occlusion}$ value into \kitti's $\{0, 1, 2, 3\}$ labels by equally dividing the interval into 4 parts. We describe in \autoref{alg:occlusion} the detail of how we compute $\text{occlusion}$ value for each object.

\begin{algorithm}
	\SetKwInOut{Input}{Input}
	\SetKwInOut{Output}{Output}
	\caption{Computing occlusion of objects from a single scene}\label{alg:occlusion}
	\vspace{0.25\baselineskip}
	\Input{Image height $H$, image width $W$, object list $objs$.}
	\BlankLine
	$canvas \leftarrow Array([H, W]) $\;
	\For{$x \in \{0, 1, \dots, W-1\}$}{
    	\For{$y \in \{0, 1, \dots, H-1\}$}{
    		$canvas[x, y] \leftarrow -1 $\;
    	} 
	} 
	$objs \leftarrow Sort(objs, key=depth, order=descending)$\;
	\For{$obj \in objs$}{
	    $[x_1, x_2, y_1, y_2] = obj.bounding\_bbox$\;
    	\For{$x \in \{x_1, x_1+1, \dots, x_2-1\}$}{
        	\For{$y \in \{y_1, y_1+1, \dots, y_2-1\}$}{
        		$canvas[x, y] \leftarrow obj.id $\;
        	} 
    	} 
	} 
	\For{$obj \in objs$}{
	    $[x_1, x_2, y_1, y_2] = obj.bounding\_bbox$\;
	    $cnt \leftarrow 0$\;
    	\For{$x \in \{x_1, x_1+1, \dots, x_2-1\}$}{
        	\For{$y \in \{y_1, y_1+1, \dots, y_2-1\}$}{
        	    \If{$canvas[x, y]=obj.id$}{
        	        $cnt \leftarrow cnt + 1$\;
        	    }
        	} 
    	} 
		$obj.occlusion \leftarrow 1 - \frac{cnt}{(x_2 - x_1) \times (y_2 - y_1)} $\;
	} 
\end{algorithm}

\begin{table*}[th!]
\caption{3D object detection results across multiple datasets using the original \kitti evaluation metric (pixel thresholds). We apply
\PRCNN
~\cite{shi2019pointrcnn}. We report average precision (AP) of the \emph{Car} category in bird's-eye view and 3D (AP$_{\text{BEV}}$ / AP$_{\text{3D}}$, $\text{IoU}=0.7$) and compare object detection accuracy of different difficulties. The results are less comparable due to misaligned difficulty partitions among datasets. Red color: best generalization (per column and per setting); blue color: worst generalization; bold font: within-domain results.}
\label{tab:5x52d}
\centering
\begin{tabular}{c|l|ccccc}
\hline
Setting & Source$\backslash$Target & \kitti & \argo & \nusc & \lyft & \waymo \\
\hline
\multirow{5}{*}{Easy} & \kitti & \color{black}{\textbf{88.0}} \color{black}{/} \color{black}{\textbf{82.3}} & \color{blue}{44.2} \color{black}{/} \color{black}{21.4} & \color{blue}{27.5} \color{black}{/} \color{blue}{7.1} & \color{blue}{72.3} \color{black}{/} \color{black}{45.5} & \color{blue}{42.1} \color{black}{/} \color{blue}{10.6} \\
 & \argo & \color{black}{68.6} \color{black}{/} \color{black}{31.5} & \color{black}{\textbf{69.9}} \color{black}{/} \color{black}{\textbf{43.6}} & \color{black}{28.3} \color{black}{/} \color{black}{11.4} & \color{black}{76.8} \color{black}{/} \color{black}{56.4} & \color{black}{73.5} \color{black}{/} \color{black}{34.2} \\
 & \nusc & \color{blue}{49.4} \color{black}{/} \color{black}{13.2} & \color{black}{57.0} \color{black}{/} \color{blue}{16.5} & \color{black}{\textbf{43.4}} \color{black}{/} \color{black}{\textbf{21.3}} & \color{red}{83.0} \color{black}{/} \color{blue}{31.8} & \color{black}{71.7} \color{black}{/} \color{black}{28.2} \\
 & \lyft & \color{red}{72.6} \color{black}{/} \color{red}{38.9} & \color{red}{66.9} \color{black}{/} \color{red}{33.2} & \color{red}{35.5} \color{black}{/} \color{black}{13.1} & \color{black}{\textbf{86.4}} \color{black}{/} \color{black}{\textbf{77.1}} & \color{red}{78.0} \color{black}{/} \color{red}{54.6} \\
 & \waymo & \color{black}{52.0} \color{black}{/} \color{blue}{13.1} & \color{black}{64.9} \color{black}{/} \color{black}{29.4} & \color{black}{31.5} \color{black}{/} \color{red}{14.3} & \color{black}{82.5} \color{black}{/} \color{red}{68.8} & \color{black}{\textbf{85.3}} \color{black}{/} \color{black}{\textbf{71.7}} \\
\hline\multirow{5}{*}{Moderate} & \kitti & \color{black}{\textbf{86.0}} \color{black}{/} \color{black}{\textbf{74.7}} & \color{blue}{44.9} \color{black}{/} \color{black}{22.3} & \color{blue}{26.2} \color{black}{/} \color{blue}{8.3} & \color{blue}{63.2} \color{black}{/} \color{black}{36.3} & \color{blue}{43.9} \color{black}{/} \color{blue}{12.3} \\
 & \argo & \color{black}{65.2} \color{black}{/} \color{black}{36.6} & \color{black}{\textbf{69.8}} \color{black}{/} \color{black}{\textbf{44.2}} & \color{black}{27.6} \color{black}{/} \color{black}{11.8} & \color{black}{68.5} \color{black}{/} \color{black}{43.6} & \color{black}{72.1} \color{black}{/} \color{black}{35.1} \\
 & \nusc & \color{blue}{45.4} \color{black}{/} \color{blue}{12.1} & \color{black}{56.5} \color{black}{/} \color{blue}{17.1} & \color{black}{\textbf{40.7}} \color{black}{/} \color{black}{\textbf{21.2}} & \color{black}{73.4} \color{black}{/} \color{blue}{26.3} & \color{black}{68.1} \color{black}{/} \color{black}{30.7} \\
 & \lyft & \color{red}{67.3} \color{black}{/} \color{red}{38.3} & \color{black}{62.4} \color{black}{/} \color{red}{35.3} & \color{red}{33.6} \color{black}{/} \color{black}{12.3} & \color{black}{\textbf{79.6}} \color{black}{/} \color{black}{\textbf{66.8}} & \color{red}{77.3} \color{black}{/} \color{red}{53.1} \\
 & \waymo & \color{black}{51.5} \color{black}{/} \color{black}{14.9} & \color{red}{64.4} \color{black}{/} \color{black}{29.8} & \color{black}{28.9} \color{black}{/} \color{red}{13.7} & \color{red}{75.5} \color{black}{/} \color{red}{58.2} & \color{black}{\textbf{85.6}} \color{black}{/} \color{black}{\textbf{67.9}} \\
\hline\multirow{5}{*}{Hard} & \kitti & \color{black}{\textbf{85.7}} \color{black}{/} \color{black}{\textbf{74.8}} & \color{blue}{42.5} \color{black}{/} \color{black}{22.2} & \color{blue}{24.9} \color{black}{/} \color{blue}{8.8} & \color{blue}{62.0} \color{black}{/} \color{black}{34.9} & \color{blue}{41.4} \color{black}{/} \color{blue}{12.6} \\
 & \argo & \color{black}{63.5} \color{black}{/} \color{red}{37.8} & \color{black}{\textbf{69.8}} \color{black}{/} \color{black}{\textbf{42.8}} & \color{black}{26.8} \color{black}{/} \color{red}{14.5} & \color{black}{65.9} \color{black}{/} \color{black}{44.4} & \color{black}{68.5} \color{black}{/} \color{black}{36.7} \\
 & \nusc & \color{blue}{42.2} \color{black}{/} \color{blue}{11.1} & \color{black}{53.2} \color{black}{/} \color{blue}{16.7} & \color{black}{\textbf{40.2}} \color{black}{/} \color{black}{\textbf{20.5}} & \color{black}{73.0} \color{black}{/} \color{blue}{27.8} & \color{black}{66.8} \color{black}{/} \color{black}{29.0} \\
 & \lyft & \color{red}{65.0} \color{black}{/} \color{black}{37.0} & \color{red}{62.8} \color{black}{/} \color{red}{35.8} & \color{red}{30.6} \color{black}{/} \color{black}{11.7} & \color{black}{\textbf{79.7}} \color{black}{/} \color{black}{\textbf{67.3}} & \color{red}{76.6} \color{black}{/} \color{red}{53.8} \\
 & \waymo & \color{black}{48.9} \color{black}{/} \color{black}{14.4} & \color{black}{61.6} \color{black}{/} \color{black}{29.0} & \color{black}{28.4} \color{black}{/} \color{black}{14.1} & \color{red}{75.5} \color{black}{/} \color{red}{55.8} & \color{black}{\textbf{80.2}} \color{black}{/} \color{black}{\textbf{67.6}} \\
\hline
\end{tabular}
\end{table*}

\begin{table}[]
\caption{Percentage (\%) of data (total annotated cars with depths $\in [0, 70]$ meters)  in each difficult partition with old / new difficulty metric. The new \emph{easy} threshold selects much fewer data than the old metric on all datasets except \kitti.}
\label{tab:partition}
\centering
\begin{tabular}{c|l|ccc}
\hline
 & Dataset & Easy & Moderate & Hard \\
\hline
\parbox[t]{2mm}{\multirow{5}{*}{\rotatebox[origin=c]{90}{Training Set}}} & \kitti & 21.7 / 21.6 & 55.5 / 67.8 & 76.1 / 91.0 \\
 & \argo & 27.7 / 14.9 & 40.5 / 40.5 & 59.6 / 59.6 \\
 & \nusc & 31.9 / 13.9 & 47.2 / 47.2 & 64.8 / 64.8 \\
 & \lyft & 25.0 / 15.5 & 50.4 / 54.9 & 64.9 / 70.5 \\
 & \waymo & 29.0 / 10.7 & 40.1 / 40.1 & 58.7 / 58.7 \\
\hline
\parbox[t]{2mm}{\multirow{5}{*}{\rotatebox[origin=c]{90}{Validation Set}}} & \kitti & 20.4 / 20.4 & 55.4 / 65.5 & 77.1 / 88.4 \\
 & \argo & 29.2 / 14.3 & 41.7 / 41.7 & 60.6 / 60.6 \\
 & \nusc & 38.3 / 18.4 & 53.6 / 53.6 & 68.5 / 68.5 \\
 & \lyft & 25.3 / 15.5 & 52.5 / 57.7 & 66.9 / 73.3 \\
 & \waymo & 30.3 / 10.3 & 42.3 / 42.3 & 60.9 / 60.9 \\
\hline
\end{tabular}
\end{table}

\section{The New Difficulty Metric}
\label{suppl-sec:metric}
In \autoref{ssec:exp_setup} of the main paper, we develop a new difficulty metric to evaluate object detection (\ie, how to define easy, moderate, and hard cases) so as to better align different datasets. 
Concretely, \kitti defines its \emph{easy}, \emph{moderate}, and \emph{hard} cases according to truncation, occlusion, and 2D bounding box height (in pixels) of ground-truth annotations. The 2D box height (the threshold at 40 pixels) is meant to differentiate far-away and nearby objects: the \emph{easy} cases only contain nearby objects. However, since the datasets we compare are collected using cameras of different focal lengths and contain images of different resolutions, directly applying the \kitti definition may not well align datasets. For example, a car at $50$ meters is treated as a moderate case in \kitti but may be treated as a easy case in other datasets.

To resolve this issue, we re-define detection difficulty based on object truncation, occlusion, and \textbf{depth range (in meters)}, which completely removes the influences of cameras. In developing this new metric we hope to achieve similar case partitions to the original metric of \kitti. To this end, we estimate the distance thresholds with

\begin{equation}
    D = \frac{f_v \times H}{h},
\end{equation}
where $D$ denotes depth, $f_v$ denotes vertical camera focal length, and $H$ and $h$ are object height in the 3D camera space and the 2D image space, respectively. For a car of average height ($1.53$ meters) in \kitti, the corresponding depth for $40$ pixels is $27.03$ meters.
We therefore select $30$ meters as the new threshold to differentiate \emph{easy} from \emph{moderate} and \emph{hard} cases. 
For \emph{moderate} and \emph{hard} cases, we disregard cars with depths larger than $70$ meters since most of the annotated cars in \kitti are within this range.
\autoref{tab:partition} shows the comparison between old and new difficulty partitions. The new metric contains fewer easy cases than the old metric for all but the \kitti dataset. This is because that the other datasets use either larger focal lengths or resolutions: the objects  in images are therefore larger than in \kitti. We note that, the moderate cases contain all the easy cases, and the hard cases contain all the easy and moderate cases.

We also report in \autoref{tab:5x52d} the detection results within and across datasets using the old metric, in comparison to \autoref{tab:cross_inference} of the main paper which uses the new metric. One notable difference is that for the easy cases in the old metric, both the within and across domain performances drop for all but \kitti datasets, since many far-away cars (which are hard to detect) in the other datasets are treated as easy cases in the old metric.

\section{Dataset discrepancy}
\label{suppl-sec:dataset}

\begin{figure}[!htbp]
    \centering
    \includegraphics[width=\linewidth]{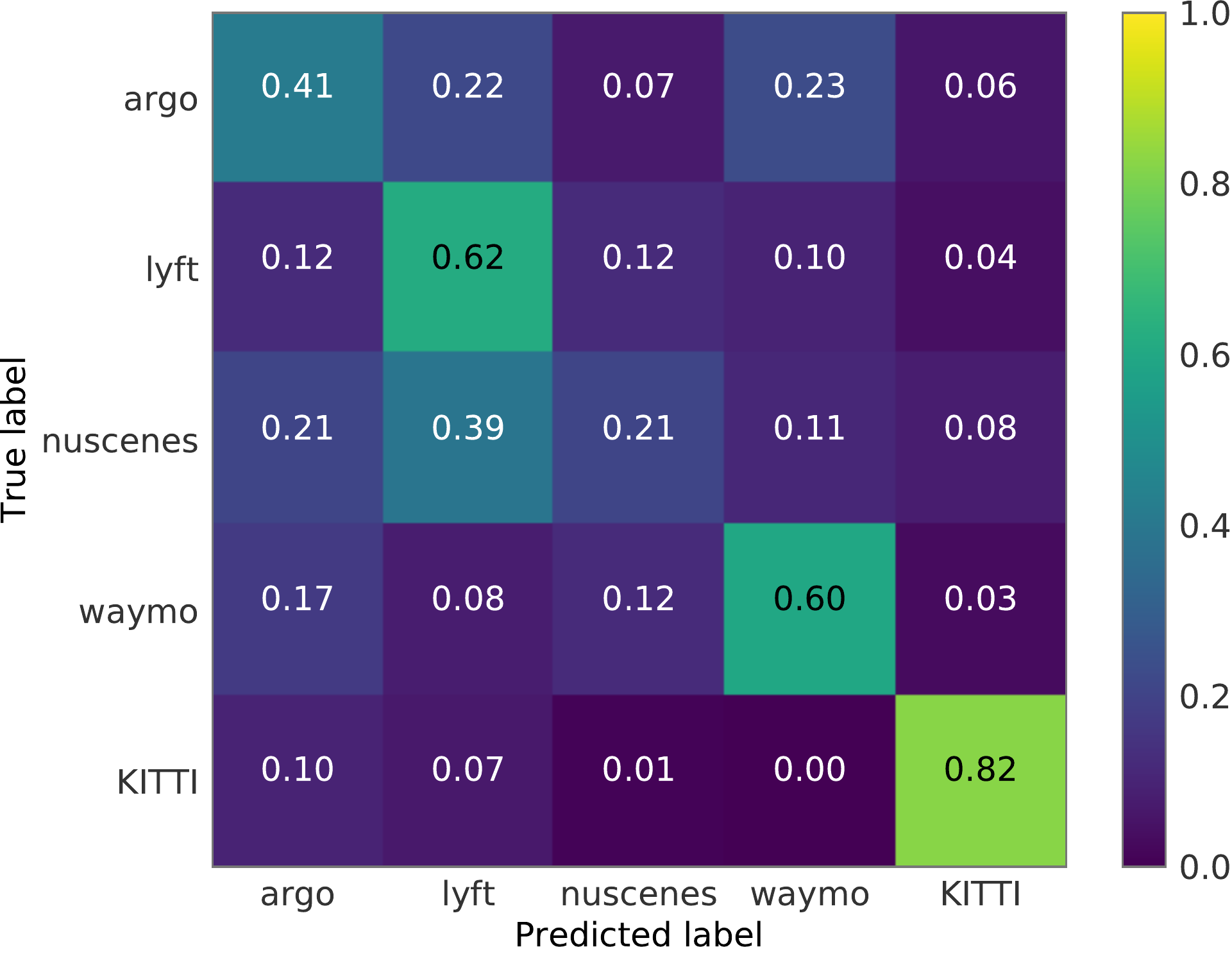}
    \caption{The confusion matrix of predicting which dataset an object belongs to using SVM with the RBF kernel. We take the $(\text{height}, \text{width}, \text{length})$ of car objects as inputs and the corresponding labels are the datasets the objects belong to.}
    \label{fig:label_cm}
    \vskip-5pt
\end{figure}

We have shown the box size distributions of each dataset in \autoref{fig:stat} of the main paper. We also calculate the mean of the bounding box sizes in~\autoref{table:size}. There is a huge gap of size between \kitti and the other four datasets. In addition, we train an SVM classifier with the RBF kernel to predict which dataset a bounding box belongs to and present the confusion matrix result in \autoref{fig:label_cm} (row: ground truth; column: prediction). The model has a very high confidence to distinguish \kitti from the other datasets. 

\begin{table}[t!]
\caption{The average size (meters) of 3D ground truth bounding boxes of the five datasets.}
\label{table:size}
\centering
\begin{tabular}{l|c|c|c}
\hline
Dataset   & Width & Height & Length \\ \hline
KITTI     & 1.62   & 1.53    & 3.89    \\
Argoverse & 1.96   & 1.69    & 4.51    \\
nuScenes  & 1.96   & 1.73    & 4.64    \\
Lyft      & 1.91   & 1.71    & 4.73    \\
Waymo     & 2.11   & 1.79    & 4.80    \\ \hline
\end{tabular}
\end{table}

\begin{figure}[!htbp]
    \centering
    \includegraphics[width=\linewidth]{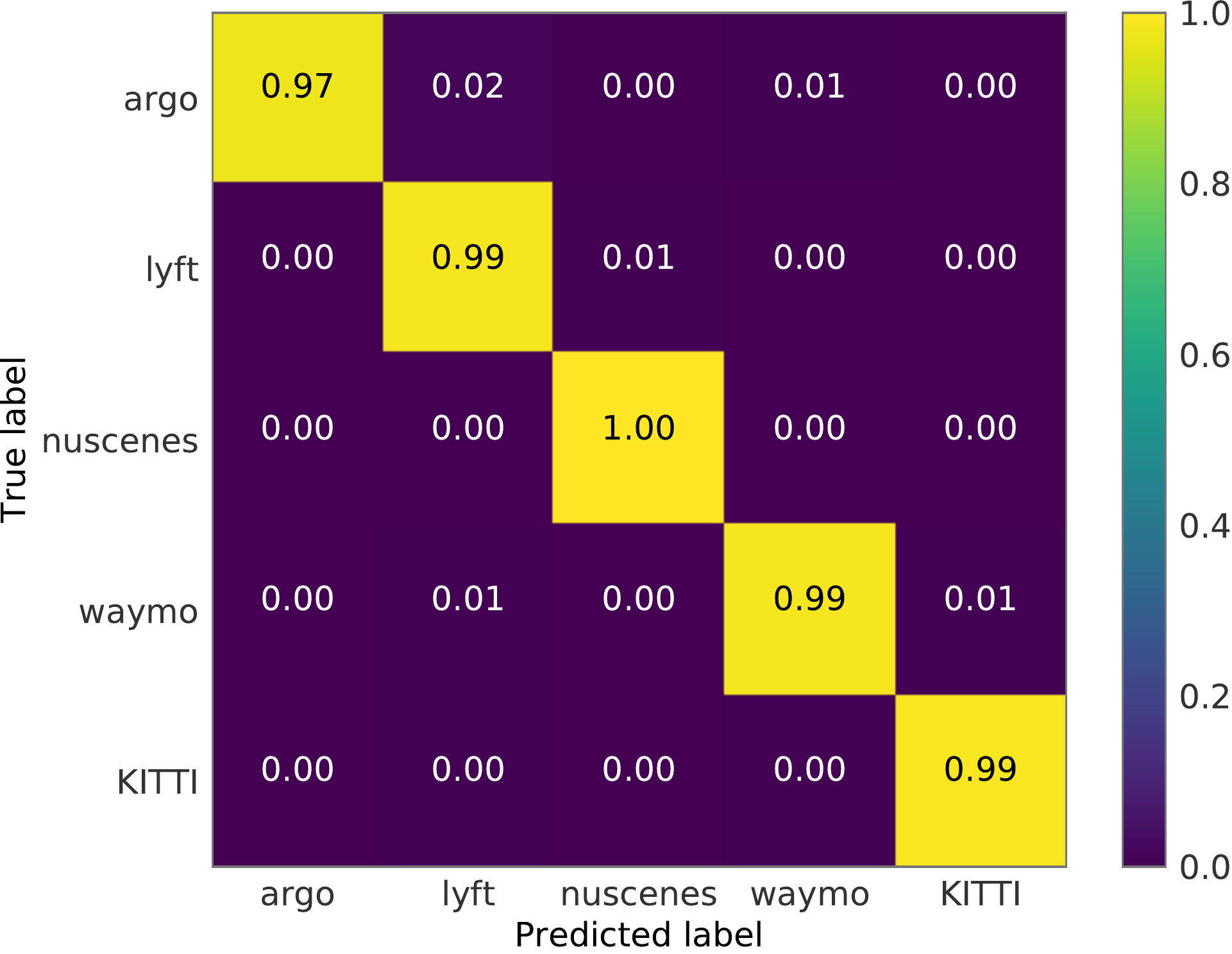}
    \caption{The confusion matrix of predicting which dataset a \textbf{Car} point cloud belongs to using PointNet++. We extract the points inside the ground truth \textbf{Car} bounding box as inputs, and the corresponding labels are the datasets the bounding boxes belong to.}
    \label{fig:ptc_cm}
    \vskip-5pt
\end{figure}

We further train a point cloud classifier to tell which dataset a point cloud of car belongs to, using PointNet++~\cite{qi2017pointnet++} as the backbone. For each dataset, we sample 8,000 object point cloud instances as training examples and 1,000 as testing examples.  We show the confusion matrix in \autoref{fig:ptc_cm}. The classifier can almost perfectly classify the point clouds. Compared to \autoref{fig:label_cm}, we argue that not only the bounding box sizes, but also the point cloud styles (e.g., density, number of laser beams, etc) of cars contribute to dataset discrepancy. Interestingly, while the second factor seems to be more informative in differentiating datasets, the first factor is indeed the main cause of poor transfer performance among datasets\footnote{As mentioned in the main paper, \PRCNN applies point re-sampling so that every scene (in RPN) and object proposal (in RCNN) will have the same numbers of input points. Such an operation could reduce the point cloud differences across domains.}.

\section{PIXOR Results}
\label{suppl-sec:pixor}

We report object detection results using PXIOR~\cite{yang2018pixor}, which takes voxelized tensors instead of point clouds as input. We implement the algorithm ourselves, achieving comparable results as the ones in~\cite{yang2018pixor}. We report the results in~\autoref{tab:pixor5x5}. PXIOR performs pretty well if the model is trained and tested within the same dataset, suggesting that its model design does not over-fit to \kitti.

We also see a clear performance drop when we train a model on one dataset and test it on the other datasets. The drop is more severe than applying the \PRCNN detector in many cases. We surmise that the re-sampling operation used in \PRCNN might make the difference. We therefore apply the same re-sampling operation on the input point cloud before inputting it to \PIXOR. \autoref{tab:pixor_subsample} shows the results: re-sampling does improve the performance in applying the \waymo detector to other datasets.
This is likely because \waymo has the most LiDAR points on average and re-sampling reduces the number, making it more similar to that of the other datasets.
We expect that tuning the number of points in re-sampling can further boost the performance.

\begin{table*}[!htbp]
\caption{3D object detection across multiple datasets (evaluated on the validation sets). We report average precision (AP) of the \emph{Car} category in the bird's-eye view (AP$_{\text{BEV}}$) at $\text{IoU}=0.7$, using PIXOR~\cite{yang2018pixor}. We report results at different difficulties (following the \kitti benchmark, but we replace the $40$, $25$, $25$ pixel thresholds on 2D bounding boxes with $30$, $70$, $70$ meters on object depths, for \emph{Easy}, \emph{Moderate}, and \emph{Hard} cases, respectively) and different depth ranges (using the same truncation and occlusion thresholds as \kitti \emph{Hard} case). The results show a significant performance drop in cross-dataset inference. We indicate the best generalization results per column and per setting
by red fonts and the worst by blue fonts. We indicate in-domain results by bold fonts.}
\label{tab:pixor5x5}
\centering
\begin{tabular}{c|l|ccccc}
\hline
Setting & Source$\backslash$Target & \kitti & \argo & \nusc & \lyft & \waymo \\ \hline
\multirow{5}{*}{Easy} & \kitti & \color{black}{\textbf{87.2}} & \color{blue}{31.5} & \color{black}{39.4} & \color{blue}{65.1} & \color{blue}{28.3} \\
 & \argo & \color{red}{57.4} & \color{black}{\textbf{79.2}} & \color{red}{49.0} & \color{red}{89.4} & \color{black}{69.3} \\
 & \nusc & \color{black}{40.4} & \color{black}{66.3} & \color{black}{\textbf{56.8}} & \color{black}{79.7} & \color{black}{40.7} \\
 & \lyft & \color{black}{53.1} & \color{red}{71.4} & \color{black}{45.5} & \color{black}{\textbf{90.7}} & \color{red}{75.2} \\
 & \waymo & \color{blue}{9.8} & \color{black}{63.6} & \color{blue}{38.1} & \color{black}{82.5} & \color{black}{\textbf{87.2}} \\
\hline\multirow{5}{*}{Moderate} & \kitti & \color{black}{\textbf{72.8}} & \color{blue}{25.9} & \color{blue}{20.0} & \color{blue}{37.4} & \color{blue}{23.4} \\
 & \argo & \color{red}{42.5} & \color{black}{\textbf{67.3}} & \color{black}{24.4} & \color{black}{57.9} & \color{black}{55.0} \\
 & \nusc & \color{black}{30.3} & \color{black}{46.3} & \color{black}{\textbf{30.1}} & \color{black}{50.5} & \color{black}{35.3} \\
 & \lyft & \color{black}{38.7} & \color{red}{58.5} & \color{red}{24.9} & \color{black}{\textbf{78.1}} & \color{red}{56.8} \\
 & \waymo & \color{blue}{10.0} & \color{black}{47.4} & \color{black}{21.0} & \color{red}{62.4} & \color{black}{\textbf{76.4}} \\
\hline\multirow{5}{*}{Hard} & \kitti & \color{black}{\textbf{68.2}} & \color{blue}{28.4} & \color{blue}{18.8} & \color{blue}{34.5} & \color{blue}{24.1} \\
 & \argo & \color{red}{43.0} & \color{black}{\textbf{64.7}} & \color{black}{22.7} & \color{black}{57.7} & \color{red}{55.2} \\
 & \nusc & \color{black}{27.1} & \color{black}{46.0} & \color{black}{\textbf{29.8}} & \color{black}{50.6} & \color{black}{35.8} \\
 & \lyft & \color{black}{35.8} & \color{red}{54.3} & \color{red}{24.8} & \color{black}{\textbf{78.0}} & \color{black}{53.7} \\
 & \waymo & \color{blue}{11.1} & \color{black}{46.9} & \color{black}{21.3} & \color{red}{63.4} & \color{black}{\textbf{74.8}} \\
\hline\multirow{5}{*}{0-30m} & \kitti & \color{black}{\textbf{87.2}} & \color{blue}{39.5} & \color{blue}{38.9} & \color{blue}{62.2} & \color{blue}{32.1} \\
 & \argo & \color{red}{60.0} & \color{black}{\textbf{82.4}} & \color{red}{49.9} & \color{red}{88.0} & \color{black}{72.8} \\
 & \nusc & \color{black}{38.7} & \color{black}{63.0} & \color{black}{\textbf{55.1}} & \color{black}{79.4} & \color{black}{43.9} \\
 & \lyft & \color{black}{50.7} & \color{red}{73.5} & \color{black}{48.4} & \color{black}{\textbf{90.5}} & \color{red}{76.4} \\
 & \waymo & \color{blue}{12.9} & \color{black}{65.7} & \color{black}{42.7} & \color{black}{83.1} & \color{black}{\textbf{88.3}} \\
\hline\multirow{5}{*}{30m-50m} & \kitti & \color{black}{\textbf{50.3}} & \color{blue}{29.4} & \color{red}{9.1} & \color{blue}{31.0} & \color{blue}{26.1} \\
 & \argo & \color{red}{23.7} & \color{black}{\textbf{66.1}} & \color{blue}{0.8} & \color{black}{54.5} & \color{red}{56.5} \\
 & \nusc & \color{black}{18.6} & \color{black}{44.9} & \color{black}{\textbf{12.3}} & \color{black}{48.1} & \color{black}{37.3} \\
 & \lyft & \color{black}{17.5} & \color{red}{50.4} & \color{black}{7.0} & \color{black}{\textbf{77.0}} & \color{black}{53.0} \\
 & \waymo & \color{blue}{8.1} & \color{black}{41.3} & \color{black}{4.5} & \color{red}{62.1} & \color{black}{\textbf{78.0}} \\
\hline\multirow{5}{*}{50m-70m} & \kitti & \color{black}{\textbf{12.0}} & \color{blue}{3.0} & \color{black}{3.0} & \color{blue}{9.0} & \color{blue}{10.1} \\
 & \argo & \color{black}{4.8} & \color{black}{\textbf{31.7}} & \color{blue}{0.3} & \color{black}{20.6} & \color{red}{31.3} \\
 & \nusc & \color{red}{9.1} & \color{black}{13.4} & \color{black}{\textbf{9.1}} & \color{black}{21.6} & \color{black}{20.9} \\
 & \lyft & \color{black}{6.5} & \color{black}{19.1} & \color{black}{9.1} & \color{black}{\textbf{61.2}} & \color{black}{29.9} \\
 & \waymo & \color{blue}{1.7} & \color{red}{20.6} & \color{red}{9.1} & \color{red}{39.7} & \color{black}{\textbf{53.3}} \\
\hline
\end{tabular}
\end{table*}

\begin{table*}[!htbp]
\caption{3D object detection across multiple datasets (evaluated on the validation sets). The setting is exactly the same as~\autoref{tab:pixor5x5}, except that we perform \PRCNN re-sampling on the input point cloud before applying the \PIXOR detector.}
\label{tab:pixor_subsample}
\centering
\begin{tabular}{c|l|ccccc}
\hline
Setting & Source$\backslash$Target & \kitti & \argo & \nusc & \lyft & \waymo \\ \hline
\multirow{5}{*}{Easy} & \kitti & \color{black}{\textbf{85.9}} & \color{blue}{22.3} & \color{blue}{35.7} & \color{blue}{56.2} & \color{blue}{13.4} \\
 & \argo & \color{black}{59.4} & \color{black}{\textbf{80.5}} & \color{black}{47.1} & \color{red}{89.3} & \color{black}{66.5} \\
 & \nusc & \color{blue}{14.5} & \color{black}{57.1} & \color{black}{\textbf{66.2}} & \color{black}{73.4} & \color{black}{44.2} \\
 & \lyft & \color{red}{66.6} & \color{red}{73.8} & \color{red}{52.2} & \color{black}{\textbf{90.7}} & \color{red}{77.3} \\
 & \waymo & \color{black}{28.6} & \color{black}{66.0} & \color{black}{52.0} & \color{black}{84.2} & \color{black}{\textbf{86.7}} \\
\hline\multirow{5}{*}{Moderate} & \kitti & \color{black}{\textbf{70.3}} & \color{blue}{19.1} & \color{blue}{18.9} & \color{blue}{33.5} & \color{blue}{14.8} \\
 & \argo & \color{black}{43.0} & \color{black}{\textbf{66.5}} & \color{black}{24.1} & \color{black}{57.9} & \color{black}{52.6} \\
 & \nusc & \color{blue}{12.6} & \color{black}{46.9} & \color{black}{\textbf{36.5}} & \color{black}{52.6} & \color{black}{35.7} \\
 & \lyft & \color{red}{49.3} & \color{red}{54.4} & \color{red}{28.6} & \color{black}{\textbf{79.4}} & \color{red}{59.2} \\
 & \waymo & \color{black}{23.8} & \color{black}{51.4} & \color{black}{26.7} & \color{red}{69.0} & \color{black}{\textbf{77.1}} \\
\hline\multirow{5}{*}{Hard} & \kitti & \color{black}{\textbf{67.2}} & \color{blue}{20.0} & \color{blue}{17.4} & \color{blue}{33.1} & \color{blue}{15.0} \\
 & \argo & \color{black}{42.8} & \color{black}{\textbf{63.8}} & \color{black}{22.3} & \color{black}{57.7} & \color{black}{52.5} \\
 & \nusc & \color{blue}{14.4} & \color{black}{44.6} & \color{black}{\textbf{35.7}} & \color{black}{53.1} & \color{black}{36.0} \\
 & \lyft & \color{red}{45.5} & \color{red}{54.5} & \color{red}{27.6} & \color{black}{\textbf{79.3}} & \color{red}{58.5} \\
 & \waymo & \color{black}{24.0} & \color{black}{54.2} & \color{black}{26.4} & \color{red}{70.3} & \color{black}{\textbf{77.3}} \\
\hline\multirow{5}{*}{0-30m} & \kitti & \color{black}{\textbf{85.8}} & \color{blue}{28.6} & \color{blue}{33.2} & \color{blue}{56.6} & \color{blue}{14.7} \\
 & \argo & \color{black}{61.5} & \color{black}{\textbf{82.7}} & \color{black}{48.6} & \color{red}{88.3} & \color{black}{65.0} \\
 & \nusc & \color{blue}{20.2} & \color{black}{61.5} & \color{black}{\textbf{64.4}} & \color{black}{75.4} & \color{black}{48.4} \\
 & \lyft & \color{red}{62.9} & \color{red}{71.9} & \color{black}{54.3} & \color{black}{\textbf{90.7}} & \color{red}{78.3} \\
 & \waymo & \color{black}{31.0} & \color{black}{65.9} & \color{red}{55.4} & \color{black}{85.9} & \color{black}{\textbf{88.2}} \\
\hline\multirow{5}{*}{30m-50m} & \kitti & \color{black}{\textbf{48.8}} & \color{blue}{22.0} & \color{black}{4.5} & \color{blue}{29.3} & \color{blue}{16.8} \\
 & \argo & \color{black}{21.1} & \color{black}{\textbf{69.7}} & \color{blue}{2.3} & \color{black}{55.1} & \color{black}{54.6} \\
 & \nusc & \color{blue}{8.6} & \color{black}{42.8} & \color{black}{\textbf{15.9}} & \color{black}{52.7} & \color{black}{40.5} \\
 & \lyft & \color{red}{25.1} & \color{red}{53.1} & \color{red}{10.3} & \color{black}{\textbf{78.6}} & \color{red}{59.9} \\
 & \waymo & \color{black}{16.7} & \color{black}{52.4} & \color{black}{9.8} & \color{red}{68.8} & \color{black}{\textbf{78.8}} \\
\hline\multirow{5}{*}{50m-70m} & \kitti & \color{black}{\textbf{15.7}} & \color{blue}{3.4} & \color{black}{0.4} & \color{blue}{7.5} & \color{blue}{12.0} \\
 & \argo & \color{red}{9.4} & \color{black}{\textbf{29.5}} & \color{blue}{0.1} & \color{black}{22.2} & \color{black}{30.4} \\
 & \nusc & \color{blue}{0.7} & \color{black}{12.8} & \color{black}{\textbf{9.1}} & \color{black}{23.1} & \color{black}{16.4} \\
 & \lyft & \color{black}{7.3} & \color{black}{18.8} & \color{black}{3.0} & \color{black}{\textbf{63.6}} & \color{red}{30.8} \\
 & \waymo & \color{black}{2.3} & \color{red}{23.5} & \color{red}{9.1} & \color{red}{45.1} & \color{black}{\textbf{56.8}} \\
\hline
\end{tabular}
\end{table*}

\section{Additional Results Using \PRCNN}
\label{suppl-sec:tables}

\subsection{Complete tables}

We show the complete tables across five datasets by replacing the predicted box sizes with the ground truth sizes (cf. \autoref{ssec:detector_perf} in the main paper) in~\autoref{big-tab-gt}, by few-shot fine-tuning in~\autoref{big-tab-fs}, by statistical normalization in~\autoref{big-tab-sn}, and by output transformation in~\autoref{big-tab-ot}. For statistical normalization and output transformation, we see smaller improvements (or even some degradation) among datasets collected in the USA than between datasets collected in Germany and the USA. 

\begin{table*}
\caption{Cross-dataset performance by assigning ground-truth box sizes to detected cars while keeping their centers and rotations unchanged. We report \APBEV / \AP of the \emph{Car} category at $\text{IoU}=0.7$, using \PRCNN~\cite{shi2019pointrcnn}. We indicate the best generalization results per column and per setting
by red fonts and the worst by blue fonts. We indicate in-domain results by bold fonts.}
\label{big-tab-gt}
\centering
\begin{tabular}{c|l|ccccc}
\hline
Setting & Source$\backslash$Target & \kitti & \argo & \nusc & \lyft & \waymo \\
\hline
\multirow{5}{*}{Easy} & \kitti & \color{black}{\textbf{95.6}} \color{black}{/} \color{black}{\textbf{84.6}} & \color{red}{80.5} \color{black}{/} \color{black}{65.7} & \color{red}{66.5} \color{black}{/} \color{red}{33.5} & \color{red}{89.8} \color{black}{/} \color{black}{74.8} & \color{red}{90.3} \color{black}{/} \color{red}{77.1} \\
 & \argo & \color{blue}{80.0} \color{black}{/} \color{black}{59.2} & \color{black}{\textbf{83.1}} \color{black}{/} \color{black}{\textbf{77.3}} & \color{blue}{54.2} \color{black}{/} \color{blue}{26.7} & \color{blue}{87.5} \color{black}{/} \color{black}{75.6} & \color{black}{89.1} \color{black}{/} \color{black}{74.7} \\
 & \nusc & \color{black}{80.5} \color{black}{/} \color{black}{63.9} & \color{blue}{77.4} \color{black}{/} \color{blue}{52.5} & \color{black}{\textbf{74.8}} \color{black}{/} \color{black}{\textbf{46.4}} & \color{black}{89.4} \color{black}{/} \color{blue}{65.2} & \color{blue}{85.6} \color{black}{/} \color{blue}{62.9} \\
 & \lyft & \color{black}{83.9} \color{black}{/} \color{blue}{58.4} & \color{black}{80.2} \color{black}{/} \color{black}{67.2} & \color{black}{65.2} \color{black}{/} \color{black}{29.2} & \color{black}{\textbf{90.3}} \color{black}{/} \color{black}{\textbf{87.3}} & \color{black}{89.9} \color{black}{/} \color{black}{73.9} \\
 & \waymo & \color{red}{86.1} \color{black}{/} \color{red}{78.2} & \color{black}{79.2} \color{black}{/} \color{red}{72.7} & \color{black}{63.1} \color{black}{/} \color{black}{30.0} & \color{black}{88.3} \color{black}{/} \color{red}{86.1} & \color{black}{\textbf{90.2}} \color{black}{/} \color{black}{\textbf{86.2}} \\
\hline\multirow{5}{*}{Moderate} & \kitti & \color{black}{\textbf{81.4}} \color{black}{/} \color{black}{\textbf{72.6}} & \color{black}{64.5} \color{black}{/} \color{black}{50.9} & \color{red}{35.0} \color{black}{/} \color{black}{18.2} & \color{black}{74.6} \color{black}{/} \color{black}{54.3} & \color{black}{79.4} \color{black}{/} \color{black}{63.0} \\
 & \argo & \color{black}{66.9} \color{black}{/} \color{black}{51.0} & \color{black}{\textbf{73.6}} \color{black}{/} \color{black}{\textbf{60.1}} & \color{blue}{28.2} \color{black}{/} \color{black}{17.6} & \color{blue}{67.6} \color{black}{/} \color{black}{52.3} & \color{black}{77.3} \color{black}{/} \color{black}{61.5} \\
 & \nusc & \color{blue}{61.4} \color{black}{/} \color{blue}{47.3} & \color{blue}{59.0} \color{black}{/} \color{blue}{36.2} & \color{black}{\textbf{41.7}} \color{black}{/} \color{black}{\textbf{25.4}} & \color{black}{72.4} \color{black}{/} \color{blue}{45.1} & \color{blue}{69.2} \color{black}{/} \color{blue}{50.6} \\
 & \lyft & \color{black}{71.4} \color{black}{/} \color{black}{49.4} & \color{red}{68.5} \color{black}{/} \color{black}{49.3} & \color{black}{34.6} \color{black}{/} \color{blue}{17.4} & \color{black}{\textbf{84.2}} \color{black}{/} \color{black}{\textbf{66.9}} & \color{red}{79.7} \color{black}{/} \color{red}{64.7} \\
 & \waymo & \color{red}{73.7} \color{black}{/} \color{red}{60.6} & \color{black}{68.0} \color{black}{/} \color{red}{54.9} & \color{black}{30.8} \color{black}{/} \color{red}{18.4} & \color{red}{75.0} \color{black}{/} \color{red}{63.2} & \color{black}{\textbf{86.4}} \color{black}{/} \color{black}{\textbf{74.4}} \\
\hline\multirow{5}{*}{Hard} & \kitti & \color{black}{\textbf{82.5}} \color{black}{/} \color{black}{\textbf{71.9}} & \color{black}{64.0} \color{black}{/} \color{black}{49.3} & \color{red}{31.4} \color{black}{/} \color{red}{17.7} & \color{black}{73.1} \color{black}{/} \color{black}{53.0} & \color{black}{77.2} \color{black}{/} \color{black}{59.1} \\
 & \argo & \color{black}{65.6} \color{black}{/} \color{black}{52.5} & \color{black}{\textbf{73.6}} \color{black}{/} \color{black}{\textbf{59.2}} & \color{blue}{27.5} \color{black}{/} \color{black}{16.6} & \color{blue}{65.3} \color{black}{/} \color{black}{52.2} & \color{black}{75.8} \color{black}{/} \color{black}{58.4} \\
 & \nusc & \color{blue}{61.3} \color{black}{/} \color{blue}{45.7} & \color{blue}{55.3} \color{black}{/} \color{blue}{33.5} & \color{black}{\textbf{40.9}} \color{black}{/} \color{black}{\textbf{25.4}} & \color{black}{72.6} \color{black}{/} \color{blue}{43.6} & \color{blue}{68.3} \color{black}{/} \color{blue}{46.2} \\
 & \lyft & \color{black}{72.0} \color{black}{/} \color{black}{52.0} & \color{black}{65.4} \color{black}{/} \color{black}{49.8} & \color{black}{31.2} \color{black}{/} \color{blue}{16.5} & \color{black}{\textbf{84.8}} \color{black}{/} \color{black}{\textbf{67.2}} & \color{red}{78.2} \color{black}{/} \color{red}{63.6} \\
 & \waymo & \color{red}{75.3} \color{black}{/} \color{red}{60.7} & \color{red}{67.8} \color{black}{/} \color{red}{51.9} & \color{black}{30.2} \color{black}{/} \color{black}{17.0} & \color{red}{75.2} \color{black}{/} \color{red}{61.9} & \color{black}{\textbf{80.8}} \color{black}{/} \color{black}{\textbf{68.9}} \\
\hline\hline\multirow{5}{*}{0-30m} & \kitti & \color{black}{\textbf{89.2}} \color{black}{/} \color{black}{\textbf{86.7}} & \color{black}{82.3} \color{black}{/} \color{black}{70.2} & \color{black}{62.8} \color{black}{/} \color{red}{35.1} & \color{red}{89.8} \color{black}{/} \color{black}{76.2} & \color{red}{90.4} \color{black}{/} \color{red}{78.7} \\
 & \argo & \color{black}{83.3} \color{black}{/} \color{black}{68.7} & \color{black}{\textbf{86.1}} \color{black}{/} \color{black}{\textbf{80.2}} & \color{blue}{56.8} \color{black}{/} \color{black}{31.9} & \color{blue}{88.3} \color{black}{/} \color{black}{77.3} & \color{black}{89.8} \color{black}{/} \color{black}{78.1} \\
 & \nusc & \color{blue}{76.5} \color{black}{/} \color{blue}{62.5} & \color{blue}{80.9} \color{black}{/} \color{blue}{55.2} & \color{black}{\textbf{74.2}} \color{black}{/} \color{black}{\textbf{49.1}} & \color{black}{89.4} \color{black}{/} \color{blue}{67.7} & \color{blue}{87.5} \color{black}{/} \color{blue}{62.6} \\
 & \lyft & \color{black}{86.7} \color{black}{/} \color{black}{62.7} & \color{red}{84.2} \color{black}{/} \color{black}{69.3} & \color{red}{63.1} \color{black}{/} \color{blue}{31.7} & \color{black}{\textbf{90.5}} \color{black}{/} \color{black}{\textbf{88.5}} & \color{black}{90.2} \color{black}{/} \color{black}{77.2} \\
 & \waymo & \color{red}{88.0} \color{black}{/} \color{red}{75.8} & \color{black}{82.8} \color{black}{/} \color{red}{76.3} & \color{black}{62.0} \color{black}{/} \color{black}{32.9} & \color{black}{88.7} \color{black}{/} \color{red}{86.8} & \color{black}{\textbf{90.5}} \color{black}{/} \color{black}{\textbf{88.1}} \\
\hline\multirow{5}{*}{30m-50m} & \kitti & \color{black}{\textbf{71.6}} \color{black}{/} \color{black}{\textbf{56.4}} & \color{black}{63.1} \color{black}{/} \color{black}{40.4} & \color{black}{11.1} \color{black}{/} \color{blue}{9.1} & \color{black}{74.3} \color{black}{/} \color{black}{52.9} & \color{red}{80.4} \color{black}{/} \color{black}{64.3} \\
 & \argo & \color{black}{43.2} \color{black}{/} \color{black}{27.0} & \color{black}{\textbf{74.5}} \color{black}{/} \color{black}{\textbf{53.9}} & \color{blue}{9.5} \color{black}{/} \color{black}{9.1} & \color{blue}{67.4} \color{black}{/} \color{black}{49.3} & \color{black}{78.8} \color{black}{/} \color{black}{62.9} \\
 & \nusc & \color{blue}{37.1} \color{black}{/} \color{blue}{25.0} & \color{blue}{49.0} \color{black}{/} \color{blue}{18.3} & \color{black}{\textbf{17.4}} \color{black}{/} \color{black}{\textbf{10.4}} & \color{black}{71.0} \color{black}{/} \color{blue}{42.2} & \color{blue}{75.4} \color{black}{/} \color{blue}{50.5} \\
 & \lyft & \color{black}{52.8} \color{black}{/} \color{black}{31.4} & \color{black}{61.9} \color{black}{/} \color{black}{35.6} & \color{red}{11.3} \color{black}{/} \color{black}{9.1} & \color{black}{\textbf{85.1}} \color{black}{/} \color{black}{\textbf{65.9}} & \color{black}{80.4} \color{black}{/} \color{red}{65.4} \\
 & \waymo & \color{red}{57.6} \color{black}{/} \color{red}{38.5} & \color{red}{63.5} \color{black}{/} \color{red}{45.8} & \color{black}{10.0} \color{black}{/} \color{red}{9.1} & \color{red}{75.4} \color{black}{/} \color{red}{62.5} & \color{black}{\textbf{87.7}} \color{black}{/} \color{black}{\textbf{74.9}} \\
\hline\multirow{5}{*}{50m-70m} & \kitti & \color{black}{\textbf{30.8}} \color{black}{/} \color{black}{\textbf{15.1}} & \color{black}{23.6} \color{black}{/} \color{black}{9.7} & \color{red}{1.6} \color{black}{/} \color{red}{1.0} & \color{black}{50.0} \color{black}{/} \color{black}{24.0} & \color{black}{53.3} \color{black}{/} \color{black}{31.2} \\
 & \argo & \color{black}{13.7} \color{black}{/} \color{red}{10.1} & \color{black}{\textbf{34.2}} \color{black}{/} \color{black}{\textbf{11.7}} & \color{blue}{0.5} \color{black}{/} \color{blue}{0.0} & \color{blue}{37.2} \color{black}{/} \color{blue}{19.8} & \color{black}{51.9} \color{black}{/} \color{black}{31.7} \\
 & \nusc & \color{blue}{9.2} \color{black}{/} \color{black}{5.7} & \color{blue}{15.8} \color{black}{/} \color{blue}{4.3} & \color{black}{\textbf{9.8}} \color{black}{/} \color{black}{\textbf{9.1}} & \color{black}{46.9} \color{black}{/} \color{black}{20.1} & \color{blue}{43.4} \color{black}{/} \color{blue}{23.2} \\
 & \lyft & \color{red}{17.2} \color{black}{/} \color{black}{8.1} & \color{black}{28.2} \color{black}{/} \color{red}{11.4} & \color{black}{1.1} \color{black}{/} \color{black}{0.1} & \color{black}{\textbf{64.2}} \color{black}{/} \color{black}{\textbf{39.9}} & \color{red}{57.8} \color{black}{/} \color{red}{36.3} \\
 & \waymo & \color{black}{13.1} \color{black}{/} \color{blue}{4.9} & \color{red}{29.2} \color{black}{/} \color{black}{11.4} & \color{black}{0.9} \color{black}{/} \color{black}{0.0} & \color{red}{53.0} \color{black}{/} \color{red}{29.4} & \color{black}{\textbf{65.9}} \color{black}{/} \color{black}{\textbf{45.2}} \\
\hline
\end{tabular}
\end{table*}

{\begin{table*}
\caption{Cross-dataset performance by few-shot fine-tuning using $10$ labeled target domain instances (average over five rounds of experiments). We report \APBEV / \AP of the \emph{Car} category at $\text{IoU}=0.7$, using \PRCNN~\cite{shi2019pointrcnn}. We indicate the best generalization results per column and per setting
by red fonts and the worst by blue fonts. We indicate in-domain results by bold fonts.}
\label{big-tab-fs}
\centering
\begin{tabular}{c|l|ccccc}
\hline
Setting & Source$\backslash$Target & \kitti & \argo & \nusc & \lyft & \waymo \\
\hline
\multirow{5}{*}{Easy} & \kitti & \color{black}{\textbf{88.0}} \color{black}{/} \color{black}{\textbf{82.5}} & \color{black}{75.8} \color{black}{/} \color{black}{49.2} & \color{black}{54.7} \color{black}{/} \color{red}{21.7} & \color{red}{89.0} \color{black}{/} \color{black}{78.1} & \color{black}{87.4} \color{black}{/} \color{red}{70.9} \\
 & \argo & \color{blue}{80.0} \color{black}{/} \color{blue}{49.7} & \color{black}{\textbf{74.2}} \color{black}{/} \color{black}{\textbf{42.0}} & \color{black}{54.0} \color{black}{/} \color{black}{19.2} & \color{blue}{86.6} \color{black}{/} \color{blue}{63.5} & \color{black}{86.6} \color{black}{/} \color{blue}{56.3} \\
 & \nusc & \color{black}{83.8} \color{black}{/} \color{black}{58.7} & \color{blue}{68.7} \color{black}{/} \color{blue}{33.7} & \color{black}{\textbf{73.4}} \color{black}{/} \color{black}{\textbf{38.1}} & \color{black}{88.4} \color{black}{/} \color{black}{67.7} & \color{blue}{84.3} \color{black}{/} \color{black}{59.8} \\
 & \lyft & \color{red}{85.3} \color{black}{/} \color{red}{72.5} & \color{black}{73.5} \color{black}{/} \color{black}{48.9} & \color{red}{56.5} \color{black}{/} \color{black}{17.7} & \color{black}{\textbf{90.2}} \color{black}{/} \color{black}{\textbf{87.3}} & \color{red}{89.1} \color{black}{/} \color{black}{70.4} \\
 & \waymo & \color{black}{81.0} \color{black}{/} \color{black}{67.0} & \color{red}{76.9} \color{black}{/} \color{red}{55.2} & \color{blue}{51.0} \color{black}{/} \color{blue}{16.7} & \color{black}{88.3} \color{black}{/} \color{red}{81.0} & \color{black}{\textbf{90.1}} \color{black}{/} \color{black}{\textbf{85.3}} \\
\hline\multirow{5}{*}{Moderate} & \kitti & \color{black}{\textbf{80.6}} \color{black}{/} \color{black}{\textbf{68.9}} & \color{black}{60.7} \color{black}{/} \color{black}{37.3} & \color{black}{28.7} \color{black}{/} \color{red}{12.5} & \color{black}{74.2} \color{black}{/} \color{black}{53.4} & \color{black}{75.9} \color{black}{/} \color{black}{55.3} \\
 & \argo & \color{black}{68.8} \color{black}{/} \color{blue}{42.8} & \color{black}{\textbf{66.5}} \color{black}{/} \color{black}{\textbf{34.4}} & \color{black}{27.5} \color{black}{/} \color{black}{11.2} & \color{blue}{65.4} \color{black}{/} \color{blue}{40.2} & \color{black}{75.3} \color{black}{/} \color{blue}{46.7} \\
 & \nusc & \color{black}{67.2} \color{black}{/} \color{black}{45.5} & \color{blue}{54.5} \color{black}{/} \color{blue}{24.2} & \color{black}{\textbf{40.7}} \color{black}{/} \color{black}{\textbf{21.2}} & \color{black}{71.9} \color{black}{/} \color{black}{44.0} & \color{blue}{72.8} \color{black}{/} \color{black}{47.0} \\
 & \lyft & \color{red}{73.9} \color{black}{/} \color{red}{56.2} & \color{black}{61.0} \color{black}{/} \color{black}{35.3} & \color{red}{30.3} \color{black}{/} \color{blue}{10.6} & \color{black}{\textbf{83.7}} \color{black}{/} \color{black}{\textbf{65.5}} & \color{red}{78.3} \color{black}{/} \color{red}{57.9} \\
 & \waymo & \color{blue}{66.8} \color{black}{/} \color{black}{51.8} & \color{red}{65.7} \color{black}{/} \color{red}{41.8} & \color{blue}{26.7} \color{black}{/} \color{black}{11.0} & \color{red}{75.1} \color{black}{/} \color{red}{54.8} & \color{black}{\textbf{85.9}} \color{black}{/} \color{black}{\textbf{67.9}} \\
\hline\multirow{5}{*}{Hard} & \kitti & \color{black}{\textbf{81.9}} \color{black}{/} \color{black}{\textbf{66.7}} & \color{black}{59.8} \color{black}{/} \color{black}{36.5} & \color{black}{27.5} \color{black}{/} \color{red}{12.4} & \color{black}{71.8} \color{black}{/} \color{black}{52.9} & \color{black}{70.1} \color{black}{/} \color{black}{54.4} \\
 & \argo & \color{black}{66.3} \color{black}{/} \color{blue}{43.0} & \color{black}{\textbf{67.9}} \color{black}{/} \color{black}{\textbf{37.3}} & \color{black}{26.9} \color{black}{/} \color{black}{11.8} & \color{blue}{66.0} \color{black}{/} \color{blue}{42.0} & \color{black}{70.3} \color{black}{/} \color{blue}{43.9} \\
 & \nusc & \color{blue}{64.7} \color{black}{/} \color{black}{44.5} & \color{blue}{52.0} \color{black}{/} \color{blue}{23.4} & \color{black}{\textbf{40.2}} \color{black}{/} \color{black}{\textbf{20.5}} & \color{black}{71.0} \color{black}{/} \color{black}{44.3} & \color{blue}{68.7} \color{black}{/} \color{black}{44.3} \\
 & \lyft & \color{red}{74.1} \color{black}{/} \color{red}{56.2} & \color{black}{61.9} \color{black}{/} \color{black}{37.0} & \color{red}{28.6} \color{black}{/} \color{blue}{11.1} & \color{black}{\textbf{79.3}} \color{black}{/} \color{black}{\textbf{65.5}} & \color{red}{76.9} \color{black}{/} \color{red}{55.6} \\
 & \waymo & \color{black}{68.1} \color{black}{/} \color{black}{52.9} & \color{red}{62.3} \color{black}{/} \color{red}{39.3} & \color{blue}{26.7} \color{black}{/} \color{black}{11.7} & \color{red}{74.7} \color{black}{/} \color{red}{55.2} & \color{black}{\textbf{80.4}} \color{black}{/} \color{black}{\textbf{67.7}} \\
\hline\hline\multirow{5}{*}{0-30m} & \kitti & \color{black}{\textbf{88.8}} \color{black}{/} \color{black}{\textbf{84.9}} & \color{black}{73.6} \color{black}{/} \color{black}{55.2} & \color{black}{54.0} \color{black}{/} \color{red}{23.6} & \color{red}{89.3} \color{black}{/} \color{black}{77.6} & \color{black}{88.7} \color{black}{/} \color{black}{74.1} \\
 & \argo & \color{black}{84.0} \color{black}{/} \color{blue}{56.9} & \color{black}{\textbf{81.2}} \color{black}{/} \color{black}{\textbf{52.2}} & \color{black}{54.0} \color{black}{/} \color{black}{22.6} & \color{blue}{87.7} \color{black}{/} \color{blue}{68.7} & \color{black}{88.3} \color{black}{/} \color{blue}{60.7} \\
 & \nusc & \color{blue}{81.2} \color{black}{/} \color{black}{59.8} & \color{blue}{70.5} \color{black}{/} \color{blue}{40.1} & \color{black}{\textbf{73.2}} \color{black}{/} \color{black}{\textbf{42.8}} & \color{black}{88.8} \color{black}{/} \color{black}{69.6} & \color{blue}{86.2} \color{black}{/} \color{black}{62.4} \\
 & \lyft & \color{red}{87.5} \color{black}{/} \color{red}{73.9} & \color{black}{78.1} \color{black}{/} \color{black}{54.3} & \color{red}{56.9} \color{black}{/} \color{black}{21.2} & \color{black}{\textbf{90.4}} \color{black}{/} \color{black}{\textbf{88.5}} & \color{red}{89.4} \color{black}{/} \color{red}{74.8} \\
 & \waymo & \color{black}{84.8} \color{black}{/} \color{black}{71.0} & \color{red}{79.4} \color{black}{/} \color{red}{56.6} & \color{blue}{52.8} \color{black}{/} \color{blue}{20.8} & \color{black}{88.8} \color{black}{/} \color{red}{79.1} & \color{black}{\textbf{90.4}} \color{black}{/} \color{black}{\textbf{87.2}} \\
\hline\multirow{5}{*}{30m-50m} & \kitti & \color{black}{\textbf{70.2}} \color{black}{/} \color{black}{\textbf{51.4}} & \color{black}{59.0} \color{black}{/} \color{black}{29.9} & \color{red}{9.5} \color{black}{/} \color{black}{6.1} & \color{black}{73.7} \color{black}{/} \color{black}{50.4} & \color{black}{78.1} \color{black}{/} \color{black}{57.2} \\
 & \argo & \color{black}{47.9} \color{black}{/} \color{blue}{23.8} & \color{black}{\textbf{70.8}} \color{black}{/} \color{black}{\textbf{34.0}} & \color{black}{7.3} \color{black}{/} \color{blue}{2.0} & \color{blue}{65.4} \color{black}{/} \color{blue}{36.9} & \color{black}{78.1} \color{black}{/} \color{black}{48.5} \\
 & \nusc & \color{blue}{45.0} \color{black}{/} \color{black}{25.1} & \color{blue}{51.4} \color{black}{/} \color{blue}{17.1} & \color{black}{\textbf{17.1}} \color{black}{/} \color{black}{\textbf{4.1}} & \color{black}{71.5} \color{black}{/} \color{black}{41.5} & \color{blue}{74.2} \color{black}{/} \color{blue}{48.0} \\
 & \lyft & \color{red}{57.7} \color{black}{/} \color{red}{33.3} & \color{red}{62.4} \color{black}{/} \color{black}{29.5} & \color{blue}{6.5} \color{black}{/} \color{black}{3.3} & \color{black}{\textbf{83.8}} \color{black}{/} \color{black}{\textbf{62.7}} & \color{red}{79.7} \color{black}{/} \color{red}{59.9} \\
 & \waymo & \color{black}{49.2} \color{black}{/} \color{black}{29.2} & \color{black}{60.6} \color{black}{/} \color{red}{34.7} & \color{black}{9.4} \color{black}{/} \color{red}{6.3} & \color{red}{75.1} \color{black}{/} \color{red}{52.6} & \color{black}{\textbf{87.5}} \color{black}{/} \color{black}{\textbf{68.8}} \\
\hline\multirow{5}{*}{50m-70m} & \kitti & \color{black}{\textbf{28.8}} \color{black}{/} \color{black}{\textbf{12.0}} & \color{black}{20.1} \color{black}{/} \color{black}{6.3} & \color{red}{3.3} \color{black}{/} \color{red}{1.2} & \color{black}{46.8} \color{black}{/} \color{black}{19.4} & \color{black}{45.2} \color{black}{/} \color{black}{24.3} \\
 & \argo & \color{blue}{8.1} \color{black}{/} \color{blue}{3.8} & \color{black}{\textbf{33.0}} \color{black}{/} \color{black}{\textbf{12.7}} & \color{blue}{0.4} \color{black}{/} \color{blue}{0.0} & \color{blue}{38.0} \color{black}{/} \color{blue}{10.3} & \color{black}{51.1} \color{black}{/} \color{black}{23.4} \\
 & \nusc & \color{black}{12.9} \color{black}{/} \color{black}{5.7} & \color{blue}{15.5} \color{black}{/} \color{blue}{2.6} & \color{black}{\textbf{9.1}} \color{black}{/} \color{black}{\textbf{9.1}} & \color{black}{47.0} \color{black}{/} \color{black}{14.9} & \color{blue}{44.3} \color{black}{/} \color{blue}{19.3} \\
 & \lyft & \color{red}{17.5} \color{black}{/} \color{red}{8.0} & \color{black}{26.8} \color{black}{/} \color{red}{9.1} & \color{black}{2.5} \color{black}{/} \color{black}{0.0} & \color{black}{\textbf{62.7}} \color{black}{/} \color{black}{\textbf{33.1}} & \color{red}{54.0} \color{black}{/} \color{red}{27.2} \\
 & \waymo & \color{black}{10.5} \color{black}{/} \color{black}{4.8} & \color{red}{27.6} \color{black}{/} \color{black}{7.3} & \color{black}{1.3} \color{black}{/} \color{black}{0.0} & \color{red}{51.2} \color{black}{/} \color{red}{19.9} & \color{black}{\textbf{63.5}} \color{black}{/} \color{black}{\textbf{41.1}} \\
\hline
\end{tabular}
\end{table*}}

{\begin{table*}
\caption{Cross-dataset performance by fine-tuning with source data after statistical normalization. We report \APBEV / \AP of the \emph{Car} category at $\text{IoU}=0.7$, using \PRCNN~\cite{shi2019pointrcnn}. We indicate the best generalization results per column and per setting
by red fonts and the worst by blue fonts. We indicate in-domain results by bold fonts.}
\label{big-tab-sn}
\centering
\begin{tabular}{c|l|ccccc}
\hline
Setting & Source$\backslash$Target & \kitti & \argo & \nusc & \lyft & \waymo \\
\hline
\multirow{5}{*}{Easy} & \kitti & \color{black}{\textbf{88.0}} \color{black}{/} \color{black}{\textbf{82.5}} & \color{black}{74.7} \color{black}{/} \color{red}{48.2} & \color{red}{60.8} \color{black}{/} \color{red}{23.9} & \color{black}{88.3} \color{black}{/} \color{black}{73.3} & \color{black}{84.6} \color{black}{/} \color{black}{53.3} \\
 & \argo & \color{blue}{76.2} \color{black}{/} \color{black}{46.1} & \color{black}{\textbf{79.2}} \color{black}{/} \color{black}{\textbf{57.8}} & \color{blue}{48.3} \color{black}{/} \color{blue}{18.6} & \color{blue}{84.8} \color{black}{/} \color{black}{65.0} & \color{black}{84.8} \color{black}{/} \color{black}{49.2} \\
 & \nusc & \color{black}{83.2} \color{black}{/} \color{blue}{35.6} & \color{blue}{72.0} \color{black}{/} \color{blue}{25.3} & \color{black}{\textbf{73.4}} \color{black}{/} \color{black}{\textbf{38.1}} & \color{red}{88.7} \color{black}{/} \color{blue}{38.1} & \color{blue}{76.6} \color{black}{/} \color{blue}{43.3} \\
 & \lyft & \color{red}{83.5} \color{black}{/} \color{red}{72.1} & \color{black}{74.4} \color{black}{/} \color{black}{44.0} & \color{black}{57.8} \color{black}{/} \color{black}{21.1} & \color{black}{\textbf{90.2}} \color{black}{/} \color{black}{\textbf{87.3}} & \color{red}{86.3} \color{black}{/} \color{red}{66.4} \\
 & \waymo & \color{black}{82.1} \color{black}{/} \color{black}{48.7} & \color{red}{75.0} \color{black}{/} \color{black}{44.4} & \color{black}{54.9} \color{black}{/} \color{black}{20.7} & \color{black}{85.7} \color{black}{/} \color{red}{80.0} & \color{black}{\textbf{90.1}} \color{black}{/} \color{black}{\textbf{85.3}} \\
\hline\multirow{5}{*}{Moderate} & \kitti & \color{black}{\textbf{80.6}} \color{black}{/} \color{black}{\textbf{68.9}} & \color{black}{61.5} \color{black}{/} \color{red}{38.2} & \color{red}{32.9} \color{black}{/} \color{red}{16.4} & \color{red}{73.7} \color{black}{/} \color{red}{53.1} & \color{black}{74.9} \color{black}{/} \color{black}{49.4} \\
 & \argo & \color{blue}{67.2} \color{black}{/} \color{black}{40.5} & \color{black}{\textbf{69.9}} \color{black}{/} \color{black}{\textbf{44.2}} & \color{blue}{24.7} \color{black}{/} \color{black}{11.1} & \color{blue}{63.3} \color{black}{/} \color{black}{38.9} & \color{black}{72.0} \color{black}{/} \color{black}{43.6} \\
 & \nusc & \color{black}{67.4} \color{black}{/} \color{blue}{31.0} & \color{blue}{55.6} \color{black}{/} \color{blue}{17.9} & \color{black}{\textbf{40.7}} \color{black}{/} \color{black}{\textbf{21.2}} & \color{black}{71.1} \color{black}{/} \color{blue}{24.5} & \color{blue}{66.6} \color{black}{/} \color{blue}{32.2} \\
 & \lyft & \color{red}{73.6} \color{black}{/} \color{red}{57.9} & \color{black}{59.7} \color{black}{/} \color{black}{33.3} & \color{black}{30.4} \color{black}{/} \color{blue}{10.9} & \color{black}{\textbf{83.7}} \color{black}{/} \color{black}{\textbf{65.5}} & \color{red}{75.5} \color{black}{/} \color{red}{51.3} \\
 & \waymo & \color{black}{71.3} \color{black}{/} \color{black}{47.1} & \color{red}{62.3} \color{black}{/} \color{black}{31.7} & \color{black}{28.8} \color{black}{/} \color{black}{11.5} & \color{black}{71.5} \color{black}{/} \color{black}{52.6} & \color{black}{\textbf{85.9}} \color{black}{/} \color{black}{\textbf{67.9}} \\
\hline\multirow{5}{*}{Hard} & \kitti & \color{black}{\textbf{81.9}} \color{black}{/} \color{black}{\textbf{66.7}} & \color{black}{60.6} \color{black}{/} \color{red}{37.1} & \color{red}{31.9} \color{black}{/} \color{red}{15.8} & \color{red}{73.1} \color{black}{/} \color{red}{53.5} & \color{black}{69.4} \color{black}{/} \color{black}{49.4} \\
 & \argo & \color{black}{68.5} \color{black}{/} \color{black}{41.9} & \color{black}{\textbf{69.9}} \color{black}{/} \color{black}{\textbf{42.8}} & \color{blue}{24.3} \color{black}{/} \color{black}{10.9} & \color{blue}{61.6} \color{black}{/} \color{black}{40.2} & \color{black}{68.2} \color{black}{/} \color{black}{42.7} \\
 & \nusc & \color{blue}{65.2} \color{black}{/} \color{blue}{30.8} & \color{blue}{52.5} \color{black}{/} \color{blue}{17.2} & \color{black}{\textbf{40.2}} \color{black}{/} \color{black}{\textbf{20.5}} & \color{black}{67.3} \color{black}{/} \color{blue}{28.6} & \color{blue}{65.7} \color{black}{/} \color{blue}{30.4} \\
 & \lyft & \color{red}{75.2} \color{black}{/} \color{red}{58.9} & \color{red}{60.8} \color{black}{/} \color{black}{31.8} & \color{black}{29.5} \color{black}{/} \color{black}{14.4} & \color{black}{\textbf{79.3}} \color{black}{/} \color{black}{\textbf{65.5}} & \color{red}{75.5} \color{black}{/} \color{red}{53.2} \\
 & \waymo & \color{black}{73.0} \color{black}{/} \color{black}{49.7} & \color{black}{60.2} \color{black}{/} \color{black}{32.5} & \color{black}{28.4} \color{black}{/} \color{blue}{10.9} & \color{black}{71.6} \color{black}{/} \color{black}{53.3} & \color{black}{\textbf{80.4}} \color{black}{/} \color{black}{\textbf{67.7}} \\
\hline\hline\multirow{5}{*}{0-30m} & \kitti & \color{black}{\textbf{88.8}} \color{black}{/} \color{black}{\textbf{84.9}} & \color{black}{73.1} \color{black}{/} \color{red}{54.2} & \color{red}{60.0} \color{black}{/} \color{red}{29.2} & \color{black}{88.8} \color{black}{/} \color{black}{75.4} & \color{black}{87.1} \color{black}{/} \color{black}{60.1} \\
 & \argo & \color{blue}{83.3} \color{black}{/} \color{black}{53.9} & \color{black}{\textbf{83.3}} \color{black}{/} \color{black}{\textbf{63.3}} & \color{blue}{51.5} \color{black}{/} \color{blue}{23.0} & \color{blue}{86.3} \color{black}{/} \color{black}{68.4} & \color{black}{87.3} \color{black}{/} \color{black}{59.7} \\
 & \nusc & \color{black}{83.6} \color{black}{/} \color{blue}{42.8} & \color{blue}{72.8} \color{black}{/} \color{blue}{27.2} & \color{black}{\textbf{73.2}} \color{black}{/} \color{black}{\textbf{42.8}} & \color{red}{88.9} \color{black}{/} \color{blue}{47.1} & \color{blue}{78.5} \color{black}{/} \color{blue}{45.9} \\
 & \lyft & \color{red}{87.4} \color{black}{/} \color{red}{73.6} & \color{black}{78.7} \color{black}{/} \color{black}{51.8} & \color{black}{58.7} \color{black}{/} \color{black}{26.8} & \color{black}{\textbf{90.4}} \color{black}{/} \color{black}{\textbf{88.5}} & \color{red}{87.9} \color{black}{/} \color{red}{72.4} \\
 & \waymo & \color{black}{85.7} \color{black}{/} \color{black}{59.0} & \color{red}{79.9} \color{black}{/} \color{black}{50.5} & \color{black}{57.6} \color{black}{/} \color{black}{24.3} & \color{black}{87.2} \color{black}{/} \color{red}{75.8} & \color{black}{\textbf{90.4}} \color{black}{/} \color{black}{\textbf{87.2}} \\
\hline\multirow{5}{*}{30m-50m} & \kitti & \color{black}{\textbf{70.2}} \color{black}{/} \color{black}{\textbf{51.4}} & \color{red}{61.5} \color{black}{/} \color{red}{31.5} & \color{red}{11.0} \color{black}{/} \color{black}{2.3} & \color{red}{73.8} \color{black}{/} \color{red}{52.2} & \color{red}{78.1} \color{black}{/} \color{red}{54.9} \\
 & \argo & \color{black}{48.9} \color{black}{/} \color{black}{25.7} & \color{black}{\textbf{72.2}} \color{black}{/} \color{black}{\textbf{39.5}} & \color{blue}{5.0} \color{black}{/} \color{black}{4.5} & \color{blue}{61.0} \color{black}{/} \color{black}{32.4} & \color{black}{74.4} \color{black}{/} \color{black}{46.2} \\
 & \nusc & \color{blue}{44.9} \color{black}{/} \color{blue}{18.6} & \color{blue}{45.6} \color{black}{/} \color{blue}{7.3} & \color{black}{\textbf{17.1}} \color{black}{/} \color{black}{\textbf{4.1}} & \color{black}{70.1} \color{black}{/} \color{blue}{18.1} & \color{blue}{67.9} \color{black}{/} \color{blue}{31.6} \\
 & \lyft & \color{red}{58.3} \color{black}{/} \color{red}{38.0} & \color{black}{57.2} \color{black}{/} \color{black}{18.5} & \color{black}{6.5} \color{black}{/} \color{red}{4.5} & \color{black}{\textbf{83.8}} \color{black}{/} \color{black}{\textbf{62.7}} & \color{black}{77.2} \color{black}{/} \color{black}{52.4} \\
 & \waymo & \color{black}{57.3} \color{black}{/} \color{black}{36.3} & \color{black}{54.9} \color{black}{/} \color{black}{20.1} & \color{black}{9.1} \color{black}{/} \color{blue}{1.5} & \color{black}{71.3} \color{black}{/} \color{black}{48.4} & \color{black}{\textbf{87.5}} \color{black}{/} \color{black}{\textbf{68.8}} \\
\hline\multirow{5}{*}{50m-70m} & \kitti & \color{black}{\textbf{28.8}} \color{black}{/} \color{black}{\textbf{12.0}} & \color{black}{23.8} \color{black}{/} \color{black}{5.6} & \color{black}{3.0} \color{black}{/} \color{red}{2.3} & \color{red}{49.9} \color{black}{/} \color{red}{22.2} & \color{black}{46.8} \color{black}{/} \color{black}{25.1} \\
 & \argo & \color{blue}{9.1} \color{black}{/} \color{blue}{2.6} & \color{black}{\textbf{29.9}} \color{black}{/} \color{black}{\textbf{6.9}} & \color{blue}{0.2} \color{black}{/} \color{black}{0.1} & \color{blue}{28.9} \color{black}{/} \color{black}{8.8} & \color{black}{46.2} \color{black}{/} \color{black}{21.2} \\
 & \nusc & \color{black}{9.4} \color{black}{/} \color{black}{5.1} & \color{blue}{14.8} \color{black}{/} \color{blue}{2.3} & \color{black}{\textbf{9.1}} \color{black}{/} \color{black}{\textbf{9.1}} & \color{black}{40.7} \color{black}{/} \color{blue}{5.2} & \color{blue}{36.4} \color{black}{/} \color{blue}{14.9} \\
 & \lyft & \color{red}{21.1} \color{black}{/} \color{red}{6.7} & \color{black}{21.2} \color{black}{/} \color{black}{4.9} & \color{red}{4.5} \color{black}{/} \color{black}{0.0} & \color{black}{\textbf{62.7}} \color{black}{/} \color{black}{\textbf{33.1}} & \color{red}{52.1} \color{black}{/} \color{red}{25.3} \\
 & \waymo & \color{black}{14.4} \color{black}{/} \color{black}{5.7} & \color{red}{27.7} \color{black}{/} \color{red}{11.0} & \color{black}{1.0} \color{black}{/} \color{blue}{0.0} & \color{black}{46.9} \color{black}{/} \color{black}{22.0} & \color{black}{\textbf{63.5}} \color{black}{/} \color{black}{\textbf{41.1}} \\
\hline
\end{tabular}
\end{table*}}

\begin{table*}
\caption{Cross-dataset performance by output transformation: directly adjusting the predicted box size by adding the difference of mean sizes between domains. We report \APBEV / \AP of the \emph{Car} category at $\text{IoU}=0.7$, using \PRCNN~\cite{shi2019pointrcnn}. We indicate the best generalization results per column and per setting
by red fonts and the worst by blue fonts. We indicate in-domain results by bold fonts.}
\label{big-tab-ot}
\centering
\begin{tabular}{c|l|ccccc}
\hline
Setting & Source$\backslash$Target & \kitti & \argo & \nusc & \lyft & \waymo \\
\hline
\multirow{5}{*}{Easy} & \kitti & \color{black}{\textbf{88.0}} \color{black}{/} \color{black}{\textbf{82.5}} & \color{blue}{72.7} \color{black}{/} \color{blue}{9.0} & \color{black}{55.0} \color{black}{/} \color{blue}{10.4} & \color{black}{88.2} \color{black}{/} \color{blue}{23.5} & \color{black}{86.1} \color{black}{/} \color{blue}{16.2} \\
 & \argo & \color{blue}{53.3} \color{black}{/} \color{black}{5.7} & \color{black}{\textbf{79.2}} \color{black}{/} \color{black}{\textbf{57.8}} & \color{blue}{52.6} \color{black}{/} \color{black}{21.3} & \color{blue}{87.1} \color{black}{/} \color{black}{66.1} & \color{black}{87.6} \color{black}{/} \color{black}{56.1} \\
 & \nusc & \color{red}{75.4} \color{black}{/} \color{red}{31.5} & \color{black}{73.3} \color{black}{/} \color{black}{27.9} & \color{black}{\textbf{73.4}} \color{black}{/} \color{black}{\textbf{38.1}} & \color{red}{89.2} \color{black}{/} \color{black}{44.3} & \color{blue}{78.4} \color{black}{/} \color{black}{35.5} \\
 & \lyft & \color{black}{71.9} \color{black}{/} \color{black}{4.7} & \color{red}{77.1} \color{black}{/} \color{black}{48.0} & \color{red}{63.1} \color{black}{/} \color{black}{24.5} & \color{black}{\textbf{90.2}} \color{black}{/} \color{black}{\textbf{87.3}} & \color{red}{89.2} \color{black}{/} \color{red}{73.9} \\
 & \waymo & \color{black}{64.0} \color{black}{/} \color{blue}{3.9} & \color{black}{74.3} \color{black}{/} \color{red}{54.8} & \color{black}{58.8} \color{black}{/} \color{red}{25.2} & \color{black}{88.3} \color{black}{/} \color{red}{85.3} & \color{black}{\textbf{90.1}} \color{black}{/} \color{black}{\textbf{85.3}} \\
\hline\multirow{5}{*}{Moderate} & \kitti & \color{black}{\textbf{80.6}} \color{black}{/} \color{black}{\textbf{68.9}} & \color{black}{59.9} \color{black}{/} \color{blue}{7.9} & \color{black}{30.8} \color{black}{/} \color{blue}{6.8} & \color{black}{70.1} \color{black}{/} \color{blue}{17.8} & \color{black}{69.1} \color{black}{/} \color{blue}{13.1} \\
 & \argo & \color{blue}{52.2} \color{black}{/} \color{black}{7.3} & \color{black}{\textbf{69.9}} \color{black}{/} \color{black}{\textbf{44.2}} & \color{blue}{27.5} \color{black}{/} \color{black}{11.7} & \color{blue}{66.9} \color{black}{/} \color{black}{42.1} & \color{black}{74.3} \color{black}{/} \color{black}{45.5} \\
 & \nusc & \color{black}{58.5} \color{black}{/} \color{red}{27.3} & \color{blue}{56.8} \color{black}{/} \color{black}{20.4} & \color{black}{\textbf{40.7}} \color{black}{/} \color{black}{\textbf{21.2}} & \color{black}{71.3} \color{black}{/} \color{black}{27.3} & \color{blue}{67.8} \color{black}{/} \color{black}{26.2} \\
 & \lyft & \color{red}{60.8} \color{black}{/} \color{black}{5.6} & \color{black}{62.7} \color{black}{/} \color{black}{37.6} & \color{red}{33.5} \color{black}{/} \color{black}{12.5} & \color{black}{\textbf{83.7}} \color{black}{/} \color{black}{\textbf{65.5}} & \color{red}{78.4} \color{black}{/} \color{red}{60.8} \\
 & \waymo & \color{black}{54.9} \color{black}{/} \color{blue}{3.7} & \color{red}{62.9} \color{black}{/} \color{red}{40.4} & \color{black}{30.1} \color{black}{/} \color{red}{14.5} & \color{red}{74.3} \color{black}{/} \color{red}{59.8} & \color{black}{\textbf{85.9}} \color{black}{/} \color{black}{\textbf{67.9}} \\
\hline\multirow{5}{*}{Hard} & \kitti & \color{black}{\textbf{81.9}} \color{black}{/} \color{black}{\textbf{66.7}} & \color{black}{59.3} \color{black}{/} \color{blue}{9.3} & \color{black}{27.8} \color{black}{/} \color{blue}{7.6} & \color{black}{66.5} \color{black}{/} \color{blue}{19.1} & \color{black}{68.7} \color{black}{/} \color{blue}{13.9} \\
 & \argo & \color{blue}{53.5} \color{black}{/} \color{black}{8.6} & \color{black}{\textbf{69.9}} \color{black}{/} \color{black}{\textbf{42.8}} & \color{blue}{26.7} \color{black}{/} \color{black}{14.5} & \color{blue}{64.6} \color{black}{/} \color{black}{43.0} & \color{black}{70.0} \color{black}{/} \color{black}{44.2} \\
 & \nusc & \color{black}{59.5} \color{black}{/} \color{red}{27.8} & \color{blue}{53.6} \color{black}{/} \color{black}{19.9} & \color{black}{\textbf{40.2}} \color{black}{/} \color{black}{\textbf{20.5}} & \color{black}{67.6} \color{black}{/} \color{black}{28.5} & \color{blue}{66.3} \color{black}{/} \color{black}{26.0} \\
 & \lyft & \color{red}{63.1} \color{black}{/} \color{black}{6.9} & \color{red}{63.4} \color{black}{/} \color{black}{38.6} & \color{red}{30.4} \color{black}{/} \color{black}{13.3} & \color{black}{\textbf{79.3}} \color{black}{/} \color{black}{\textbf{65.5}} & \color{red}{77.3} \color{black}{/} \color{red}{57.3} \\
 & \waymo & \color{black}{58.0} \color{black}{/} \color{blue}{4.1} & \color{black}{60.5} \color{black}{/} \color{red}{39.2} & \color{black}{29.4} \color{black}{/} \color{red}{14.6} & \color{red}{74.0} \color{black}{/} \color{red}{57.2} & \color{black}{\textbf{80.4}} \color{black}{/} \color{black}{\textbf{67.7}} \\
\hline\hline\multirow{5}{*}{0-30m} & \kitti & \color{black}{\textbf{88.8}} \color{black}{/} \color{black}{\textbf{84.9}} & \color{blue}{73.0} \color{black}{/} \color{blue}{13.7} & \color{black}{56.2} \color{black}{/} \color{blue}{13.9} & \color{black}{88.4} \color{black}{/} \color{blue}{27.5} & \color{black}{87.7} \color{black}{/} \color{blue}{22.2} \\
 & \argo & \color{blue}{64.9} \color{black}{/} \color{black}{10.1} & \color{black}{\textbf{83.3}} \color{black}{/} \color{black}{\textbf{63.3}} & \color{blue}{55.2} \color{black}{/} \color{black}{27.0} & \color{blue}{87.8} \color{black}{/} \color{black}{69.9} & \color{black}{87.9} \color{black}{/} \color{black}{62.6} \\
 & \nusc & \color{black}{74.6} \color{black}{/} \color{red}{36.6} & \color{black}{73.7} \color{black}{/} \color{black}{32.0} & \color{black}{\textbf{73.2}} \color{black}{/} \color{black}{\textbf{42.8}} & \color{red}{89.2} \color{black}{/} \color{black}{46.2} & \color{blue}{79.6} \color{black}{/} \color{black}{41.6} \\
 & \lyft & \color{red}{74.8} \color{black}{/} \color{black}{9.1} & \color{red}{81.2} \color{black}{/} \color{red}{55.8} & \color{red}{61.2} \color{black}{/} \color{red}{27.2} & \color{black}{\textbf{90.4}} \color{black}{/} \color{black}{\textbf{88.5}} & \color{red}{89.6} \color{black}{/} \color{red}{77.2} \\
 & \waymo & \color{black}{71.3} \color{black}{/} \color{blue}{4.4} & \color{black}{78.4} \color{black}{/} \color{black}{55.7} & \color{black}{60.5} \color{black}{/} \color{black}{25.8} & \color{black}{88.7} \color{black}{/} \color{red}{85.0} & \color{black}{\textbf{90.4}} \color{black}{/} \color{black}{\textbf{87.2}} \\
\hline\multirow{5}{*}{30m-50m} & \kitti & \color{black}{\textbf{70.2}} \color{black}{/} \color{black}{\textbf{51.4}} & \color{black}{56.1} \color{black}{/} \color{blue}{5.4} & \color{black}{10.8} \color{black}{/} \color{black}{9.1} & \color{black}{67.4} \color{black}{/} \color{blue}{10.7} & \color{black}{73.6} \color{black}{/} \color{blue}{10.4} \\
 & \argo & \color{blue}{35.1} \color{black}{/} \color{black}{9.1} & \color{black}{\textbf{72.2}} \color{black}{/} \color{black}{\textbf{39.5}} & \color{blue}{9.5} \color{black}{/} \color{blue}{0.3} & \color{blue}{66.3} \color{black}{/} \color{black}{39.1} & \color{black}{77.5} \color{black}{/} \color{black}{44.9} \\
 & \nusc & \color{black}{35.5} \color{black}{/} \color{red}{15.5} & \color{blue}{47.4} \color{black}{/} \color{black}{7.8} & \color{black}{\textbf{17.1}} \color{black}{/} \color{black}{\textbf{4.1}} & \color{black}{69.9} \color{black}{/} \color{black}{22.5} & \color{blue}{68.7} \color{black}{/} \color{black}{21.1} \\
 & \lyft & \color{red}{43.3} \color{black}{/} \color{blue}{3.9} & \color{red}{60.8} \color{black}{/} \color{black}{25.4} & \color{red}{11.2} \color{black}{/} \color{black}{9.1} & \color{black}{\textbf{83.8}} \color{black}{/} \color{black}{\textbf{62.7}} & \color{red}{79.5} \color{black}{/} \color{red}{61.4} \\
 & \waymo & \color{black}{39.8} \color{black}{/} \color{black}{4.5} & \color{black}{58.1} \color{black}{/} \color{red}{34.9} & \color{black}{9.9} \color{black}{/} \color{red}{9.1} & \color{red}{74.5} \color{black}{/} \color{red}{57.5} & \color{black}{\textbf{87.5}} \color{black}{/} \color{black}{\textbf{68.8}} \\
\hline\multirow{5}{*}{50m-70m} & \kitti & \color{black}{\textbf{28.8}} \color{black}{/} \color{black}{\textbf{12.0}} & \color{black}{20.5} \color{black}{/} \color{blue}{1.0} & \color{red}{1.5} \color{black}{/} \color{red}{1.0} & \color{black}{41.3} \color{black}{/} \color{black}{6.8} & \color{black}{42.6} \color{black}{/} \color{blue}{4.2} \\
 & \argo & \color{black}{8.0} \color{black}{/} \color{blue}{0.8} & \color{black}{\textbf{29.9}} \color{black}{/} \color{black}{\textbf{6.9}} & \color{blue}{0.5} \color{black}{/} \color{blue}{0.0} & \color{blue}{35.6} \color{black}{/} \color{black}{14.2} & \color{black}{49.2} \color{black}{/} \color{black}{20.3} \\
 & \nusc & \color{black}{7.8} \color{black}{/} \color{red}{5.1} & \color{blue}{15.3} \color{black}{/} \color{black}{3.0} & \color{black}{\textbf{9.1}} \color{black}{/} \color{black}{\textbf{9.1}} & \color{black}{41.4} \color{black}{/} \color{blue}{5.6} & \color{blue}{37.0} \color{black}{/} \color{black}{12.0} \\
 & \lyft & \color{red}{12.7} \color{black}{/} \color{black}{0.9} & \color{red}{25.6} \color{black}{/} \color{black}{6.0} & \color{black}{1.1} \color{black}{/} \color{black}{0.0} & \color{black}{\textbf{62.7}} \color{black}{/} \color{black}{\textbf{33.1}} & \color{red}{54.9} \color{black}{/} \color{red}{30.4} \\
 & \waymo & \color{blue}{7.7} \color{black}{/} \color{black}{1.1} & \color{black}{25.5} \color{black}{/} \color{red}{6.5} & \color{black}{0.9} \color{black}{/} \color{black}{0.0} & \color{red}{50.8} \color{black}{/} \color{red}{22.3} & \color{black}{\textbf{63.5}} \color{black}{/} \color{black}{\textbf{41.1}} \\
\hline
\end{tabular}
\end{table*}

\subsection{Online sales data} In the main paper, for statistical normalization we leverage the average car size of each dataset.
Here we collect car sales data from Germany and the USA in the past four years. The average car size $(h, w, l)$ is $(1.75, 1.93, 5.15)$ in the USA  and $(1.49, 1.79, 4.40)$ in Germany. The difference is $(0.26, 0.14, 0.75)$, not far from $(0.20, 0.37, 0.78)$ between \kitti and the other datasets. The gap can be reduced by further  considering locations (e.g., \argo from Miami and Pittsburgh, USA) and earlier data (\kitti was collected in 2011).

In~\autoref{tab:web}, we show the results of adapting a detector trained on \kitti to other datasets using statistical normalization with the car sales data: $(\Delta h, \Delta w, \Delta l)$ is $(0.26, 0.15, 0.75)$. The performance is slightly worse than using the statistics of the datasets. Nevertheless, compared to directly applying the source domain detector, statistical normalization with the car sales data still shows notable improvements.

\begin{table}
\caption{Statistical normalization using the mean sizes of datasets versus car sales data. Direct: directly applying the source domain detector.}
\vskip-5pt
\label{tab:web}
\centering
\tabcolsep 2pt
\begin{tabular}{c|l|c|c|c}
& & \multicolumn{3}{c}{From \kitti (\kitti as the source)} \\ 
\hline
Setting & Dataset & Direct & Datasets & Car sales data \\
\hline
\multirow{4}{*}{Easy} & \argo & \small{\color{black}{55.8} \color{black}{/} \color{black}{27.7}} & \small{\color{black}{74.7} \color{black}{/} \color{black}{48.2}} & \small{\color{black}{68.6} \color{black}{/} \color{black}{32.8}} \\
 & \nusc & \small{\color{black}{47.4} \color{black}{/} \color{black}{13.3}} & \small{\color{black}{60.8} \color{black}{/} \color{black}{23.9}} & \small{\color{black}{62.0} \color{black}{/} \color{black}{24.4}} \\
 & \lyft & \small{\color{black}{81.7} \color{black}{/} \color{black}{51.8}} & \small{\color{black}{88.3} \color{black}{/} \color{black}{73.3}} & \small{\color{black}{88.9} \color{black}{/} \color{black}{69.9}} \\
 & \waymo & \small{\color{black}{45.2} \color{black}{/} \color{black}{11.9}} & \small{\color{black}{84.6} \color{black}{/} \color{black}{53.3}} & \small{\color{black}{66.7} \color{black}{/} \color{black}{22.8}} \\
\hline\multirow{4}{*}{Moderate} & \argo & \small{\color{black}{44.9} \color{black}{/} \color{black}{22.3}} & \small{\color{black}{61.5} \color{black}{/} \color{black}{38.2}} & \small{\color{black}{57.7} \color{black}{/} \color{black}{29.1}} \\
 & \nusc & \small{\color{black}{26.2} \color{black}{/} \color{black}{8.3}} & \small{\color{black}{32.9} \color{black}{/} \color{black}{16.4}} & \small{\color{black}{32.6} \color{black}{/} \color{black}{13.0}} \\
 & \lyft & \small{\color{black}{61.8} \color{black}{/} \color{black}{33.7}} & \small{\color{black}{73.7} \color{black}{/} \color{black}{53.1}} & \small{\color{black}{72.6} \color{black}{/} \color{black}{47.6}} \\
 & \waymo & \small{\color{black}{43.9} \color{black}{/} \color{black}{12.3}} & \small{\color{black}{74.9} \color{black}{/} \color{black}{49.4}} & \small{\color{black}{61.8} \color{black}{/} \color{black}{22.9}} \\
\hline\multirow{4}{*}{Hard} & \argo & \small{\color{black}{42.5} \color{black}{/} \color{black}{22.2}} & \small{\color{black}{60.6} \color{black}{/} \color{black}{37.1}} & \small{\color{black}{54.0} \color{black}{/} \color{black}{30.0}} \\
 & \nusc & \small{\color{black}{24.9} \color{black}{/} \color{black}{8.8}} & \small{\color{black}{31.9} \color{black}{/} \color{black}{15.8}} & \small{\color{black}{29.8} \color{black}{/} \color{black}{13.2}} \\
 & \lyft & \small{\color{black}{57.4} \color{black}{/} \color{black}{34.2}} & \small{\color{black}{73.1} \color{black}{/} \color{black}{53.5}} & \small{\color{black}{71.7} \color{black}{/} \color{black}{45.7}} \\
 & \waymo & \small{\color{black}{41.5} \color{black}{/} \color{black}{12.6}} & \small{\color{black}{69.4} \color{black}{/} \color{black}{49.4}} & \small{\color{black}{62.7} \color{black}{/} \color{black}{25.1}} \\
\hline\hline\multirow{4}{*}{0-30m} & \argo & \small{\color{black}{58.4} \color{black}{/} \color{black}{34.7}} & \small{\color{black}{73.1} \color{black}{/} \color{black}{54.2}} & \small{\color{black}{71.0} \color{black}{/} \color{black}{44.0}} \\
 & \nusc & \small{\color{black}{47.9} \color{black}{/} \color{black}{14.9}} & \small{\color{black}{60.0} \color{black}{/} \color{black}{29.2}} & \small{\color{black}{60.1} \color{black}{/} \color{black}{26.1}} \\
 & \lyft & \small{\color{black}{77.8} \color{black}{/} \color{black}{54.2}} & \small{\color{black}{88.8} \color{black}{/} \color{black}{75.4}} & \small{\color{black}{89.2} \color{black}{/} \color{black}{72.5}} \\
 & \waymo & \small{\color{black}{48.0} \color{black}{/} \color{black}{14.0}} & \small{\color{black}{87.1} \color{black}{/} \color{black}{60.1}} & \small{\color{black}{72.4} \color{black}{/} \color{black}{30.2}} \\
\hline\multirow{4}{*}{30m-50m} & \argo & \small{\color{black}{46.5} \color{black}{/} \color{black}{19.0}} & \small{\color{black}{61.5} \color{black}{/} \color{black}{31.5}} & \small{\color{black}{57.4} \color{black}{/} \color{black}{20.0}} \\
 & \nusc & \small{\color{black}{9.8} \color{black}{/} \color{black}{4.5}} & \small{\color{black}{11.0} \color{black}{/} \color{black}{2.3}} & \small{\color{black}{5.7} \color{black}{/} \color{black}{3.0}} \\
 & \lyft & \small{\color{black}{60.1} \color{black}{/} \color{black}{34.5}} & \small{\color{black}{73.8} \color{black}{/} \color{black}{52.2}} & \small{\color{black}{72.2} \color{black}{/} \color{black}{42.7}} \\
 & \waymo & \small{\color{black}{50.5} \color{black}{/} \color{black}{21.4}} & \small{\color{black}{78.1} \color{black}{/} \color{black}{54.9}} & \small{\color{black}{66.8} \color{black}{/} \color{black}{35.5}} \\
\hline\multirow{4}{*}{50m-70m} & \argo & \small{\color{black}{9.2} \color{black}{/} \color{black}{3.0}} & \small{\color{black}{23.8} \color{black}{/} \color{black}{5.6}} & \small{\color{black}{16.8} \color{black}{/} \color{black}{4.5}} \\
 & \nusc & \small{\color{black}{1.1} \color{black}{/} \color{black}{0.0}} & \small{\color{black}{3.0} \color{black}{/} \color{black}{2.3}} & \small{\color{black}{1.0} \color{black}{/} \color{black}{0.1}} \\
 & \lyft & \small{\color{black}{33.2} \color{black}{/} \color{black}{9.6}} & \small{\color{black}{49.9} \color{black}{/} \color{black}{22.2}} & \small{\color{black}{46.0} \color{black}{/} \color{black}{18.8}} \\
 & \waymo & \small{\color{black}{27.1} \color{black}{/} \color{black}{12.0}} & \small{\color{black}{46.8} \color{black}{/} \color{black}{25.1}} & \small{\color{black}{44.2} \color{black}{/} \color{black}{18.0}} \\
\hline
\end{tabular}
\end{table}

\subsection{Pedestrian}
We calculate the statistics of pedestrians, as in \autoref{tab:ped}. There are smaller differences among datasets. We therefore expect a smaller improvement by statistical normalization.

\begin{table}[t]
    \centering
    \small
    \tabcolsep 2.5pt
    \caption{Dataset statistics on pedestrians (meters)}
    \begin{tabular}{c|c|c|c|c|c}
         & KITTI & Argoverse & nuScenes & Lyft & Waymo \\ \hline
         H & 1.76$\pm$0.11 & 1.84$\pm$0.15 & 1.78$\pm$0.18 & 1.76$\pm$0.18 & 1.75$\pm$0.20 \\\hline
         W & 0.66$\pm$0.14 & 0.78$\pm$0.14& 0.67$\pm$0.14 & 0.76$\pm$0.14 & 0.85$\pm$0.15 \\\hline
         L & 0.84$\pm$0.23 & 0.78$\pm$0.14 & 0.73$\pm$0.19 & 0.78$\pm$0.17& 0.90$\pm$0.19 \\\hline
    \end{tabular}
    \label{tab:ped}
\end{table}
\section{Qualitative Results}
\label{suppl-sec:Quali}
\begin{figure*}[htbp]
	\centering
	\includegraphics[width=\linewidth]{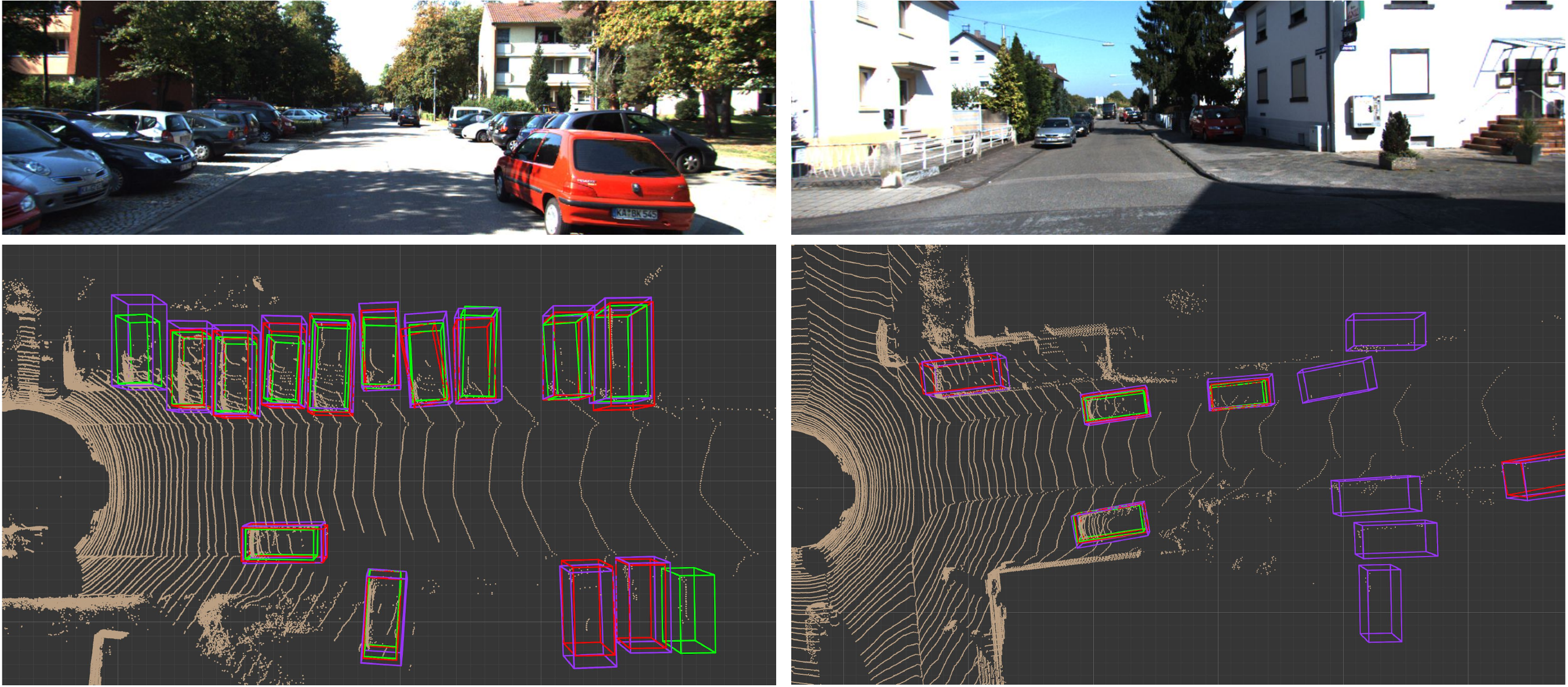}
	\vskip -5pt
	\caption{3D prediction on \kitti using the model trained on \waymo before and after statistical normalization refinement. The green boxes are the ground truth \emph{car} bounding boxes. Purple boxes and magenta boxes are predictions by the model before and after statistical normalization refinement, respectively. The left column demonstrates that statistical normalization is able to resize bounding box predictions to the correct sizes, while the right case shows that it also can reduce false positive rates.}
	\label{fig:qual}
	\vskip -5pt
\end{figure*}

We further show qualitative results of statistical normalization refinement. We train a \PRCNN detector on \waymo and test it on \kitti. We compare its car detection before and after statistical normalization refinement in \autoref{fig:qual}. Statistical normalization can not only improve the predicted bounding box sizes, but also reduce false positive rates.

\end{document}